\documentclass[10pt,journal,compsoc]{IEEEtran}

\ifCLASSOPTIONcompsoc
  \usepackage[nocompress]{cite}
\else
  \usepackage{cite}
\fi

\ifCLASSINFOpdf
\else

\fi
\usepackage{times}

\usepackage{soul}
\usepackage{url}
\usepackage[hidelinks]{hyperref}
\usepackage[utf8]{inputenc}
\usepackage[small]{caption}
\usepackage{graphicx}
\usepackage{amsmath}
\usepackage{booktabs}
\usepackage{xcolor}

\usepackage{amsmath}
\usepackage{xcolor,soul,framed} 
\usepackage{xspace}
\usepackage{array}
\usepackage{eqparbox}
\usepackage{url}
\usepackage{longtable}
\usepackage{lipsum}
\usepackage{blindtext}
\usepackage{makecell}
\usepackage{mathtools}
\usepackage{commath}
\usepackage{multirow}
\usepackage{tabularx}
\usepackage[normalem]{ulem}
\usepackage{xspace}
\usepackage{algorithm}
\usepackage{algpseudocode}
\usepackage{amsfonts}
\usepackage{amssymb}% http://ctan.org/pkg/amssymb
\usepackage{pifont}% http://ctan.org/pkg/pifont
\hypersetup{
    colorlinks=true,
    filecolor=magenta,      
    urlcolor=cyan,
}

\begin{document}
\title{Deep Gait Recognition: A Survey}
\author{Alireza~Sepas-Moghaddam,~\IEEEmembership{Member,}
        and~Ali~Etemad,~\IEEEmembership{Senior Member,~IEEE}% <-this %
\thanks{Alireza Sepas-Moghaddam and Ali Etemad are with the Department of Electrical and Computer Engineering, and Ingenuity Labs Research Institute, Queen's University, Kingston, ON, Canada. Contact information: alireza.sepasmoghaddam@queensu.ca. The authors would like to thank the BMO Bank of Montreal and Mitacs for funding this research.}}

% [[]]\markboth{IEEE Transactions on Biometrics, Behavior, and Identity Science}%
% {Shell \MakeLowercase{\textit{et al.}}: Bare Demo of IEEEtran.cls for Biometrics Council Journals}

\IEEEtitleabstractindextext{%
\begin{abstract}
Gait recognition is an appealing biometric modality which aims to identify individuals based on the way they walk. Deep learning has reshaped the research landscape in this area since 2015 through the ability to automatically learn discriminative representations. Gait recognition methods based on deep learning now dominate the state-of-the-art in the field and have fostered real-world applications. In this paper, we present a comprehensive overview of breakthroughs and recent developments in gait recognition with deep learning, and cover broad topics including datasets, test protocols, state-of-the-art solutions, challenges, and future research directions. We first review the commonly used gait datasets along with the principles designed for evaluating them. We then propose a novel taxonomy made up of four separate dimensions namely body representation, temporal representation, feature representation, and neural architecture, to help characterize and organize the research landscape and literature in this area. Following our proposed taxonomy, a comprehensive survey of gait recognition methods using deep learning is presented with discussions on their performances, characteristics, advantages, and limitations. We conclude this survey with a discussion on current challenges and mention a number of promising directions for future research in gait recognition.
\end{abstract}

}

\maketitle

\IEEEdisplaynontitleabstractindextext

\IEEEpeerreviewmaketitle

\IEEEraisesectionheading{\section{Introduction}\label{sec:introduction}}

\IEEEPARstart{G}{AIT}, defined as the way people walk, contains relevant cues about human subjects~\cite{R1}. As a result, it has been widely used in different application areas such as affect analysis~\cite{Ali1,Ali2,Ali3}, sport science~\cite{sport1, sport2}, health~\cite{pato, clinic, clinic2}, and user identification~\cite{survey1, survey2, nambiar}. Gait information can be captured using a number of sensing modalities such as wearable sensors attached to the human body, for instance accelerometers, gyroscopes, and force and pressure sensors~\cite{wear}. Non-wearable gait recognition systems predominantly use vision, and are therefore mostly known as vision-based gait recognition. These systems capture gait data using imaging sensors with no cooperation from the subjects and even from far away distances~\cite{survey4}. The focus of this paper is to survey vision-based gait recognition systems that have mainly relied on deep learning. We focus solely on vision-based gait recognition as a comprehensive review paper has recently been published, surveying wearable-based gait recognition approaches~\cite{wear}.

The performance of vision-based gait recognition systems, hereafter only referred to only as gait recognition, can be affected by \textit{i)} variations in the appearance of the individual, such as carrying a handbag/backpack or wearing items of clothing such as a hat or a coat; \textit{ii)} variations in the camera viewpoint; \textit{iii)} occlusion factors, for instance where parts of the subject's body are partially covered by an object or by a part of the subject's own body in certain viewpoints (known as self-occlusion)~\cite{incomp, bar3}; and \textit{iv)} variations in the environment, such as complex backgrounds~\cite{background} and high or low levels of lighting~\cite{R7}, which generally make the segmentation and recognition processes more difficult.

\begin{figure}[!t]
\centering
\includegraphics[width=0.95\columnwidth]{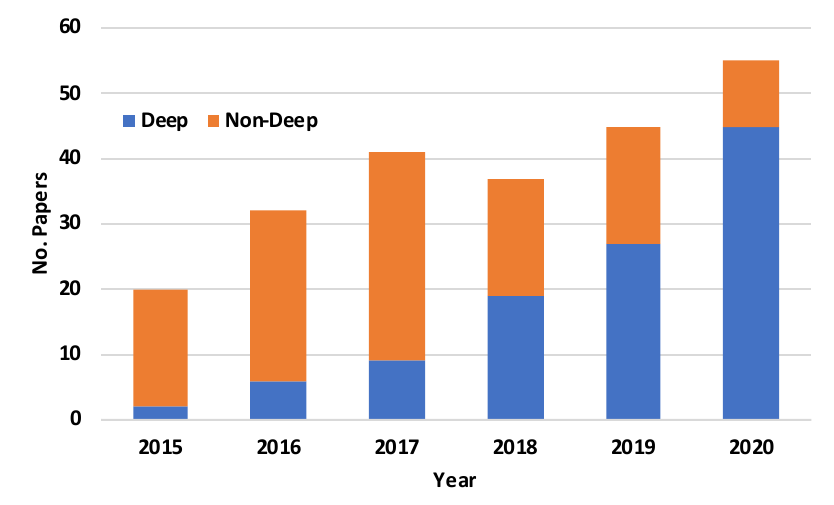}
\caption{The number of gait recognition papers published after 2015 using non-deep ({orange}) and deep ({blue}) gait recognition methods. These papers have been published in top-tier journals and conferences in the field. Journal publications include IEEE Transactions ($19\%$) including \textit{T-PAMI}, \textit{T-IP}, \textit{T-IFS}, \textit{T-MM}, \textit{T-CSVT}, and \textit{T-Biom}, as well as other top journals ($24\%$) such as \textit{Pattern Recognition} and \textit{Pattern Recognition Letter}. Conference publications include highly ranked computer vision and machine learning conferences ($22\%$) including \textit{CVPR}, \textit{AAAI}, \textit{ICCV}, \textit{ECCV}, \textit{ACCV}, \textit{BMVC}, as well as other top relevant conferences ($35\%$) such as \textit{ICASSP}, \textit{ICIP}, \textit{ICPR}, \textit{ICME}, \textit{ACM Multimedia}, and \textit{IJCB}. The figure shows clear opposing trends between the two approaches, indicating that, unsurprisingly, deep learning methods have become the dominant approach is recent years.}
\label{fig:Evolution}
\end{figure}

\begin{figure*}[!t]
\centering
\includegraphics[width=0.9\textwidth]{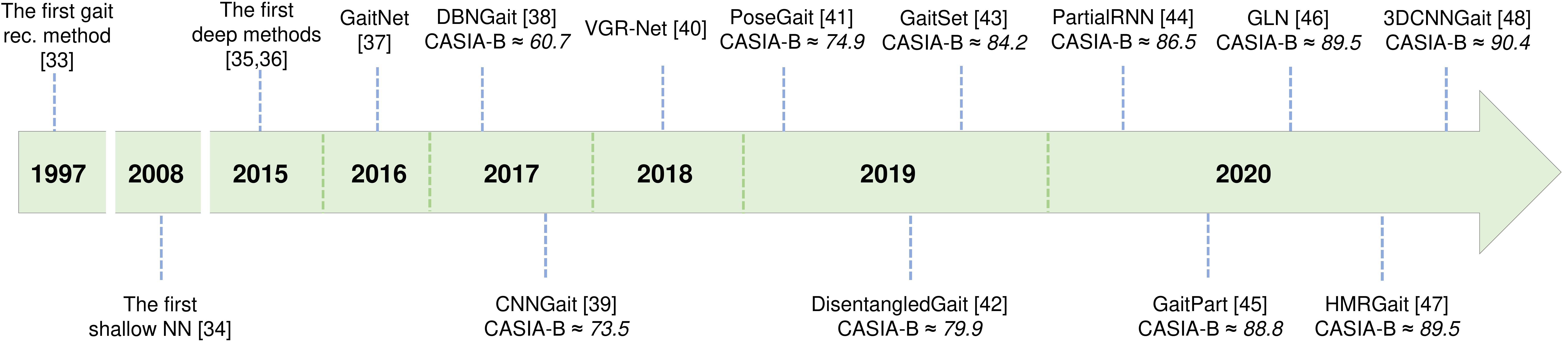}
\caption{The evolution of deep gait recognition methods.}
\label{fig:timeline}
\end{figure*}

\subsection{Unique Characteristics of Gait Recognition}
Gait recognition systems pose challenges that are unique to this field, making it a problem that demands an independent treatment. From a `biometrics' perspective, gait recognition has several unique characteristics that distinguishes it from other biometric modalities. For instance, in contrast to many other biometric systems such as face~\cite{face}, ear~\cite{ear1}, iris\cite{iris, iris2}, and fingerprint~\cite{finger} recognition that require subjects to be quite close to acquisition systems, gait data can be captured from far away distances~\cite{survey4}. As a result, gait recognition videos may often be recorded with low spatial resolution, hence many details regarding the scene become challenging to detect by automated systems. Moreover, while most biometric recognition systems need the subjects’ active cooperation towards acquisition, gait recognition data can be captured in a discrete manner~\cite{survey1}. As a result, the likelihood of recording gait patterns in an uncontrolled and non-obscured manner considerably increases. Interestingly, this very property makes gait difficult to forge by imposters, making it reliable for sensitive applications such as crime analysis~\cite{taa1}.
% Finally, many biometric modalities could be fully hidden from the observer, such as occluded faces with full masks, while gait features can not be fully hidden from the camera~\cite{taa1}.

What makes some of the challenges in gait recognition unique and distinct from general `computer vision' problems is that most gait recognition methods learn representations from analysis of the \textit{skeletons} or \textit{silhouettes} of subjects. Meanwhile, other visual classification problems often heavily rely on derived features from \textit{texture} in addition to shape and structure information. 
% The development of gait recognition systems has been influenced by person re-identification~\cite{reid_survey} and human activity recognition~\cite{activity} fields. 
For example, despite the similarities of computer vision problems such as `person re-identification'~\cite{reid_survey} and `human activity recognition'~\cite{activity} to gait recognition, gait data still pose challenges and properties that are unique to this field. 
% systems unique challenges that gait recognition problem poses specific specific challenges that are unique when compared to person re-identification~\cite{reid_survey} and human activity recognition~\cite{activity} fields.
Specifically, person re-identification methods identify subjects across multiple non-overlapping surveillance cameras, or possibly from the same camera but at different time instances. To this end, these methods aim to learn representations that capture appearance characteristics of individuals such as clothing and skin color tone, that are shared across multiple cameras~\cite{reid_feat}. On the contrary, gait recognition methods aim to learn suitable representations with which \textit{walking patterns} can be disentangled from the visual appearance of the subjects and subsequently used for classification~\cite{survey1}.
When comparing gait recognition to human activity recognition methods~\cite{wang2019deep}, the goal of the latter is to identify specific movements or actions of a subject from video clips, which can be considered as `macro' motion patterns. Meanwhile, gait characteristics can be considered nuanced `micro' patterns that sit on top of a specific activity class, namely \textit{walking}. As a result, the detection of such subtle discriminative information are often more challenging than those dealt with for activity recognition. Furthermore, given the subtlety of gait patterns that make them unique to different subjects, they can often be highly influenced by the temporary personal state of the subject, for instance, fatigue~\cite{fatigue}, excitement and fear~\cite{gaitaffect}, and even injuries~\cite{Injury}.

% where the type of the activity can be varied from a single to group actions. 
% The activities can also be changed across the video and the methods also need to consider the human-to-human and human-to-object interactions.  
% Accordingly, the features of interest for action recognition should be learned across a sequence of atomic activities; these features are highly correlated to the particular actions and interactions, as opposed to the gait recognition features that are associated with characteristic changes in the the cyclic combination of the body movements.

% personal state of the user e.g. fatigue, excitement, fear, injury can impact... while these generally do not impact activity the closest field to gait recognition, i.e., activity recognition methods, e.g., golf or tennis swing... gait sits on top of activity therefore much more nuanced.

\subsection{Motivation}

In the past two decades, many gait recognition methods have been developed to tackle the above-mentioned problems. In recent years, there has been a clear trend in migrating from non-deep methods to deep learning-based solutions for gait recognition. To visualize this trend, we present Figure \ref{fig:Evolution}, which illustrates the number of gait recognition papers published after 2015. It is observed that the majority of gait recognition methods in 2019 and 2020 have been designed based on deep neural networks. In Figure \ref{fig:timeline}, we illustrate the evolution of some of the most important gait recognition methods along with their associated accuracy on the CASIA-B~\cite{CASIA} (perhaps the most popular dataset for gait recognition) when available. The first gait recognition method was proposed in 1997~\cite{R3}, followed by the first shallow neural network for gait recognition in 2008~\cite{firstNN}, consisting of only one input layer, one hidden layer, and one output layer. In 2015, the field witnessed significant breakthroughs, notably due to the popularization of deep neural networks~\cite{SOTA1, SOTA2}. The method entitled GaitNet~\cite{B8} was then proposed in 2016 based on a 6-layer convolutional neural network (CNN). In 2017, DBNGait~\cite{DBN2} was proposed based on a deep belief network (DBN), and in~\cite{B6} three different deep CNN architectures with different depths and architectures were fused for gait recognition. VGR-Net~\cite{DIC} was one of the important contributions in 2018, followed by the introduction of several significant methods in 2019, including PoseGait~\cite{IETBiom}, DisentangledGait~\cite{B3}, and GaitSet~\cite{B1}, where the best recognition accuracy of 84.2\% was achieved by GaitSet~\cite{B1}. Remarkable advances have been made in 2020, notably by the appearance of several highly efficient methods, including PartialRNN~\cite{TBiom}, GaitPart~\cite{CVPR20201}, GLN~\cite{ECCV}, HMRGait~\cite{ACCV}, and 3DCNNGait~\cite{MM}. The current state-of-the-art results on CASIA-B dataset~\cite{CASIA} have been reported by 3DCNNGait~\cite{MM} with a recognition accuracy of 90.4\%.

% It would be a challenging task to compile a survey on the available gait recognition methods due to the diversity of the methods developed, notably after the appearance of deep methods.
Several survey papers~\cite{survey1, survey2,survey3, survey4, skl3, wear, nambiar, tax1} have so far reviewed recent advances in gait recognition, where some of these papers, for instance~\cite{skl3, wear, tax1}, have focused on \textit{non-vision-based} gait recognition methods. The most recent survey papers on \textit{vision-based} gait recognition are~\cite{survey1, survey2,survey3,survey4,nambiar}, which only cover the papers published until mid 2018. Nonetheless, many important breakthroughs in gait recognition with deep learning have occurred since 2019, as observed in Figures \ref{fig:Evolution} and \ref{fig:timeline}. Additionally, none of the surveys~\cite{survey1, survey2,survey3, survey4,nambiar} have specifically focused on deep learning methods for gait recognition. 

\subsection{Contribution}
% \subsection{Review Methodology and Contributions}
This paper surveys the most recent advances in gait recognition until the end of January 2021, providing insights into both technical and performance aspects of the deep gait recognition methods in a systematic way. In this context, we first proposes a novel taxonomy with four dimensions, i.e., body representation, temporal representation, feature representation, and neural architecture, to help characterize and organize the available methods. Following our proposed taxonomy, a comprehensive survey of all the available deep gait recognition methods is presented, along with discussions on their characteristics and performances. We have established certain search protocols to make sure other scholars can confidently use this survey in their future research.

Our key contributions are summarized as follows:
\begin{itemize}
    \item We propose a novel taxonomy with four dimensions to characterize and organize the available deep gait recognition methods.
    \item We provide a taxonomy-guided review on the evolution of the deep gait recognition methods, where most of these methods have not been reviewed in previous surveys. This provides insights for new topic exploration and future algorithm design.
    \item We present comparisons between the state-of-the-art using the available results reported on large-scale public gait datasets, providing insight into the effectiveness of different deep gait recognition methods.
    \item We review 15 publicly available vision-based datasets for gait recognition, along with their associated test protocols.
    \item We discuss a number of open challenges and identify important future research directions that will be of benefit to researchers working on gait recognition. 
\end{itemize}

% 3) DL is great... it has changed everything\\
% 4) impact of DL in gait + some stats\\
% 5) problem statement: in recent surveys, none of them are deep and vision-based and many works have been done since 2018 + contributions\\
% 6) searched keywords, how many papers\\
% 7) timeline...\\
% 8) structure of the paper + the reason for the structure
\subsection{Organization}

The rest of this survey is structured as follows. We start with describing the systematic approach used to collect the papers and review the literature. Next, in Section 3, we review the available gait datasets along with their associated test protocols. We then use these datasets and protocols to report the existing performance results when reviewing the deep gait recognition methods. Section 4 presents our proposed taxonomy. Following, Section 5 surveys the state-of-the-art on deep gait recognition and discusses the evolutional trends of deep gait recognition over the past few years. Finally, Section 6 discusses some deep gait recognition challenges and identifies a number of future research areas.

\section{Review Methodology}

We employed a search protocol to ensure other scholars can confidently use this survey in their future research. To do so, we first discovered candidate papers through the Google Scholar~\cite{google} search engines and digital libraries, namely IEEE Xplore~\cite{IEEE}, ACM Digital Library~\cite{ACM}, ScienceDirect~\cite{ELS}, and CVF Open Access~\cite{CVF}. Our search terms included combinations of the following queries: ``gait recognition'', ``gait identification'', ``gait biometric'', ``neural architecture'', ``deep learning'', and ``deep representations''. We then filtered the search results, thus excluding papers that neither use deep learning methods for gait recognition nor demonstrate enough technical clarity/depth. To be more specific about the `clarity/depth' criteria, we excluded the papers that \textit{i}) use non-vision sensors for gait recognition; \textit{ii}) do not propose a new solution; \textit{iii}) use non-standard or private datasets for performance evaluation; \textit{iv}) do not compare the performance of their solution to the state-of-the-art. In cases where other modalities were combined with vision-based sensors, only the technical solution focusing on the vision-based aspect was studied.

\begin{table*}
  \centering
  \setlength\tabcolsep{2.3pt}
    \caption{Summary of well-known gait datasets used in the literature.}
    \begin{tabular}{l|l|l|l|l|l|l}
    \hline
    \textbf {Dataset}& \textbf{Year}& \textbf{Data Type} & \textbf{\# of Subjects}& \textbf{Environment} & \textbf{\# of} & \textbf{Variations} \\
    \textbf { }& \textbf{ }& \textbf{ } & \textbf{\& Sequences}& \textbf{ } & \textbf{Views} & \textbf{ } \\
    \hline\hline
     CMU MoBo~\cite{DB-MOBO}& 2001 & RGB; Silhouette & 25 / 600 & Indoor & 6 & 3 Walking Speeds; Carrying a Ball\\
    %  &  & Silhouette &  &  &  & Carrying a Ball\\
    % \hline
    SOTON~\cite{DB-SOTON} & 2002 & RGB; Silhouette & 115 / 2,128 & Indoor \& Outdoor & 2 & Normal Walking on a Treadmill\\
    % &  & Silhouette &  &  &  & \\
    % \hline
    % Georgia Tech~\cite{DB-Georgia} & 2003& 3D RGB & 15 & 540 & 1 & 4 Walking Speeds\\
    % \hline
    CASIA-A~\cite{DB-CASIAA}& 2003 & RGB & 20 / 240 & Outdoor & 3 & Normal Walking\\
    % \hline
    USF HumanID~\cite{DB-HumanID} & 2005 & RGB & 122 / 1,870 & Outdoor & 2 & Outdoor Walking; Carrying a Briefcase; Time Interval\\
    % &  &   &  &  &  & \\
    % &  &   &  &  &  & \\

    % \hline
    CASIA-B~\cite{CASIA} & 2006 & RGB; Silhouette & 124 / 13,680 & Indoor & 11 & Normal Walking; Carrying a Bag; Wearing a Coat\\
    % &  & Silhouette &  &  &  & \\
    % &  &   &  &  &  & \\
    
    % \hline
    CASIA-C~\cite{DB-CASIAC} & 2006 & Infrared; Silhouette & 153 / 1,530 & Outdoor & 1 & 3 Walking Speeds; Carrying a Bag \\
    % &  & Silhouette  &  &  &  & \\
    % \hline

    OU-ISIR Speed~\cite{ISRT1}  & 2010 & Silhouette & 34 / 306 & Indoor & 4 & Nine walking speeds\\
    % \hline
    
    OU-ISIR Clothing~\cite{ISRT2}  & 2010 & Silhouette & 68 / 2,746 & Indoor & 4 & Up to 32 combinations of clothing\\
    % \hline

    % OU-ISIR Speed~\cite{ISRT3}  & 2012 & Silhouette & 4,007 & 31,368 & 4 & Normal Walking\\
    % \hline

    OU-ISIR MV~\cite{ISRT4}  & 2010 & Silhouette & 168 / 4,200 & Indoor & 25 & 24 azimuth views and 1 top view\\
    % \hline

    OU-ISIR~\cite{DB-OU}  & 2012 & Silhouette & 4,007 / 31,368 & Outdoor & 4 & Normal Walking\\
    % \hline
    TUM GAID~\cite{DB-TUM}  & 2012 & RGB; Depth; Audio & 305 / 3,737  & Indoor & 1 & Normal Walking; Backpack; Wearing coating shoes\\
    % &  & Depth  &  &  &  & \\
    % &  & Audio  &  &  &  & \\
    % \hline
    % KY IR Shadow~\cite{DB-Shadow} & 2014 & Shadow Silhouette & 54 & 324 & 1 & Normal Walking; Carrying a Bag; Changing the Clothes\\
    %     &  &  &  &  &  & \\
    % &  &   &  &  &  & \\
    % \hline

    OU-ISIR LP Bag~\cite{OUBAG}  & 2017 & Silhouette & 62,528 / 187,584     & Indoor  & 1  & Seven different carried objects \\
    % \hline
    
    OU-MVLP~\cite{MVLP}  & 2018 & Silhouette & 10,307 / 259,013  & Indoor & 14 & Normal Walking\\
    % \hline
    
    CASIA-E~\cite{SOTA23, CASIAE}  & 2020 &  Silhouette  &  1014 / Undisclosed   &  Indoor \& Outdoor & 15  & 3 Scenes; Normal Walk; Carrying a Bag; Wearing a Coat  \\
    % \hline 
    
    OU-MVLP Pose~\cite{Tbiom2}  & 2020 & Skeleton & 10,307 / 259,013  & Indoor & 14 & Normal Walking\\
    \hline

    \end{tabular}%
  \label{tabDB}%
\end{table*}%

Naturally, we imposed restrictions on the date of publications to only include search results after 2014, when deep neural networks were first used for biometric recognition~\cite{deepsurvey, sundararajan2018deep}. We then used the returned results in order to perform forward and backward searches, respectively identifying the other resources that have cited the returned articles and the references cited by the returned articles. We repeated this process with the new identified resources until we collected the most relevant papers to the best of our knowledge. We eventually ended up with a final set of publications that have used deep learning for gait recognition.

\section{Test Protocols and Datasets}

\subsection{Protocols}

\begin{figure}[!t]
\centering
\includegraphics[width=1\columnwidth]{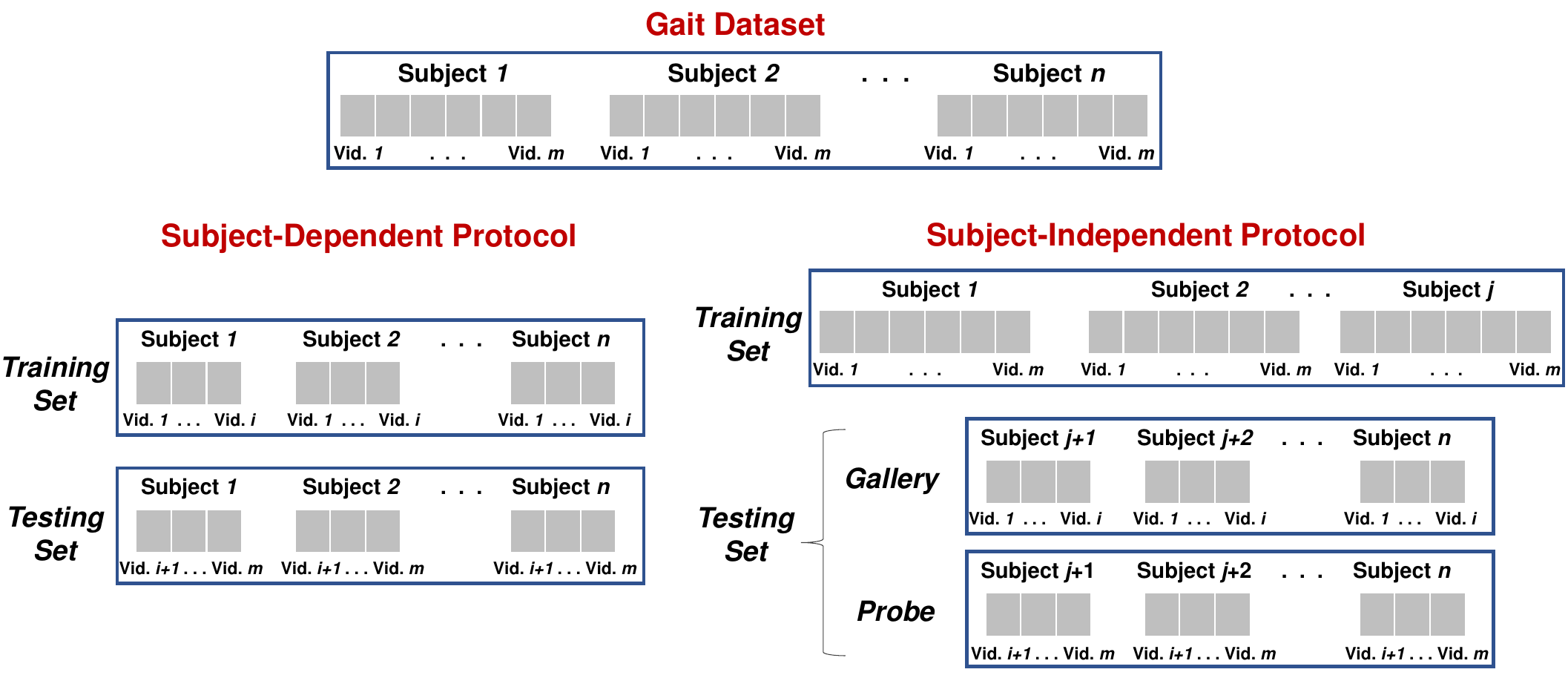}
\caption{An overview of test protocols is presented. These protocols can be categorized into subject-dependent or subject-independent according to whether test subjects appear in the training set or not.}
\label{fig:protocols}
\end{figure}

Evaluation protocols for gait recognition solutions can generally be categorized into \textit{subject-dependent} and \textit{subject-independent}. As illustrated in Figure~\ref{fig:protocols}, in the subject-dependent protocol, both the training and testing sets include samples from all the subjects. However, in the subject-independent protocol, the test subjects are disjoint from the training subjects. Here, the test data are further divided into gallery and probe sets, and the learned model on the disjoint training subjects is then used to extract features from gallery and probe sets. Finally, a classifier is used to compare the probe features with the gallery ones in order to identify the most similar gait patterns and label them as being from the same identity. Both subject-dependent and subject-independent protocols have been widely adopted for gait recognition. For example, in the TUM GAID~\cite{DB-TUM} dataset, subject-dependent protocols have been often used, while in the CASIA-B~\cite{CASIA} and OU-MVLP~\cite{MVLP} large-scale datasets, subject-independent protocols are utilized. Gait recognition results in the literature have all been measured and presented using rank-1 recognition accuracy.

\subsection{Datasets}

In order to evaluate gait recognition systems, different datasets have been collected. These datasets cover various parameters related to acquisition viewpoints, environment conditions, and appearance of the subjects. Generally speaking, large datasets, both in terms of number and distribution of samples and parameters, are often desired and preferred to allow for deep neural networks to be trained effectively. We present an overview of the main characteristics of well-known gait datasets in Table \ref{tabDB}. These characteristics include the type and modality of data, number of subjects and sequences, number of viewpoints, and also the variations covered in each dataset. To show the chronological evolution of these datasets, we have sorted these datasets by the order of release date in Table \ref{tabDB}. According to Table \ref{tabDB}, CASIA B~\cite{CASIA}, CASIA-E~\cite{SOTA23, CASIAE}, OU-ISIR MV~\cite{ISRT4}, and OU-MVLP~\cite{MVLP} cover the highest number of acquisition viewpoints while OU-ISIR~\cite{DB-OU}, OU-ISIR LP Bag~\cite{OUBAG}, and OU-MVLP~\cite{MVLP} include the highest number of gait sequences. In the following we review the gait datasets available in Table \ref{tabDB} along with the associated test protocols.

% Interestingly, these datasets are the widely used ones among the our reviewed papers for deep gait recognition. In the following we review the gait datasets available in Table \ref{tabDB} along with the associated test protocols for evaluation study.

\textbf{CMU MoBo:} The CMU Motion of Body (MoBo)~\cite{DB-MOBO} dataset is one of the first gait datasets in the literature, and consists of RGB and silhouette data from 25 different subjects who walk on a treadmill. This dataset covers three subsets including slow walking speed, fast walking speed, and walking when holding a ball. The test protocol proposed by the authors of this dataset suggests training with one subset and testing with the other subsets. All six combinations are often used to report the results.  

\textbf{SOTON}: The SOTON dataset~\cite{DB-SOTON} contains data from 115 subjects. All the sequences have been recorded both indoor and outdoor, with a fixed camera, capturing the subjects walking along a straight path. The indoor gait data has been captured from the subjects when walking on a treadmill. Different papers have divided this dataset into training and testing sets differently, and there is no pre-defined test protocol presented with the dataset.

\textbf{CASIA-A}: CASIA-A~\cite{DB-CASIAA} is a dataset that includes data from 20 subjects in outdoor environments. Participants have walked along a straight line, while three cameras positioned at 0$^{\circ}$, 45$^{\circ}$, and 90$^{\circ}$ have captured the gait videos with an average of 90 frames per a sequence. A cross-view test protocol is the most widely used protocol for this dataset, where solutions are trained with all the available views, excluding one which is then used for testing.

\textbf{USF HumanID}: The USF HumanID dataset~\cite{DB-HumanID} has been collected in the context of the HumanID gait challenge, and includes outdoor gait videos from 122 subjects who have walked in an elliptical path. This dataset covers challenging variations including carrying a briefcase, walking on different surfaces, wearing different shoes, and with acquisition times. The data has been captured from two viewing angles by left and right cameras. The evaluation study has been made available along with the dataset~\cite{DB-HumanID}, which considers 12 different test protocols with respect to the above-mentioned variations. 

\textbf{CASIA-B}: CASIA-B dataset~\cite{CASIA} is the most widely used gait dataset, and contains multi-view gait data from 124 persons in the form of both RGB and silhouettes. Acquisition has been performed from 11 different viewing angles that range from 0$^{\circ}$ to 180$^{\circ}$ with 18$^{\circ}$ increments. The dataset considers three different walking conditions namely normal walking (NM), walking with a coat (CL), and walking with a bag (BG), respectively with 6, 2, and 2 gait sequences per person per view. The most frequently used test protocol for CASIA-B is a subject-independent protocol which uses the data from the first 74 subjects for training, and the remaining 50 subjects for testing. The test data is then split into a gallery set including the first four gait sequences from the NM gait data and the probe set consists of the rest of the sequences, namely the remaining 2 NM, 2 CL, and 2 BG sequences, per each subject per each view. The results have been mostly reported for all the viewing angles angles, excluding the probe sequences with angles identical to the references.

\textbf{CASIA-C}: The CASIA-C dataset~\cite{DB-CASIAC} includes infrared and silhouette data from 153 different subjects, and the sequences have been captured under different variations at night. These variations include three different walking speeds namely slow walking (SW), normal walking (NW), and fast walking (FW), as well as carrying a bag (BW). There are 4 NW, 2 SW, 2 FW, and 2 BW sequences per each subject. As for the evaluation scheme, cross-speed walker identification tests have been considered.

\textbf{OU-ISIR Speed}: The OU-ISIR Speed dataset~\cite{ISRT1} provides silhouette data from 34 subjects. This dataset is suitable for evaluation of robustness of gait recognition methods with respect to walking speeds, as it includes nine different speeds, ranging from 2 km/h to 11 km/h, with 1 km/h interval. Cross-speed tests have been adopted for this dataset.

\textbf{OU-ISIR Clothing}: The OU-ISIR Clothing dataset~\cite{ISRT2} includes data from 68 subjects who wore up to 32 different types of clothing. Gait sequences were collected in two indoor acquisition sessions on the same day. A subject-independent test protocol has been provided along with the dataset~\cite{DB-OU}, which divides the data into pre-defined training, testing, and probe sets particularly with respect to the clothing conditions.

\textbf{OU-ISIR MV}: The OU-ISIR MV dataset~\cite{ISRT4} consists of gait silhouettes from 168 subjects with an age range of 4 to 75 years old, and almost equal number of male vs. female participants. The data has been captured from a large range of view variations, including 24 azimuth views, along with 1 top view. A cross-view test protocols have been widely adopted for this dataset.

\textbf{OU-ISIR}: The OU-ISIR dataset~\cite{DB-OU} is a large-scale gait dataset, consisting of gait data from 4,007 subjects with almost equal gender distribution, and with ages ranging from 1 to 94 years old. The gait sequences have been captured in two different acquisition sessions in indoor halls using four cameras placed at 55$^{\circ}$, 65$^{\circ}$, 75$^{\circ}$, and 85$^{\circ}$ degrees. As there are two sequences available for each subject, the test protocol uses the first sequences as gallery and the other one as probe samples. 

\textbf{TUM GAID}: The TUM GAID~\cite{DB-TUM} is a multi-modal gait dataset, including RGB, depth, and audio data from 305 subjects. For a selected set of 32 subjects, the dataset has been captured in two different outdoor acquisition sessions in winter and summer. 10 sequences have been captured from each subject, including normal walking (N), walking with a backpack (B), and walking with disposable shoe covers (S). The test protocol has been made available by the original authors, dividing the data into training, validation, and test sets. Recognition experiments are often then carried out with respect to the N, B, and S gait variations.

\textbf{OU-ISIR LP Bag}: The OU-ISIR LP Bag dataset~\cite{OUBAG} consists of gait videos from 62,528 subjects with carried objects, captured using one camera in constrained indoor environments. Three sequences have been obtained per each subject, one with a carried object and two without it. Following the test protocol proposed in \cite{OUBAG}, the training set contains data from 29,097 subjects with two sequences with and without the carried objects, and the test set includes the other 29,102 disjoint subjects. In order to divide the test set into probe and gallery sets, two approaches have been adopted, respectively under cooperative and uncooperative scenarios. For the cooperative scenario, the gallery set only contains sequences without carried objects, where the probe set includes sequences with seven different types of carried objects. In the uncooperative scenario, gallery and probe sets are randomly formed such that they both contain sequences with and without carried objects.

\begin{figure*}[!t]
\centering
\includegraphics[width=2\columnwidth]{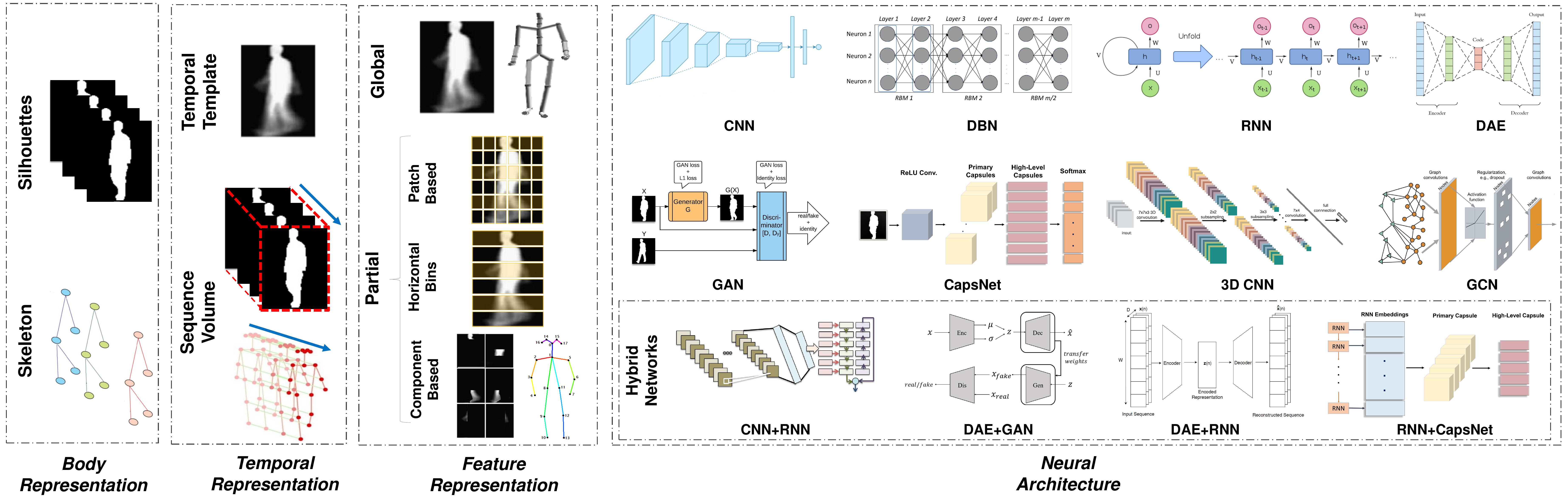}
\caption{Our taxonomy consisting of 4 dimensions: body representation, temporal representation, feature representation, and neural architecture.}
\label{fig:taxonomy}
\end{figure*}

\textbf{OU-MVLP}: The OU-MVLP dataset~\cite{MVLP} is the largest available gait dataset in term of number of gait sequences (259,013). The dataset provides videos of silhouettes and is acquired in two acquisition sessions per each subject. The gender distribution of subjects is almost equal with an age range of 2 to 87 years old. This dataset has been acquired from 14 different views, ranging from 0$^{\circ}$ to 90$^{\circ}$, and 180$^{\circ}$ to 270$^{\circ}$, where the angle change in each step is 15$^{\circ}$. Pre-determined lists of 5153 and 5154 subjects have been designated and provided with the dataset as training and testing sets respectively. For testing, sequences from the first and second acquisition sessions respectively form gallery and probe sets. In most of the recent gait recognition papers, either all or four viewing angles, notably 0$^{\circ}$, 30$^{\circ}$, 60$^{\circ}$, and 90$^{\circ}$, are considered.

\textbf{CASIA-E}: CASIA-E dataset~\cite{SOTA23} consists of silhouettes from 1014 subjects with hundreds of sequences per subject, captured in three types of scenes with a simple static, a complex static, and a complex dynamic background. The data has been captured considering three different walking conditions, including normal walking (NM), walking with a coat (CL), and walking with a bag (BG). This dataset has been acquired from 15 different angles, including two vertical views with a height of 1.2 \textit{m} and 3.5 \textit{m}, as well as 13 horizontal views ranging from 0$^{\circ}$ to 180$^{\circ}$ with 15$^{\circ}$ increments. This dataset was recently used in the \textit{TC4 Competition and Workshop on Human Identification at a Distance 2020}~\cite{Compet}, where the training set included the entire data from the first 500 subjects, while 25 sequences from the last 514 subjects were used for validation. The remaining sequences were used for testing.

\textbf{OU-MVLP Pose}: The OU-MVLP Pose dataset~\cite{Tbiom2} was built upon the OU-MVLP~\cite{MVLP}, extracting pose skeleton sequences from the RGB images available in OU-MVLP. Two subsets have been created using pre-trained versions of OpenPose~\cite{openpose} and AlphaPose~\cite{Alpha} to extract human joint information. The test protocol is similar to the one proposed for OU-MVLP~\cite{MVLP}.

\section{Proposed Taxonomy}

% It would be a challenging task to compile a survey on the available gait recognition methods due to the diversity of the methods developed, notably after the appearance of deep learning. 
In order to help illustrate an overall structure for the available gait recognition approaches, a few taxonomies have been proposed in the literature~\cite{tax1, tax2, R7}, which have organized the available solutions from different perspectives. The taxonomy proposed in~\cite{tax1} is based on the type of sensors, classifiers, and covariate factors such as occlusion types. The taxonomy in~\cite{tax2} categorizes gait recognition methods based on the type of features used. Finally, the one proposed in~\cite{R7} considers user appearance, camera, light source, and environment-related factors. Nevertheless, despite the availability of these taxonomies, none focus on deep gait recognition methods that are most successful nowadays. We thus propose a new taxonomy in this paper to better illustrate the technological landscape of gait recognition methods with a particular focus on deep learning techniques. 
% This taxonomy helps characterizing and organizing the available deep gait recognition methods, as done in Section 4, thus guiding the researchers in this area to develop more efficient recognition methods. 
Figure \ref{fig:taxonomy} presents our proposed taxonomy which considers four dimensions, namely body representation, temporal representation, feature representation, and neural architecture. The details of each of these dimensions are described in the following.

\subsection{Body Representation}
This dimension relates to the way the body is represented for recognition, which can be based on \textit{silhouettes} or \textit{skeletons}. Silhouette is the most frequently used body representation in the literature that can be easily computed by subtracting each image containing the subject from its background, followed by binarization. Gait silhouettes are proven to be effective and convenient for describing the body state in a single frame with low computational cost. This body representation forces recognition solutions to focus on `gait' as opposed to clothing and other non-gait factors that could, from the perspective of a classifier, be used for identification.
% and with high ability to preserve privacy.
% , as well as forcing solutions to focus on body and motion for identification as opposed to facial features. 
A sequence of silhouettes can represent useful gait features such as speed, cadence, leg angles, gait cycle time, step length, stride length, and the the ratio between swing and stance phases~\cite{sil, TanmayThesis}. It can also be processed to extract motion data, for example using optical flow calculation~\cite{opt1, opt2, IETBiom}. Nonetheless, gait silhouettes are more sensitive to changes in the appearance of the individuals, for instance via different clothing and carrying conditions. 
% Additionally, the privacy of the participants can be preserved by providing gait data in form of binary gait silhouettes, which is why some gait datasets such as OU-ISIR~\cite{DB-OU} and OU-MVLP~\cite{MVLP}, only provide silhouettes. 
% It should be noted that some of the important gait datasets, such as OU-ISIR~\cite{DB-OU} and OU-MVLP~\cite{MVLP} datasets, only provide gait data in form of binary gait silhouettes to preserve the privacy of the participants, so naturally the body silhouette is the only available gait representation for gait recognition when using those datasets.  

Skeleton body representation can be captured using depth cameras~\cite{skl1} or alternatively be estimated using pose-estimation methods~\cite{pose}. Static and dynamic features, for instance stride length, speed, distances, and angles between joints, can be obtained from skeleton joints~\cite{skl3}. Gait recognition methods based on this type of body representation are generally more robust against viewpoint changes due to the consideration of joint positions~\cite{skel}, as opposed to silhouette-based methods. Skeleton-based methods are also more robust against appearance changes~\cite{TanmayThesis} as the pose-estimation step generally learns to detect body joints over different clothing conditions, which is not the case for gait silhouettes. However, since these approaches rely heavily on accurate detection of body joints, they are generally more sensitive to occlusions~\cite{TanmayThesis}. Additionally, the use of pose-estimators imposes a computational overhead to these recognition systems~\cite{posecomp}.

\subsection{Temporal Representation}

This dimension deals with approaches used to represent the temporal information in gait sequences. Two types of representations, \textit{templates} and \textit{volumes}, have been commonly used in the literature. Following we describe these representations.

\begin{figure}[!t]
\centering
\includegraphics[width=0.70\columnwidth]{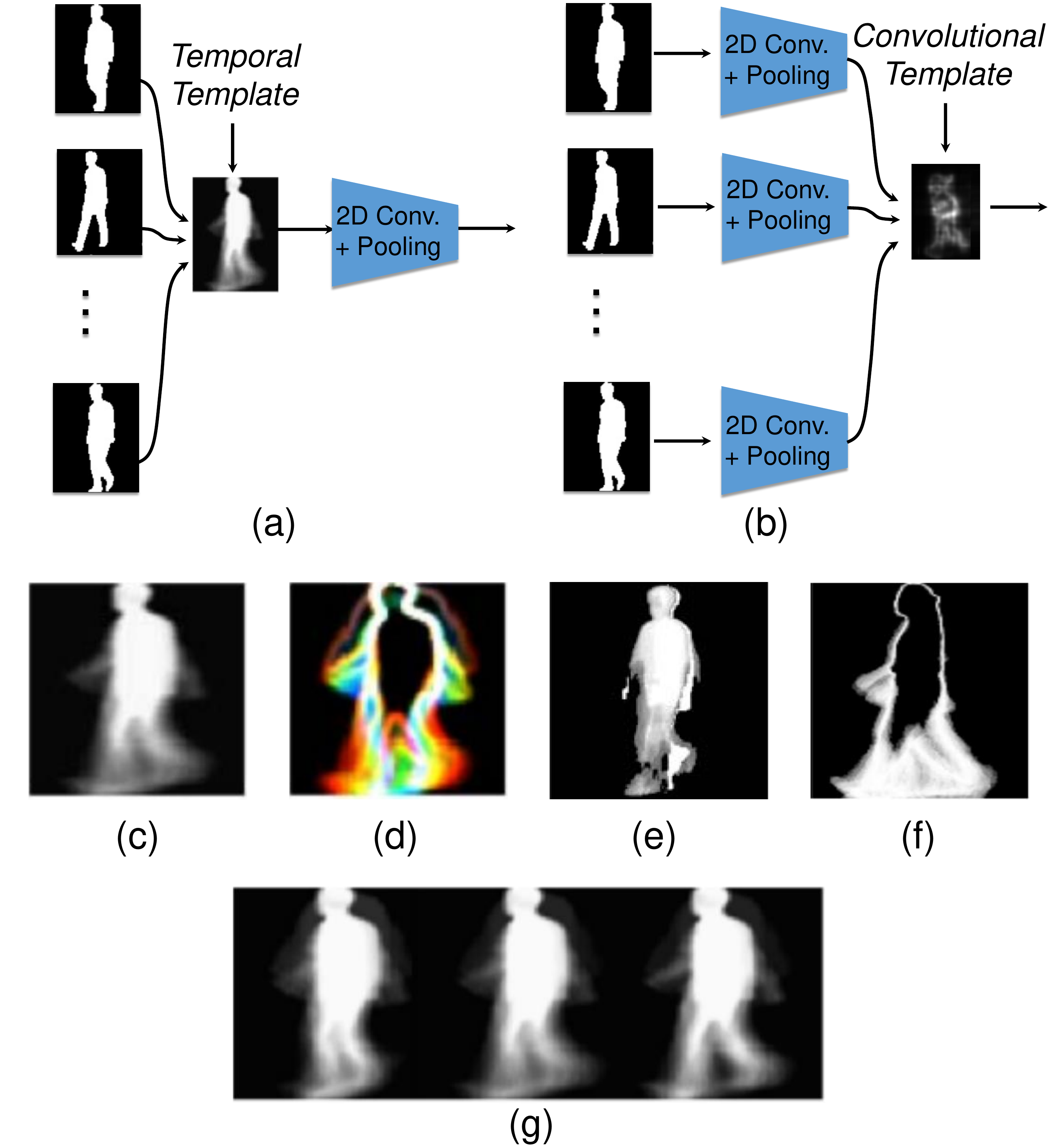}
\caption{Overview of temporal representations. Generating templates in: (a) the initial layer of a deep network; (b) an intermediate layer of the network after several convolution and pooling layers. Illustration of (c) GEI~\cite{GEI}, (d) CGI~\cite{GEI1}, (e) FDEI~\cite{FDEI}, (f) GEnI~\cite{GENI}, and (g) PEI~\cite{B7} temporal gait templates.}
\label{fig:template_Layer}
\end{figure}

Templates aggregate temporal walking information over a sequence of silhouettes in a single map, for example by averaging the silhouettes over at least one gait cycle. This operation enables recognition solutions to be independent of the number of frames once template maps have been created. With respect to deep gait recognition architectures, gait silhouettes can be aggregated in the initial layer of a network (Figure \ref{fig:template_Layer}-a), also known as \textit{temporal templates}, where the aggregated map can then be processed by subsequent layers~\cite{GEI, GEI1, FDEI, GENI, B7}. Gait silhouettes can alternatively be aggregated in an intermediate layer of the network after several convolution and pooling layers (Figure \ref{fig:template_Layer}-b), also known as \textit{convolutional template}~\cite{B1, TBiom}. 
% In this context, convolutional maps obtained using several convolution and pooling layers are combined to from a convolutional template. 
Examples of temporal templates include: (\textit{i}) gait energy images (GEI)~\cite{GEI}, which average gait silhouettes over one period/sequence (Figure \ref{fig:template_Layer}-c); (\textit{ii}) chrono gait images (CGI)~\cite{GEI1}, which extract the contour in each gait image to be then encoded using a multi-channel mapping function in the form a single map (Figure \ref{fig:template_Layer}-d); (\textit{iii}) frame-difference energy images (FDEI)~\cite{FDEI}, which preserve the kinetic information using clustering and denoising algorithms, notably when the silhouettes are incomplete (Figure \ref{fig:template_Layer}-e); (\textit{iv}) gait entropy images (GEnI)~\cite{GENI}, computing entropy for each pixel in the gait frames to be then averaged in a single gait template (Figure \ref{fig:template_Layer}-f); and (\textit{v}) period energy images (PEI)~\cite{B7}, a generalization of GEI that preserves more spatial and temporal information by exploiting a multi-channel mapping function based on the amplitudes of frames (Figure \ref{fig:template_Layer}-g). Examples of convolutional templates include set pooling~\cite{B1} and gait convolutional energy maps (GCEM)~\cite{TBiom}, which average convolutional maps obtained by several convolution and pooling layers, over the whole sequence.

To preserve and learn from the order and relationship of frames in gait sequences, instead of aggregating them, sequence volume representations have be adopted (see Figure~\ref{fig:taxonomy}, second box from the left). Then, to learn the temporal information, two different approaches have been adopted. In the first approach, the temporal dynamics over the sequences are learned using recurrent learning strategies, for example recurrent neural networks, where each frame is processed with respect to its relationships with the previous frames~\cite{SOTA30, LSTM3, ACCV}. The second approach first creates 3D tensors from spatio-temporal information available in sequences, where the depth of the tensors represent the temporal information. These tensors are then learned, for example using 3D CNNs~\cite{MM,SOTA22, arXiv} or graph convolutional networks (GCNs)~\cite{GCN2}. 
% The features learned using such approaches are generally more robust to camera and appearance changes. 

% \begin{figure}[!t]
% \centering
% \includegraphics[width=0.85\columnwidth]{Images/Template.pdf}
% \caption{Illustration of (a) GEI~\cite{GEI}, (b) CGI~\cite{GEI1}, (c) FDEI~\cite{FDEI}, (d) GEnI~\cite{GENI} and e) PEI~\cite{B7} temporal gait templates.}
% \label{fig:template}
% \end{figure}

\subsection{Feature Representation}

This dimension encapsulates the region of support for representation learning, which can be either \textit{global} or \textit{partial}. 
% Global representations are learned when the silhouette or skeleton information are incorporated as a whole. 
% Global representation indicates that silhouette or skeleton information is holistically learned.
The process of learning silhouettes or skeletons holistically is referred to as global representation learning. 
On the other hand, when learning partial representations, gait data is split into local regions, e.g., patches, body components, and vertical/horizontal bins (see Figure \ref{fig:taxonomy}, third box from the left). These local regions are then further processed, for example by recurrent neural networks~\cite{TBiom}, capsule networks~\cite{ICPR}, attention-based networks~\cite{part}, or fully connected layers~\cite{B1}. Methods based on global representations tend to be more sensitive to occlusions and appearance changes as well as missing key body parts~\cite{R9, B1}. On the other hand, partial regions often maintain different contributions towards the final recognition performance, thus learning their importance can improve the overall performance of gait recognition methods~\cite{CVPR20201, B1}. Additionally, the relations between these partial features can be learned, to preserve positional attributes such as scale, rotation, and location, which improve the robustness of gait recognition methods against orientation and view changes~\cite{ICPR, TBiom}.

\subsection{Neural Architectures}

Deep neural networks (DNNs) capture high-level abstractions using hierarchical architectures of multiple nonlinear transformations. Different neural architectures have been designed for gait recognition problems, whose descriptions are provided below.

\subsubsection{Convolutional Neural Networks}

Convolutional neural networks (CNNs) have been used the most for gait recognition. {These models are generally used to learn an embedding where the body shape, represented as a silhouette or skeleton, is encoded in the feature space.
% where the model reads either across each frame of the input video or a temporal template in order to extract feature representation.
}
Specifically, CNNs generally consist of different types of layers including convolutional, pooling, and fully connected layers. Convolutional layers convolve learned filters with the input image to create activation feature maps that capture features with varying levels of detail. The convolutional layers also include activation functions such as a ReLU~\cite{ReLU} or a tanh~\cite{tanh} functions, to increase the non-linearity in the output. 
% learn the correlations between neighboring pixels, in form of activation feature maps.
Pooling layers then reduce the spatial size of the feature maps by using nonlinear down-sampling strategies, such as average or maximum pooling, thus decreasing the complexity of the network. Fully connected layers are finally used to learn the resulting 2D feature maps into 1D vectors for further processing.

To better analyze CNNs adopted in the state-of-the-art gait recognition methods, we provide an overview of the most successful used architectures in Table \ref{tab:CNN}. Note that for the methods combining CNNs with other types of networks, e.g., autoencoder, capsule, and Long Short-Term Memory (LSTM), we only present the architectures of the CNN components in the table. {As can be seen, there is no need for state-of-the-art gait recognition models to exploit very deep CNN architectures. This is due to the fact that input gait data, either in the from of silhouettes or skeletons, do not present considerable complexity in terms of texture information. Hence, even fewer than 10 layers are shown to be sufficient for encoding gait frames. This is contrary to many other domains, such as face or activity recognition, where very deep networks such as ResNet~\cite{resnet} and inception~\cite{inception} are used to learn highly discriminative features.} In Table \ref{tab:CNN}, we also present the size of CNN inputs, showing a trend toward a 64$\times$64 resolution in the recent literature. Additionally, an analysis in~\cite{B6} showed that the resolutions of 64$\times$64 and 128$\times$128 lead to the best gait recognition results for several tested CNNs, where the input resolution of 128$\times$128 works slightly better than 64$\times$64. However, as a higher input resolution implies more convolutional and pooling layers, the input resolution of 64$\times$64 has been most widely adopted to limit the computational complexity of the solutions.

\subsubsection{Deep Belief Networks}

A deep belief network (DBN)~\cite{DBN} is a probabilistic generative model, composed by staking restricted Boltzmann machines (RBMs)~\cite{RBM} to extract hierarchical representations from the training data. Each RBM is a two-layer generative stochastic model, including a visible and a hidden layer, with connections between the adjacent layers and without connections between the units within each layer. The weights and biases of the units define a probability distribution over the joint states of the visible and hidden units. DBNs have been used for gait recognition in~\cite{Biom} and~\cite{DBN2}. In~\cite{Biom}, fitting, body parameters, and shape features were extracted from silhouettes to be then learned by DBNs, thus extracting more discriminative features. In~\cite{DBN2}, gait has been first represented as motion and spatial components, and two separate DBNs were trained for each component. The extracted features were finally concatenated to represent the final feature.

\subsubsection{Recurrent Neural Networks}

Recurrent Neural Networks (RNNs) have been widely applied to temporal or sequence learning problems, achieving competitive performances for different tasks~\cite{RNNSurvey}, including gait recognition~\cite{LSTM3, LSTM1, LSTM2, TBiom, ICPR, B3,B4}. A layer of RNN is typically composed of several cells, each corresponding to one input element of the sequence, e.g., one frame of a gait video. RNNs can also stacks several layers to make the model deeper, where the output of the $i^{th}$ cell in $j^{th}$ layer feeds the $i^{th}$ cell in the $(j+1)^{th}$ layer. Each cell is connected to its previous and subsequent cells, thus memorizing information from the previous time steps~\cite{RNNSurvey}. Among different RNN architectures, LSTM~\cite{LSTM} and Gated Recurrent Units (GRU)~\cite{GRU1} are the most widely used RNN architectures that have been used to learn the relationships available in a gait sequence using memory states and learnable gating functions. In an LSTM network~\cite{LSTM}, the cells have a common cell state, which keeps long-term dependencies along the entire LSTM cell chain, controlled by two gates, the so-called input and forget gates, thus allowing the network to decide when to forget the previous state or update the current state with new information. The output of each cell, the hidden state, is controlled by an output gate that allows the cell to compute its output given the updated cell state. GRU~\cite{GRU1} is another form of RNN that does not use output activation functions as opposed to LSTM. This architecture also includes an update gate that allows the network to update the current state with respect to the new information. The output of the gate, also known as the reset gate, only maintains connections with the cell input.

{There have been three different approaches to use RNNs for gait recognition systems. The first approach~\cite{LSTM3} (Figure \ref{fig:taxonomy}-a) that has been mostly adopted for skeleton representations, uses RNNs in order to learn from temporal relationships of joint positions. In the second approach~\cite{LSTM1,LSTM2} (Figure \ref{fig:taxonomy}-b), as will be discussed later in detail in Section \ref{sec:hib}, RNNs are combined with other types of the neural architectures, notably CNNs, for learning both spatial and temporal information. The last approach that has been recently adopted in~\cite{TBiom,ICPR} (Figure \ref{fig:taxonomy}-c) uses RNNs to recurrently learn the relationships between partial representations from a single gait template, for instance GCEM~\cite{TBiom}}.

\begin{table}
  \centering
  \setlength\tabcolsep{2.5pt}
%   \scriptsize
    \caption{Overview of recent CNN architectures adopted for deep gait recognition.}
    \begin{tabular}{l|l|l|l|l|l}
    \hline
    \textbf {Method}& \textbf{Input}& \textbf{Total \# of}& \textbf{\# of Conv.} & \textbf{\# of Pool.} & \textbf{\# of FC}  \\
     & \textbf{Size} & \textbf{Layers}& \textbf{Layers} & \textbf{Layers} & \textbf{Layers}  \\
    \hline\hline
    
    GEINet~\cite{B8} & 88$\times$128  & 6 & 2 & 2 & 2  \\
    Ensem. CNNs~\cite{B6} & 128$\times$128 & 7 &  3 & 2 & 2 \\
    MGANs~\cite{B7} & 64$\times$64 & 8 &  4 & 1 & 3 \\
    EV-Gait~\cite{B2} & 128$\times$128 & 9 & 6 & 0& 2   \\
    Gait-joint~\cite{B10} & 64$\times$64  & 16 & 12 & 2& 2  \\
    Gait-Set~\cite{B1} & 64$\times$64 & 9 &  6 & 2 & 1 \\ 
    Gait-RNNPart~\cite{TBiom} & 64$\times$64 & 9 &  6 & 2 & 1 \\ 
    Gait-Part~\cite{CVPR20201} & 64$\times$64 & 9 &  6 & 2 & 1 \\ 
    SMPL~\cite{ACCV} & 64$\times$64 & 5 &  3 & 1 & 1 \\ 
    Caps-Gait~\cite{ICPR} & 64$\times$64 & 9 &  6 & 2 & 1 \\ 
    
    \hline
    \end{tabular}%
  \label{tab:CNN}%
\end{table}%

\begin{figure}[!t]
\centering
\includegraphics[width=1 \columnwidth]{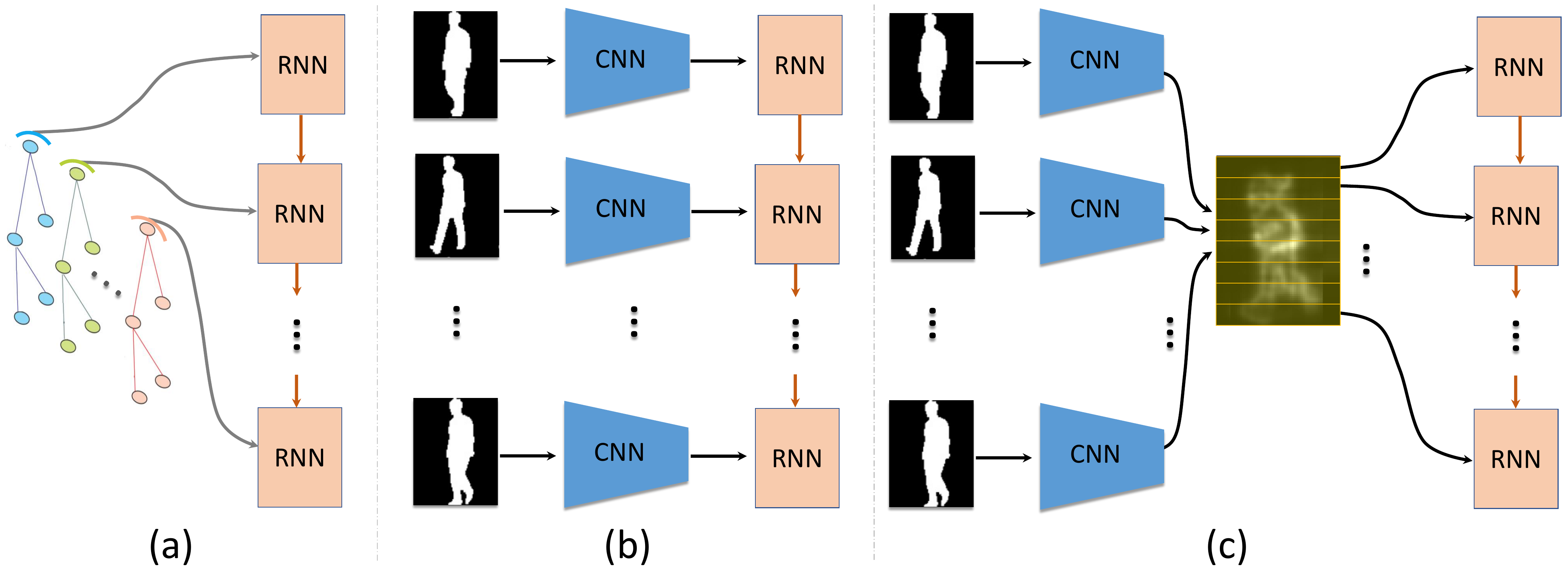}
\caption{Three different approaches for using RNNs in the context of deep gait recognition systems: (a) RNNs directly learn from the movement of joint positions; (b) RNNs are combined with CNNs; and (c) RNNs recurrently learn the relationships between partial representations in gait templates.}
\label{fig:RNN}
\end{figure}

\subsubsection{Deep AutoEncoders}

Deep auto-encoder (DAE) is a type of network that aims to extract so called bottleneck features or latent space representations, using an encoder-decoder structure. The encoder transforms the input data into a feature representation and the decoder part transforms the representation back to the original input data. The encoder generally includes several fully connected and/or convolutional layers while the decoder consists of layers that perform the inverse operations. DAE networks are generally trained with the aim of minimizing the reconstruction error that measures the difference between the original input and the reconstructed version. Once a DAE is trained, the bottleneck features which are a latent/compressed representation of the knowledge of the original input, are extracted to be used for classification, i.e., gait recognition in our case. The method proposed in~\cite{R6} uses a DAE network, first encoding the input temporal templates using four convolutional layers to extract feature. The decoder then reconstructs the input from the extracted features using four deconvolutional layers. In~\cite{DAE1}, an auto-encoder with 7 fully connected layers along with input and output layers was used to extract robust gait features. In~\cite{CVPR20202}, a DAE was used to disentangle the input temporal template into identity and covariate features. The backbone of the encoder was based on the Inception module in GoogLeNet~\cite{GoogleNet}, extracting multi-scale identity and covariate features. The decoder then took those features as input to reconstruct the temporal template using deconvolutional layers.

\subsubsection{Generative Adversarial Networks}

Generative Adversarial Networks (GANs) include a generator and a discriminator~\cite{goodfellow}, where the generator aims to deceive the discriminator by synthesizing fake samples that resemble the real ones. In turn, the discriminator aims to distinguish between the fake and real samples. As a result of this minimax game between these two components, GANs can generate realistic synthesized samples. {In the context of gait recognition, GANs can be used to solve the problem of gait variations due to clothing, viewpoints, and carrying conditions. For instance, GANs can transform gait data from one view to another, or change the type of clothing worn by the subject, or even remove a backpack that was originally carried by the subject.
% with multiple variations into the particular viewpoint, e.g., side view, with normal clothing and without the carrying conditions. 
Such disentanglement of identity from confounding factors often results in improvements in the performance of gait recognition systems~\cite{GAN5, GAN6, B7, B9, GAN3, GAN4}. Nevertheless, one of the most important challenges toward manipulating gait data is the preservation of human identity features while modifying appearance characteristics in the representation space. To this end, two discriminators are often used~\cite{B9}. The first discriminator is used to distinguish real vs. fake samples in order to ensure that the generated images appear realistic. The second discriminator is exploited to ensure that identity information are preserved by taking a pair of source and target images as input and producing a scalar probability of whether the input pair belongs to the same person or not.}

% These networks can preserve identity information while transferring gait variations such as pose and clothing along low-dimensional manifolds in a process referred to as domain adaptation. This disentangles identity and gait variations, often resulting in improvements in the performance of gait recognition systems~\cite{GAN5, GAN6, B7, B9, GAN3, GAN4}.}

Different types of GANs have recently been adopted for gait recognition. Multi-task GAN (MGANs)~\cite{B7} have been proposed for cross-view gait recognition, where a CNN is used to learn the temporal template as view-specific features in a latent space. Then, the features are transformed from one view to another using a view transform layer. The network is then trained with multi-task adversarial and pixel-wise losses. In another paper, Discriminant Gait GAN (DiGGAN)~\cite{B9} considered the mechanisms of using two independent discriminators in order to transfer GEIs form a certain viewpoint to a different viewing angle while also preserving identity information. In~\cite{GAN3} a Two-Stream GAN (TS-GAN) was proposed to learn both global and partial feature representations when transforming GEI temporal templates with different viewing angles to a GEI temporal templates with a standard view, i.e., 90$^{\circ}$.

\subsubsection{Capsule Networks}

Capsule Networks (CapsNet)~\cite{caps} have been proposed to address two important shortcomings in CNNs, namely the limits of scalar activations and poor information routing through pooling operations by respectively exploiting capsule activation values and routing-by-agreement algorithms. CapsNets are composed of capsules which are groups of neurons that explicitly encode the intrinsic viewpoint-invariant relationships available in different parts of the objects. {In the context of gait representation learning, a CapsNet can model and understand the structural relationship between the various parts of the body, such as the relationships between legs and feet, upper body and lower body, and trunk and limbs, using a learnable pose matrix. A CapsNet can also be used to model internal hierarchical representations between multiple gait silhouettes or skeleton joint coordinates of a subject in a video. This is in contrast to the standard pooling layers often used in CNNs, which fail to preserve positional attributes in the human body, such as locations, scales, rotations, and relationships between the body parts.} CapsNets generally include two blocks, primary and high-level group of capsules. The first block encodes spatial information with several layers including convolutional, reshaping, and squashing layers, followed by the second block that learns deeper part-whole relationships between hierarchical sub-parts. The concept of capsule network has been recently adopted for gait recognition~\cite{capsule1, ICPR, spyder}. The method proposed in~\cite{capsule1} first learns the properties of GEI templates using a CNN. It then uses a CapsNet with dynamic routing to retain the relationship within each template with the aim of finding more robust features. Capsule networks have also been combined with other types of deep networks in~\cite{ICPR} and ~\cite{spyder}, which we will review in Section \ref{sec:hib}.

\subsubsection{3D Convolutional Neural Networks}

{3D Convolutional neural networks (3D CNNs) have been recently adopted for gait recognition to learn spatio-temporal dynamics over whole gait sequences~\cite{MM, 3DCNNICIP, arXiv,MM}. 3D CNNs are able to extract features that are more robust to changes in camera viewpoints and the appearance of subjects. 3D CNNs take the stacked gait frames in the form of a 3D tensor as input, and then use multiple 3D convolution filters and pooling operations to extract the spatio-angular representations. The limitation of 3D CNNs for gait recognition is the lack of flexibility in processing variable length sequences. In~\cite{MM}, this shortcoming was addressed by exploiting multiple 3D CNNs to integrate temporal information at different scales.} In~\cite{3DCNNICIP}, a 3D CNN network containing 13 3D convolution filters and pooling layers along with two fully connected layers was designed for gait recognition. The method in~\cite{arXiv} is composed of several global and partial 3D convolutional layers, where the standard 3D pooling layer was modified to aggregate temporal information in local clips. 
% Finally, a state-of-the-art gait recognition method was proposed in~\cite{MM}, which simultaneously employs multiple 3D CNNs to extract spatio-temporal features. 
% This method uses multiple- temporal-scale framework for integrating temporal information at small and large scales. 

\subsubsection{Graph Convolutional Networks}

Graph convolutional networks (GCNs) have been recently developed to extend CNNs to a higher dimensional domain using arbitrarily structured graphs and graph convolution filters~\cite{GCNFilter}. {Given the inherent hierarchical and graph-like nature of the human body, GCNs can jointly model both the structural information of the human body and temporal relationships available between gait frames in order to learn discriminative and robust features with respect to camera viewpoints and subject appearances. Gait recognition methods based on GCNs consider gait sequence volumes as the spatio-temporal representations for gait recognition~\cite{GCNFilter, GCN2}}. In~\cite{GCN2}, gait features were extracted by forming a spatio-temporal graph from the available video sequences. The final features were then obtained using a joint relationship learning scheme by mapping the features onto a more discriminative subspace with respect to human body structure and walking pattern.

\subsubsection{Hybrid Networks} \label{sec:hib}

A large number of hybrid deep networks that make use of two or more types of networks have been proposed to boost the performance of gait recognition systems. Among these, architectures with CNN+RNN, DAE+GAN, DAE+RNN, and RNN+CapsNet components are the most popular in the deep gait recognition literature (see Figure \ref{fig:taxonomy}). Following we provide the descriptions and examples of these four hybrid architectures.

\textbf{CNN+RNN.} Integration of CNNs with RNNs (notably LSTM and GRU) for learning the temporal relationships following spatial encoding is perhaps the most popular approach for spatio-temporal learning, which has also been used for gait recognition in the literature. 
% In this context, CNNs first extract spatial features from the single frames with either global or partial representations; the relations between these features are then recurrently learned to extract more discriminative features. 
In~\cite{CNNLSTM1}, a deep gait recognition system was proposed by combining eight different CNN architectures with LSTM to obtain spatio-temporal features from image sequences. The proposed method in~\cite{TIP} first divides gait silhouettes into 4 horizontal parts, where each part was fed to an individual CNN with 10 layers. An attention-based LSTM was then used to output frame-level attention scores for each sequence of CNN features. The CNN features were finally multiplied by their corresponding weights to selectively focus on the most important frames for gait recognition. In~\cite{TBiom}, convolutional maps from gait frames were first learned using an 8-layer CNN. The convolutional maps were then aggregated to form GCEM templates which were then split into horizontal bins. These partial features (horizontal bins) were finally learned by an attentive bi-directional GRU to exploit the relations between these parts of the embedding.

\textbf{DAE+GAN.} Recently, DAEs have been considered as the backbone of the generator and/or discriminator components in GANs for gait recognition~\cite{GAN4, GAN5,GAN6, DAEGAN2}. GaitGAN~\cite{GAN5} and GaitGANv2~\cite{GAN6} used two discriminators with encoder-decoder structures, respectively for fake/real discrimination and identification. These two discriminators ensured that the generated gait images were realistic and that the generated images contained identity information. The Alpha-blending GAN (Ab-GAN) proposed in~\cite{GAN4} exploits an encoder-decoder network as the generator to generate gait templates without carried objects. Cycle-consistent Attentive GAN (CA-GAN) was proposed in~\cite{DAEGAN2} and used an encoder-decoder structure for gait view synthesis. The proposed GAN contains two branches to simultaneously exploit both global and partial feature representations.

\textbf{DAE+RNNs.} 
The combination of DAEs and RNNs has recently been proposed for generating sequence-based disentangled features using an LSTM RNN~\cite{B3, B4}. In this context, a deep encoder-decoder network with novel loss functions was first used to disentangle gait features, namely identity information from appearance and canonical features that mostly contain spurious information for gait recognition. A multi-layer LSTM was then used to capture temporal dynamics of the gait features to be finally aggregated for the recognition purpose~\cite{B3, B4}.

\textbf{RNNs+CapsNets.} Recurrently learned features obtained by RNNs can be treated as capsules~\cite{caps}, thus learning coupling weights between these capsules through dynamic routing. This encapsulated hierarchical part-whole relationships between the recurrently learned features that can make the hybrid network more robust against appearance and view changes. Additionally, the CapsNet can act as an attention mechanism, thus assigning more importance to the more relevant features. In~\cite{ICPR}, a CapsNet was used to treat the recurrently learned partial representations of a convolution template as capsules, thus learning the coupling weights between the partial features. This led to exploiting the relationships between the partial features while also preserving positional attributes. So, the model could generalize better to unseen gait viewpoints during testing. In ~\cite{spyder}, a capsule network with dynamic routing was used to exploit the spatial and structural relations between body parts. In this context, the recurrently learned features were first extracted using an LSTM network from a sequence of gait frames to feed the capsule network.

\begin{table*}[]
\centering
\setlength\tabcolsep{1.80pt}
\caption{Classification of deep gait recognition methods based on our proposed taxonomy.}
\begin{tabular}{l|l|l|l|l|l|l|l|l}
\hline 
\textbf{Ref.} &\textbf{Year} & \textbf{Venue}         & \textbf{Body Rep.}                 & \textbf{Temporal Rep.}     & \textbf{Feat. Rep.}    & \textbf{Neural Architecture}   & \textbf{Loss Function}      & \textbf{Dataset}                         \\ \hline\hline
\cite{TMM}      & 2015 & \textit{T-MM}           & Silhouettes           & Tmp: GEI          & Global          & CNN                   & Cross-Entropy                   & CASIA-B                         \\
\cite{SOTA1}    & 2015 & \textit{CISP}          & Silhouettes           & Tmp: GEI          & Global          & CNN                   & Undisclosed                     & CASIA-B                         \\
\hline
\cite{SOTA2}    & 2016 & \textit{ICPR}          & Skeleton              & Sequence Volume  & Partial         & LSTM                   & Undisclosed                     & CASIA-B                         \\
\cite{B8}       & 2016 & \textit{ICB}           & Silhouettes           & Tmp: GEI          & Global          & CNN                   & Cross-Entropy                   & OU-ISIR                         \\
\cite{3DCNNICIP}& 2016 & \textit{ICIP}          & Silhouettes           & Sequence Volume  & Global          & 3D CNN                & Undisclosed                      & CMU Mobo; USF HumanID           \\
\cite{SOTA3}    & 2016 & \textit{ICASSP}        & Silhouettes           & Tmp: GEI          & Global          & CNN                   & Contrastive                     & OU-ISIR                         \\
\cite{LSTM3}    & 2016 & \textit{BMVC}          & Skeleton              & Sequence Volume  & Global          & CNN + LSTM            & Cross-Entropy                    & CASIA-B; CASIA-A                \\
\hline
\cite{Biom}     & 2017 & \textit{Int. J. Biom.} & Silhouettes           & Tmp: GEI          & Partial         & DBN                   & Undisclosed                             & CASIA-B                         \\
\cite{SOTA4}    & 2017 & \textit{CVIU}          & Silhouettes           & Tmp: GEI          & Global          & CNN                   & Undisclosed                             & CASIA-B                         \\
\cite{B6}       & 2017 & \textit{IEEE T-PAMI}   & Silhouettes           & Tmp: GEI          & Global          & CNN                   & Cross-Entropy                   & CASIA-B; OU-ISIR                \\
\cite{SOTA5}    & 2017 & \textit{Applied Sci.}  & Silhouettes           & Tmp: Norm. AC     & Global          & CNN                   & Undisclosed                             & OU-ISIR                         \\
\cite{SOTA6}    & 2017 & \textit{IEEE T-CSVT}   & Silhouettes           & Tmp: GEI          & Global          & CNN                   & Triplet Loss                    & OU-ISIR                         \\
\cite{SOTA7}    & 2017 & \textit{BIOSIG}        & Silhouettes           & Tmp: GEI          & Global          & CNN                   & Undisclosed                             & TUM-GAID                        \\
\cite{SOTA31}   & 2017 & \textit{MM}            & Silhouettes           & Tmp: GEI          & Global          & CNN                   & Triplet Loss                    & OU-ISIR                         \\
\cite{SOTA30}   & 2017 & \textit{CCBR}          & Skeleton              & Sequence Volume  & Global          & CNN + LSTM             & Undisclosed                             & CASIA-B                         \\
\cite{GAN5}     & 2017 & \textit{CVPRW}         & Silhouettes           & Tmp: GEI          & Global          & GAN                   & Cross-Entropy                   & CASIA-B                         \\
\cite{DAE1}     & 2017 & \textit{Neurocomp.}    & Silhouettes           & Tmp: GEI          & Global          & DAE                   & Euclidean                       & CASIA-B; SZU RGB-D              \\
\hline
\cite{SOTA22}   & 2018 & \textit{Elect. Imaging}& Silhouettes           & Sequence Volume  & Global          & 3D CNN                & Undisclosed                             & CASIA-B                         \\
\cite{CNNLSTM1} & 2018 & \textit{IEEE Access}   & Silhouettes           & Sequence Volume  & Global          & CNN + LSTM            & Cross-Entropy                   & CASIA-C                         \\
\cite{SOTA21}   & 2018 & \textit{Neuroinform.}  & Silhouettes           & Seq. Vol. + GEI  & Global          & 3D CNN                & Contrastive                     & OU-ISIR                         \\
\cite{DIC}      & 2018 & \textit{DIC}           & Skeleton              & Tmp: GEI          & Global          & CNN                   & Cross-Entropy                  & CASIA-B                         \\
\cite{SOTA8}    & 2018 & \textit{IEEE Access }  & Silhouettes           & Sequence Volume  & Global          & CNN + LSTM            & Cross-Entropy                   & CASIA-B; OU-ISIR                \\
\cite{SOTA9}    & 2018 & \textit{ISBA}          & Silhouettes           & Sequence Volume  & Global          & 3D CNN                & Undisclosed                             & CASIA-B                         \\
\cite{DAEGAN2}  & 2018 & \textit{ICME}          & Silhouettes           & Tmp: GEI          & Part; Glob.      & DAE + GAN             & Cross-Entropy                   & CASIA-B                         \\
\cite{SOTA17}   & 2018 & \textit{JVCIR}         & Silhouettes           & Tmp: GEI          & Global          & CNN                   & Cross-Entropy                   & CASIA-B; OU-ISIR                \\
\cite{SOTA18}   & 2018 & \textit{CCBR}          & Skeleton              & Sequence Volume  & Global          & CNN + LSTM           & Undisclosed                             & CASIA-B                         \\
\hline
\cite{PRL}      & 2019 & \textit{PRL}           & Skel.; Silh.          & Sequence Volume  & Global          & LSTM                 & Undisclosed                             & CASIA-B; TUM-GAID               \\
\cite{IETBiom}  & 2019 & \textit{IET Biom.}     & Silhouettes           & Tmp: Weight Avg.  & Partial         & CNN                   & Undisclosed                             & CASIA-B; TUM; OU-ISIR      \\
\cite{B3}       & 2019 & \textit{CVPR}          & Skel.; Silh.          & Sequence Volume  & Global          & DAE + LSTM            & Multiple Loss Functi
ons         & CASIA-B; FVG                    \\
\cite{SOTA27}   & 2019 & \textit{J. Sys. Arch.} & Silhouettes           & Tmp: GEI          & Global          & DAE + GAN             & Multiple Loss Functions         & CASIA-B; OU-ISIR                \\
\cite{B10}      & 2019 & \textit{PR}            & Silhouettes           & Tmp: GEI          & Global          & CNN                   & Siamese                         & CASIA-B; SZU                    \\
\cite{B7}       & 2019 & \textit{IEEE T-IFS }   & Silhouettes           & Tmp: GEI          & Global          & GAN                   & Adversarial\&Cross-Entropy      & CASIA-B; OU-ISIR                \\
\cite{SOTA24}   & 2019 & \textit{PRL}           & Silhouettes           & Tmp: GEI          & Global          & CNN                   & Restrictive Triplet             & CASIA-B; OU-ISIR                \\
\cite{SOTA10}   & 2019 & \textit{CVPR}          & Silhouettes           & Tmp: GEI          & Global          & CNN                   & Quintuplet                      & CASIA-B; OU-ISIR LP Bag         \\
\cite{GAN3}     & 2019 & \textit{Neurocomp.}    & Silhouettes           & Tmp: GEI          & Global          & GAN                   & Pixel-wise and Entropy          & CASIA-B; OU-ISIR                \\
\cite{SOTA26}   & 2019 & \textit{IJCNN}         & Silhouettes           & Tmp: GEI          & Global          & GAN                   & Multiple Loss Functions         & CASIA-B                         \\
\cite{R6}       & 2019 & \textit{IEEE T-IFS}    & Skeleton              & Tmp: GEI          & Global          & DAE                   & Contrastive\&Triplet Loss       & OU-ISIR LP Bag; TUM-GAID        \\
\cite{opt1}     & 2019 & \textit{ICVIP}         & Silhouettes           & Tmp: Weight Avg.  & Partial         & CNN                   & View\&Cross-Entropy             & CASIA-B                         \\
\cite{SOTA25}   & 2019 & \textit{IEEE T-MM}     & Silhouettes           & Sequence Volume  & Global          & CNN + LSTM             & Contrastive                     & CASIA-B; OU-ISIR                \\
\cite{SOTA29}   & 2019 & \textit{IJCNN}         & Silhouettes           & Tmp: GEI          & Global          & GAN                   & Multiple Loss Functions         & CASIA-B                         \\
\cite{SOTA28}   & 2019 & \textit{NCAA}          & Silhouettes           & Tmp: GEI          & Global          & CNN                   & Undisclosed                             & CASIA-B; CASIA-A; OU-ISIR       \\
\cite{SOTA23}   & 2019 & \textit{PR}            & Silhouettes           & Tmp: Set pooling  & Global          & CNN                   & Center\&Soft-Max                & CASIA-B                         \\
\cite{NCAA}     & 2019 & \textit{NCAA}          & Silhouettes           & Tmp: GEI          & Global          & CNN                   & Undisclosed                             & CASIA-B; OU-ISIR                \\
\cite{capsule2} & 2019 & \textit{JVCI}          & Silhouettes           & Tmp: GEI          & Global          & CapsNet               & Standard Capsule Loss           & CASIA-B                         \\
\cite{B1}       & 2019 & \textit{AAAI}          & Silhouettes           & Tmp: Set Pooling  & Partial         & CNN                   & Batch All Triplet loss          & CASIA-B; OU-MVLP                \\
\hline
\cite{RNNAE}    & 2020 & \textit{IEEE Access}   & Skeleton              & Sequence Volume  & Global          & DAE + LSTM            & Mean Square Error               & Walking Gait                    \\
\cite{PR5}      & 2020 & \textit{PR}            & Skeleton              & Sequence Volume  & Partial         & CNN                   & Center\&Soft-Max                & CASIA-B; CASIA-E                \\
\cite{B4}       & 2020 & \textit{IEEE T-PAMI}   & Skel.; Silh.          & Sequence Volume  & Global          & DAE + LSTM            & Multiple Loss Functions         & CASIA-B; FVG                    \\
\cite{TIP}      & 2020 & \textit{IEEE T-IP }    & Silhouettes           & Sequence Volume  & Partial         & CNN + LSTM            & Angle Center                    & CASIA-B; OU-MVLP; OU-LP         \\
\cite{GAN4}     & 2020 & \textit{PR}            & Silhouettes           & Tmp: GEI          & Global          & GAN                   & Multiple Loss Functions         & OULP-BAG; OU-ISIR LP Bag        \\
\cite{SOTA11}   & 2020 & \textit{MTAP}          & Silhouettes           & Tmp: MF-GEI       & Global          & CNN                   & Undisclosed                             & CASIA-B                         \\
\cite{spyder}   & 2020 & \textit{KBS}           & Silhouettes           & Sequence Volume  & Global          & LSTM + Capsule        & Capsule\&Memory                 & CASIA-B; OU-MVLP                \\
\cite{SOTA16}   & 2020 & \textit{JINS}          & Silhouettes           & Tmp: GEI          & Global          & CNN + LSTM           & Undisclosed                             & CASIA-B; OU-ISIR                \\
\cite{CSVT}     & 2020 & \textit{IEEE T-CSVT}   & Silhouettes           & Tmp: GEI          & Global          & CNN                   & Contrastive\&Triplet Loss       & CASIA-B; OU-MVLP; OU-ISIR       \\
\cite{arXiv}    & 2020 & \textit{arXiv}         & Silhouettes           & Sequence Volume  & Global          & 3D CNN                & Triplet Loss                    & CASIA-B; OU-MVLP                \\
\cite{MTAP2}    & 2020 & \textit{MTAP}          & Silhouettes           & Tmp: GEI          & Partial         & CNN                   & Undisclosed                             & CASIA-B; OU-ISIR                \\
\cite{GCN2}     & 2020 & \textit{arXiv}         & Skeleton              & Sequence Volume  & Global          & GCN                   & Triplet Loss\&ArcFace           & CASIA-B                         \\
\cite{MTAP3}    & 2020 & \textit{MTAP}          & Silhouettes           & Tmp: GEI          & Global          & CNN                   & Undisclosed                             & CASIA-B; OU-ISIR                \\
\cite{SOTA15}   & 2020 & \textit{JIPS}          & Silhouettes           & Tmp: GEI          & Global          & CNN                   & Soft-Max                        & CASIA-B; OU-ISIR                \\
\cite{MTAP4}    & 2020 & \textit{MTAP}          & Silhouettes           & Tmp: GEI          & Global          & CNN                   & Undisclosed                             & CASIA-B                         \\
\cite{Supercom} & 2020 & \textit{J. SuperComp.} & Silhouettes           & Tmp: GEI          & Global          & CNN                   & Undisclosed                             & CASIA-B; OU-ISIR; OU-MVLP       \\
\cite{capsule2} & 2020 & \textit{ITNEC}         & Silhouettes           & Tmp: GEI          & Global          & CapsNet               & Standard Capsule Loss           & CASIA-B; OU-ISIR                \\
\cite{CVPR20201}& 2020 & \textit{CVPR}          & Silhouettes           & Tmp: Hor. Pooling & Partial         & CNN                   & Batch All Triplet loss          & CASIA-B; OU-MVLP                \\
\cite{CVPR20202}& 2020 & \textit{CVPR}          & Silhouettes           & Tmp: GEI          & Global          & DAE                   & Contrastive\&Triplet Loss       & CASIA-B; OU-ISIR LP Bag         \\
\cite{Tbiom2}   & 2020 & \textit{IEEE T-Biom}   & Skeleton              & Sequence Volume  & Global          & CNN + LSTM             & Cross-Entropy\&Center           & OUMVLP-Pose                     \\
\cite{ACCVW1}   & 2020 & \textit{ACCVW}         & Silhouettes           & Tmp: Set Pooling  & Partial         & CNN                   & Batch All Triplet loss          & CASIA-E                         \\
\cite{ACCVW2}   & 2020 & \textit{ACCVW}         & Silhouettes           & Tmp: Set Pooling  & Partial         & CNN                   & Batch All Triplet loss          & CASIA-E                         \\
\cite{ICPR}     & 2020 & \textit{ICPR}          & Silhouettes           & Tmp: Set Pooling  & Partial         & CNN + GRU + Caps.     & Triplet Loss\&Cosine Prox.    & CASIA-B; OU-MVLP                \\
\cite{TBiom}    & 2020 & \textit{IEEE T-Biom.}  & Silhouettes           & Tmp: GCEM         & Partial         & CNN + GRU             & Triplet Loss\&Cross-Entropy     & CASIA-B; OU-MVLP                \\
\cite{SOTA20}   & 2020 & \textit{IEEE Access}   & Silhouettes           & Tmp: Set Pooling  & Global          & CNN                   & Triplet Loss                    & CASIA-B                         \\
\cite{ICASSP3}  & 2019 & \textit{ICASSP}        & Silhouettes           & Tmp: Pooling      & Global          & CNN                   & Center-Ranked                   & CASIA-B; OU-MVLP                \\
\cite{SOTA19}   & 2020 & \textit{IJCB}          & Silhouettes           & Tmp: GEI          & Global          & DAE + GAN             & Center\&Soft-Max                & CASIA-B; OU-ISIR                \\
\cite{ACCV}     & 2020 & \textit{ACCV}          & Skel.; Silh.          & Sequence Volume  & Global          & CNN + LSTM            & Multiple Loss Functions         & CASIA-B; OU-MVLP                \\
\cite{MM}       & 2020 & \textit{MM}            & Silhouettes           & Sequence Volume  & Global          & 3D CNN                & Multiple Triplet Losses                    & CASIA-B; OU-ISIR                \\
\cite{ECCV}     & 2020 & \textit{ECCV}          & Silhouettes           & Tmp: Set Pooling  & Global          & CNN                   & Triplet Loss\&Cross-Entropy     & CASIA-B; OU-MVLP                \\

\hline

\end{tabular}
\label{tab:Status}
\end{table*}

% \begin{figure*}[!t]
% \centering
% \includegraphics[width=1\textwidth]{Images/Graph.eps}
% \caption{Visualization of deep gait recognition methods, according to three levels of our taxonomy and publication date.}
% \label{fig:tax}
% \end{figure*}

\begin{figure*}[!t]
\centering
\includegraphics[width=1\textwidth]{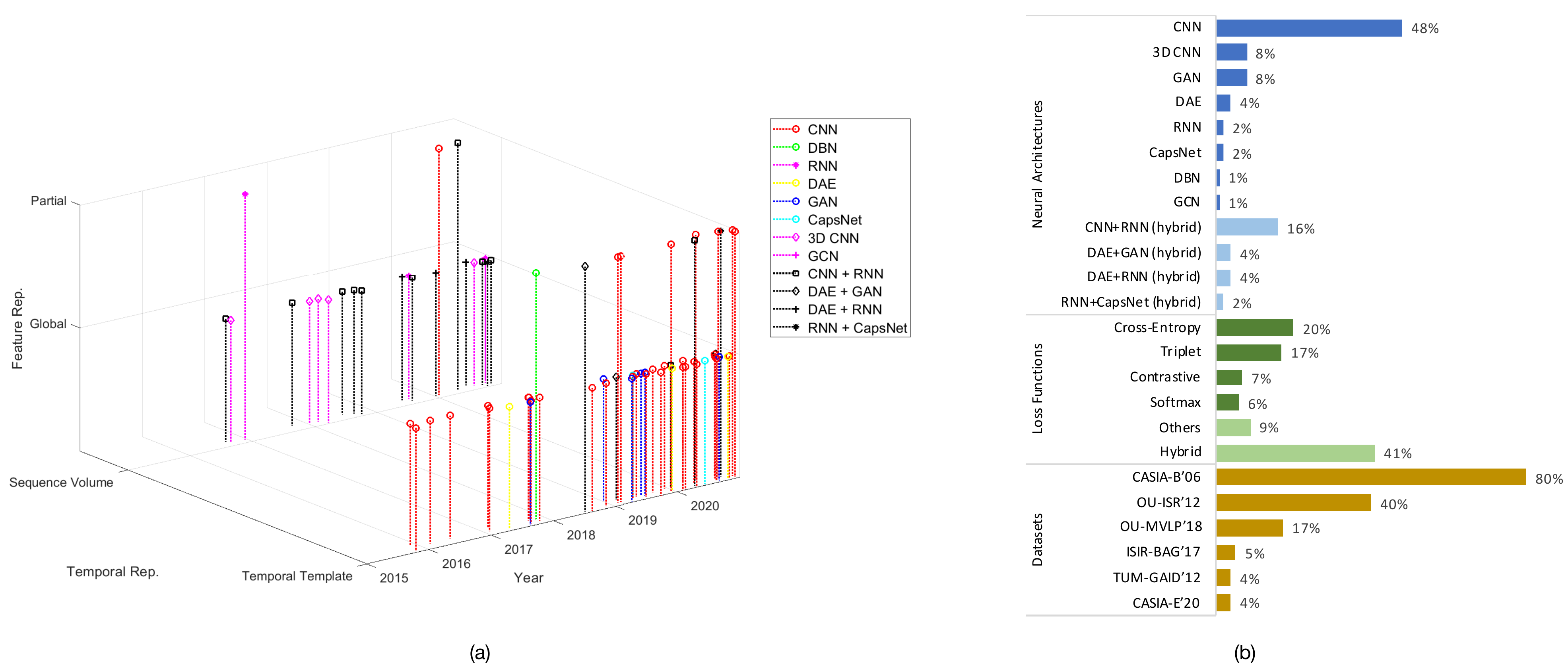}%Graph.eps
\caption{(a) Visualization of deep gait recognition methods, according to three levels of our taxonomy and publication date; (b) The frequency of different neural architectures, loss functions, and gait datasets used in the literature.}
\label{fig:tax}
\end{figure*}

\section{State-of-the-art}

% \subsection{\textit{Status quo}}

In this section, we use our proposed taxonomy to survey deep gait recognition methods available in the literature. Table~\ref{tab:Status} summarizes the main characteristics of the existing deep solutions, sorted according to their publication dates. This table categorizes the available solutions based on the dimensions associated with the proposed taxonomy. This table also includes the loss functions that the methods have used for training as well as the datasets used for performance assessment. 
% Table~\ref{tab:Status} summarizes the main characteristics of the available solutions, sorted according to their publication dates. We have additionally included information about the venues where these papers have been published and the loss functions that the methods used for training. This Table includes information about the datasets used for performance assessment. 
To better analyze the methods presented in Table~\ref{tab:Status}, we also visualize them in Figure \ref{fig:tax}(a), where each entry represents a method categorized based on the three levels of our taxonomy as well as the publication year and month. We excluded the body representation dimension in this figure for better readability. The information about temporal representation, feature representation, and publication date are shown in the \textit{x}, \textit{y}, and \textit{z} axes, respectively. Lastly, the color and marker symbol of each vertical line represent the neural architecture.

% \begin{figure}[!t]
% \centering
% % \includegraphics[width=1 \columnwidth]{Images/fig8_v2.pdf}
% % \includegraphics[width=1 \columnwidth]{Images/fig8_v3.pdf}
% \includegraphics[width=0.9 \columnwidth]{Images/fig8_v5.pdf}
% \caption{The frequency of different neural architectures, loss functions, and gait datasets used in the literature.}
% \label{fig:freq}
% \end{figure}

\subsection{Analysis and Trends}

Our analysis based on Table~\ref{tab:Status} and Figure \ref{fig:tax}(a) allows us to reach some interesting conclusions about the recent evolution and trends in deep gait recognition technologies with respect to our proposed taxonomy. Following are the key ideas from the analysis.

\textbf{Body Representation.} Silhouettes are the most widely adopted body representation for deep gait recognition, corresponding to over 81\% of the literature. Skeletons have been considered less frequently compared to silhouettes, corresponding to only 13\% of the available solutions. There have also been a few methods, i.e., approximately 5\% of the available literature, that exploit both skeleton and silhouette representations, notably using disentangled representation learning or fusion strategies. Based on our analysis,
% in Section~\ref{sec:comp}, 
high performing gait recognition methods such as~\cite{MM, ECCV,ICPR, TBiom, CVPR20201, ACCV, TIP,CSVT} have all adopted silhouette body representation. Nonetheless, due to recent advancements in effective pose-estimation techniques~\cite{skl2,posecomp, pose, openpose} capable of extracting accurate and robust skeleton data from videos, we anticipate methods based on hybrid silhouettes-skeleton body representations to gain popularity in the near future.

\textbf{Temporal Representation.} Gait templates have been the most considered representation for capturing temporal gait information, corresponding to 70\% of the proposed deep methods. Among different types of gait templates, GEI and set-pooling have been adopted the most. Around 30\% of solutions adopt sequence volumes to preserve the order of available gait frames and to learn from their relationships. Given the frequent use of convolutional templates in some of the recent high-performing literature~\cite{ECCV,ICPR, TBiom, CVPR20201, B1}, we anticipate that these templates gain further popularity and surpass temporal templates in the future.

\textbf{Feature Representation.} Our analysis shows that over 87\% of available methods are based on global feature representations, where deep features are learned by considering gait information as a whole. Recently, many interesting and high performing methods~\cite{B1, TBiom, CVPR20201, ICPR} have adopted partial representations by splitting the gait data into local regions. The performance of such techniques points to promising potential in partial representation learning for discriminating key gait features. Hence, we anticipate further research with convincing results in this area.

\textbf{Neural Architectures.}  As presented in Figure~\ref{fig:tax}(b), 2D CNNs are the most widely used DNN type for deep gait recognition with 48\% of the published solutions utilizing only 2D CNN architectures for classification. 
% Among the single deep neural networks for gait recognition, CNN is the most widely used one that can be applied to a single temporal template (48\% of the methods). 
3D CNNs and GANs are the next popular categories, each corresponding to 8\% of the literature. DAEs, RNNs, CapsNets, DBNs, and GCNs are less considered among DNNs, respectively corresponding to 4\%, 2\%, 2\%, 1\%, and 1\% of the methods. Concerning hybrid methods which constitute 26\% of the published solutions, CNN-RNN combinations are the most widely adapted approach with 16\% share,
% of the methods) to learn the spatio-temporal relationships available in a gait video.
while the combination of DAEs with GANs and RNNs corresponds to 8\% of the methods, followed by RNN-CapsNet methods that make up 2\% of the solutions. We expect that hybrid methods that make use of two or more types of DNN attract more attention in the near future and demonstrate robust performance in the field.

% \textbf{Deep Network.} CNN is the most widely used deep network for gait recognition. As conventional CNNs can only be applied to a single image, these networks cannot solely adopt sequence learning strategies. Instead, as discussed earlier, CNNs can be combined with other networks, e.g., RNNs, in the form of hybrid deep networks that learn the relationships between frames. Gait recognition methods based on RNNs, 3D CNNs, and GCNs always rely on sequence volume temporal representations; this is also usually the case for the hybrid deep networks. There have also been several deep networks, including DAEs, GANs, and CapsNets that have recently been adopted in the context of gait recognition and are expected to attract more attention in the near future. 

\textbf{Loss Functions.}
Loss functions calculate a model's error during training and should ideally be designed to efficiently capture the properties of the problem for facilitating an effective training process~\cite{sundararajan2018deep}. Figure~\ref{fig:tax}(b) shows the usage frequency of different well-known loss functions that have been used by deep gait recognition literature. Among the single loss functions, cross-entropy~\cite{cent} has been the most widely adopted with 20\% of solutions having used it. This loss function takes the output probabilities of the predicated classes
% and measure the distance between the output and the ground-truth values 
and makes the model output as close as possible to the ground-truth output. Triplet loss~\cite{TS} is the next popular type with a usage frequency of 17\%. This loss has been notably used by some of the most recent and state-of-the-art solutions~\cite{TBiom,ICPR, CVPR20201,CVPR20202, B1,MM} and ~\cite{ECCV}. This loss function compares a baseline input, also known as \textit{anchor}, to a \textit{positive} sample with the same identity, and a \textit{negative} sample with a different identity. The loss function then ensures that the dissimilarity between two feature vectors belonging to the same subject is lower than that between feature vectors belonging to two different subjects. Contrastive loss~\cite{contloss} corresponds to 7\% of the recognition methods, and uses pairs of samples including anchor-neighbor or anchor-distant. If the pair of samples is anchor-neighbor, the loss function minimizes their distance; otherwise, it increases their distance. Th next popular loss function, corresponding to 6\% of the recognition methods, is based on the softmax loss~\cite{softloss}. There have also been some other loss functions, such as arcface~\cite{arc}, center loss~\cite{centerloss}, and Euclidean loss~\cite{euc} with a combined usage frequency of 9\% that have been less considered for gait recognition. Finally, there have been two classes of deep gait recognition methods that use multiple loss functions (usage frequency of 41\%), including (\textit{i}) methods such as~\cite{GCN2,ECCV} that add together two or more loss functions to complement each other and compensate their weaknesses; and (\textit{ii}) methods that have been designed based on networks with multiple components, such GANs with generators and discriminators~\cite{SOTA26, B7, GAN3,GAN4} and hybrid networks~\cite{GCN2,B3,ECCV,ACCV}, where different loss functions have been used to train different components. We expect that deep gait recognition methods based on multiple losses attract more attention and surpass other approaches in the near future.

\textbf{Datasets.}
We tally the number of times each dataset has been used by the published literature and present the results in Figure~\ref{fig:tax}(b). Figure~\ref{fig:tax}(b) does not include the datasets that have appeared less than 3 times in Table~\ref{tab:Status}. In addition, a point to consider is that many of the literature use more than one dataset to perform the experiments. We observe that CASIA-B~\cite{CASIA} is the most widely used dataset, appearing in 80\% of the published literature, as it provides a large number of samples with variations in carrying and wearing conditions. OU-ISIR~\cite{DB-OU} was the largest gait dataset prior to 2018; we therefore found OU-ISIR to be the second most popular dataset having been used by 40\% of the solutions. Since the introduction of OU-MVLP~\cite{MVLP} in 2018, this dataset has been receiving considerable attention from the community and has been used by 18\% of the methods in a span of only 2 years. The OU-ISIR LP Bag dataset~\cite{OUBAG} only consists of gait data with carried objects, so naturally it was only considered when designing solutions for specific applications such as those intended to be invariant to carrying conditions from a single viewpoint. As a result, this dataset was used for evaluation purposes by only 5\% of methods. TUM GAID~\cite{DB-TUM} has also been less considered by the community, corresponding to 5\% of published literature. Finally, CASIA-E~\cite{SOTA23, CASIAE} which was developed in 2020 is the sixth most widely used, appearing in 4\% of the literature. However, we anticipate that this dataset will become the standard benchmark dataset for gait recognition in the near future, due to the fact that it provides hundreds of video sequences per each subject with high variations in appearance and acquisition environments.

\subsection{Performance Comparison} \label{sec:comp}
To shed more light on the performance of deep gait recognition methods, we summarize the performance of the methods tested on the three most popular gait datasets, namely  CASIA-B~\cite{CASIA}, OU-ISIR~\cite{DB-OU}, and OU-MVLP~\cite{MVLP} datasets in Tables~\ref{tab:CASIA},~\ref{tab:OU}, and~\ref{tab:OU2} respectively. To perform a fair comparison, these tables only include methods that followed the standard test protocols designed for these datasets, as discussed in Section 3.2. The results show that the method proposed in~\cite{MM} currently provides the best recognition results on CASIA-B (average performance result of 90.4\%) and OU-ISIR (performance result of 99.9\%). Concerning the OU-MVLP dataset, results show the superiority of the method proposed in~\cite{ECCV} (performance result of 89.18\%) over other methods. Apart from~\cite{MM} and~\cite{ECCV}, there are several other methods including those proposed in~\cite{ICPR, TBiom, CVPR20201, ACCV, TIP,CSVT}, whose performances are near the state-of-the-art for these datasets. Our analysis shows that some of these best performing methods, including~\cite{ICPR, TBiom, ACCV, TIP}, make use of two or more types of neural architectures to boost the performance. Some other methods including ~\cite{MM, ECCV, CVPR20201,CSVT} use multiple loss functions to complement each other and compensate their weaknesses to boost performance. This analysis reveals the effectiveness of hybrid approaches, in term of either neural architectures as well as loss functions, for achieving strong performance in the area.

\begin{table}[!t]%[!htbp]
% \footnotesize
\centering
 \setlength\tabcolsep{6pt}
\caption{State-of-the-art results on CASIA-B dataset ~\cite{CASIA}. NM, BG, and CL are respectively normal walking, walking with a bag, and walking with a coat test protocols.}
\begin{tabular}{ l l l | l l l  | l }
\hline
\multicolumn{3}{ c |}{\textbf{Method}} & \multicolumn{4}{ c }{\textbf{Performance}}  \\ 
\hline
	 \textbf{Reference} & \textbf{Year} & \textbf{Venue} &  \textbf{NM} & \textbf{BG} & \textbf{CL} &  \textbf{Average} \\ \hline\hline

	~\cite{TMM} & 2015 & \textit{IEEE T-MM} & 78.9 & --- & --- & --- \\
	~\cite{Biom} & 2017 & \textit{Int. J. Biom.} & 90.8 & 45.9 & 45.3 & 60.7\\
	~\cite{B6} & 2017 & \textit{IEEE T-PAMI} & 94.1 & 72.4 & 54.0 & 73.5 \\ 
	~\cite{DIC} & 2018 & \textit{DIC} & 83.3 & --- & 62.5 & --- \\
	~\cite{SOTA23}   & 2019 & \textit{PR} & 75.0 & --- & --- & --- \\         
	~\cite{B7} & 2019 & \textit{IEEE T-IFS} & 79.8 & --- & --- & ---\\ 
	~\cite{B10} & 2019 & \textit{PR} & 89.9 & --- & --- & --- \\
	~\cite{B3} & 2019 & \textit{CVPR} & 93.9 & 82.6 & 63.2 & 79.9 \\
	~\cite{B2} & 2019 & \textit{CVPR} & 89.9 & --- & --- & --- \\
	~\cite{IETBiom} & 2019 & \textit{IET Biom.} & 94.5  & 78.6  & 51.6 & 74.9\\
	~\cite{PRL} & 2019 & \textit{PRL} & 86.1 & --- & --- & --- \\
	~\cite{B1} & 2019 & \textit{AAAI} & 95.0 & 87.2 & 70.4 & 84.2 \\
% 	~\cite{TIFS} & 2019 & \textit{T-IFS} & & & & \\
	~\cite{B4} & 2020 & \textit{IEEE T-PAMI} & 92.3 & 88.9 & 62.3 & 81.2\\
	~\cite{TIP} & 2020 & \textit{IEEE T-IP} & 96.0 & --- & --- & --- \\
	~\cite{CSVT} & 2020 & \textit{IEEE T-CSVT} & 92.7 & --- & --- & --- \\
	~\cite{SOTA20} & 2020 & \textit{IEEE Access} & 95.1 & 87.9 & 74.0 & 85.7 \\
	~\cite{ICPR} & 2020 & \textit{ICPR} & 95.7 & 90.7 & 72.4 & 86.3\\
	~\cite{TBiom} & 2020 & \textit{IEEE T-Biom.} & 95.2 & 89.7 & 74.7 & 86.5\\
	~\cite{CVPR20201} & 2020 & \textit{CVPR} & 96.2 & 91.5 & 78.7 & 88.8 \\
    ~\cite{CVPR20202} & 2020 & \textit{CVPR} & 94.5 & --- & --- & --- \\
    ~\cite{ECCV}& 2020 & ECCV & 96.8 & \textbf{94.0} & 77.5 & 89.4 \\
% 	~\cite{arXiv} & 2020 & \textit{arXiv} & 96.4 & 92.7 & \textbf{83.0} & \textbf{90.7}\\
	~\cite{ACCV} & 2020 & \textit{ACCV} & \textbf{97.9} & 93.1 & 77.6 & 89.5\\
	~\cite{MM} & 2020 & \textit{MM} & 96.7 & 93.0 & \textbf{81.5} & \textbf{90.4} \\
\hline
\end{tabular}
\label{tab:CASIA}
\end{table}

% % \subsection{State-Of-The-Art on OU-ISIR and OU-MVLP Datasets}
% \textbf{OU-ISIR and OU-MVLP Datasets.}
% As shown in Figure \ref{fig:datasetN}, OU-ISIR~\cite{DB-OU} and OU-MVLP~\cite{MVLP} are two other datasets that have been widely used by deep recognition methods. The results obtained on these two datasets are presented in Table \ref{tab:OU}. This Table only includes the methods that followed the respective test protocol for cross-view gait recognition, as discussed in Section 3. 
% % The results show the OU-MVLP dataset~\cite{MVLP} has recently received more attentions after its introduction in 2018, when compared to OU-ISIR~\cite{DB-OU}. This is due to the fact that OU-MVLP provides deep methods with a very large number of samples (more than 250K gait video) captured from 14 viewpoints that facilitates an efficient training and a reliable evaluation. Similar to the results for CASIA-B~\cite{CASIA}, 
% Table \ref{tab:OU} shows the superiority of 
% % ~\cite{B1, ICPR, TBiom, CVPR20201, ECCV, ACCV, TIP, MM}, over other gait recognition methods. Among these methods, 
% \cite{MM} and \cite{ECCV}, respectively for OU-ISIR~\cite{DB-OU} and OU-MVLP~\cite{MVLP} datastes, over other gait recognition methods.

\begin{table}[!t]%[!htbp]
% \footnotesize
\centering
 \setlength\tabcolsep{8pt}
\caption{State-of-the-art results on OU-ISIR~\cite{DB-OU} dataset.}

\begin{tabular}{ l l l | l  }
\hline
\multicolumn{3}{ c |}{\textbf{Method}} & \multicolumn{1}{ c }{}  \\ 
% \hline
	 \textbf{Reference} & \textbf{Year} & \textbf{Venue} &   \textbf{Performance}  \\ \hline\hline

    \cite{SOTA3}      & 2016  & \textit{ICASSP}         &90.71            \\  
    \cite{B6}         & 2017  & \textit{IEEE T-PAMI}   &92.77           \\  
    \cite{SOTA5}      & 2017  & \textit{Applied Sci.}  &91.25            \\
    \cite{SOTA21}     & 2018  & \textit{Neuroinform.}  &88.56            \\
    \cite{SOTA8}      & 2018  & \textit{IEEE Access}   &95.67            \\
    \cite{IETBiom}    & 2019  & \textit{IET Biom.}     &97.40            \\
    \cite{B7}         & 2019  & \textit{IEEE T-IFS }   &93.20            \\
    \cite{SOTA24}     & 2019  & \textit{PRL}           &94.62            \\
    \cite{SOTA25}     & 2019  & \textit{IEEE T-MM}     &97.26            \\
    \cite{SOTA19}     & 2020  & \textit{IJCB}          &94.17            \\
    \cite{CSVT}       & 2020  & \textit{IEEE T-CSVT}   &98.93            \\
    \cite{TIP}        & 2020  & \textit{IEEE T-IP}     &99.27            \\
    \cite{MM}         & 2020  & \textit{MM}            &\textbf{99.90}   \\
\hline
\end{tabular}
\label{tab:OU}
\end{table}

\begin{table}[!t]%[!htbp]
% \footnotesize
\centering
 \setlength\tabcolsep{8pt}
\caption{State-of-the-art results on OU-MVLP~\cite{MVLP} dataset.}

\begin{tabular}{ l l l | l  }
\hline
\multicolumn{3}{ c |}{\textbf{Method}} & \multicolumn{1}{ c }{}  \\ 
% \hline
	 \textbf{Reference} & \textbf{Year} & \textbf{Venue} & \textbf{Performance}    \\ \hline\hline

    \cite{B1}         & 2019  & \textit{AAAI}          &  83.40   \\
    \cite{ICASSP3}    & 2020  & \textit{ICASSP}        &  57.80   \\
    \cite{CSVT}       & 2020  & \textit{IEEE T-CSVT}   &  63.10   \\
    \cite{TIP}        & 2020  & \textit{IEEE T-IP}     &  84.60   \\
    \cite{ICPR}       & 2020  & \textit{ICPR}          &  84.50   \\
    \cite{TBiom}      & 2020  & \textit{IEEE T-Biom.}  &  84.30   \\
    \cite{CVPR20201}  & 2020  & \textit{CVPR}          &  88.70   \\
    \cite{ECCV}       & 2020  & \textit{ECCV}          &  \textbf{89.18}   \\
\hline
\end{tabular}
\label{tab:OU2}
\end{table}

{\section{Vulnerability to Adversarial Attacks}}

{Traditional spoofing attacks to gait recognition systems were attempted by trained impostors, notably with similar body shapes and clothes, imitating the walking styles of target subjects~\cite{taa1, taa2, taa3}. However, these attacks are rather hard and limited as they require trained and qualified imitators. Different from the traditional spoofing attacks, \textit{adversarial attacks} to gait recognition systems have been designed to imperceptibly fool recognition systems by synthesizing input gait videos with both good quantitative similarity and visual realism.}

{Despite the strong performance of deep learning solutions in computer vision, such solutions have been surprisingly vulnerable to adversarial attacks~\cite{aas1, aas2}. These attacks introduce perturbations in visual content that can manipulate the predictions of deep models by resulting in embeddings capable of fooling the classifiers~\cite{aas3}. 
Since the introduction of the first adversarial attacks to deep neural networks in 2014~\cite{aafirst}, these models have attracted significant attention from the research community in computer vision and pattern analysis. Among the adversarial attack approaches, GANs~\cite{goodfellow, aagan1, aagan2, aagan3} have been one of the most powerful methods for image and video manipulation.}

{The first attempt to investigates the vulnerability of gait recognition systems to adversarial attacks was done in~\cite{aa1} using a GAN for synthesizing gait images. In this context, the foreground of each frame of the source video is first segmented to be then fed to the generator along with the target background. The generator is composed of two parallel encoder-decoder networks, respectively dealing with foreground and background information. The corresponding feature representations from these two networks are fused at multiple scales. Static and dynamic silhouette-based losses have been designed in order to force the model to generate more realistic results for gait recognition. Additionally, triplet loss is used to preserve the similarity between the individuals in the source and generated videos. The performance of two state-of-the-art gait recognition systems, including CNNGait~\cite{B6} and GaitSet~\cite{B1}, were evaluated using the generated gait samples. The results showed that the generated samples provide sufficient discriminative information to bypass gait recognition systems. To show how the sequence-based gait recognition is vulnerable to adversarial attacks, another study was done in~\cite{aa2}. In this study, a novel temporal sparse adversarial attack method was proposed based on a GAN to synthesize high-quality sequences of silhouette frames. To ensure imperceptibility of the proposed method, a few adversarial gait silhouettes are substituted or inserted in the sequence. Experimental results show that the state-of-the-art GaitSet~\cite{B1} method has low robustness to this adversarial attack.}

{The presented results in~\cite{aa1, aa2} suggest that adversarial attacks can surprisingly degrade the performance of deep gait recognition methods, thus posing a real threat to such recognition systems. This demonstrates the necessity of adopting efficient countermeasure techniques against adversarial attacks aimed towards deep gait recognition systems.}

\section{Challenges and Future Research Directions}

Despite the great success of gait recognition using deep learning techniques, there still remains a large number of challenges that need to be addressed in the area. Here, we further point out a few promising future research directions and open problems in this domain. These directions may facilitate future research activities and foster real-world applications.

\subsection{Disentanglement}

Complex gait data arise from the interaction between many factors such as occlusion, camera view-points, appearance of individuals, sequence order, body part motion, or lighting sources present in the data~\cite{B3, B4, CVPR20202}. These factors can interact in complex manners that can complicate the recognition task. There have recently been a growing number of methods in other research areas, such as face recognition~\cite{distface1, distface2}, action recognition~\cite{distact}, emotion recognition~\cite{dist}, and pose estimation~\cite{distpos}, that focus on learning disentangled features by extracting representations that separate the various explanatory factors in the high-dimensional space of the data~\cite{dist1, dist2, dist3}. However, the majority of available deep gait recognition methods have not explored disentanglement approaches, and hence are not explicitly able to separate the underlying structure of gait data in the form of meaningful disjoint variables. Despite the recent progress in using disentanglement approaches in a few gait recognition methods~\cite{B3, B4, CVPR20202}, there is still room for improvement. To foster further progress in this area, the adaptation of novel generative models~\cite{dist5, dist6} and loss functions~\cite{dist7} can be considered to learn more discriminative gait representations by explicitly disentangling identity and non-identity components.

\subsection{Self-supervised Learning}

A considerable majority of available deep gait recognition methods follow the supervised learning paradigm, and thus require labeled data during training. Nevertheless, in real-world applications, labeled data may not always be readily available and labels are generally expensive and time-consuming to obtain. In order to utilize unlabeled gait data to learn more efficient and generalizable gait representations, self-supervised learning~\cite{self} can be exploited. In this context, general and rich high-level semantics can be captured without using any annotated labels. Self-supervised approaches can define various pretext tasks, such as body part motion or sequence order recognition for input sequences~\cite{self1,self2}, to be solved by a network. Through learning these pretext tasks, the network can then learn generic features. The network trained with the generated pre-text labels can then be fine-tuned with the actual labels in order to recognize the identity. Among self-supervised approaches, contrastive learning methods~\cite{ssl1}, including SimCLR~\cite{ssl2}, are promising approaches that learn representations by defining an \textit{anchor} and a \textit{positive} sample in the feature space, and then aim to make the anchor separable from the \textit{negative} samples. One important challenge in using self-supervised learning in the context of gait recognition is to design effective pretext tasks to ensure the network can learn meaningful representations.
% while also maintaining a balance between the convergence and overfitting. 
Additionally, 
% most existing self-supervised methods learn representations by training a network to solve a single pretext task. However, 
joint learning of several pretext tasks in a network~\cite{self}, instead of a single pretext task, notably using several loss functions~\cite{self3}, can provide the network with more representative features~\cite{pritam1, pritam2, pritamR}.
% different pretext tasks may provide different supervision signals for gait recognition, so that their combination can provide the network with more representative features. It also brings to attention the importance of the task selection in a way to narrow the domain gap between self-supervised trained model and the downstream task, i.e. gait recognition. 
We expect these challenges to gain increased popularity in the context of deep gait recognition in the near future.

% We expect that contrastive self-supervised learning methods
% Self-supervised visual representation learning with deep learning has recently been used for person re-identification~\cite{self1,self2} and is expected to 
% will be adopted for gait recognition in future research.  

\subsection{Multi-task Learning}

Multi-task learning is generally performed to simultaneously learn multiple tasks using a shared model, thus learning more generalized and often reinforced representations~\cite{multitask1, multitask2, MLT6}. In many cases, these approaches offer advantages such as increased convergence speed, improved learning by leveraging auxiliary information, and reduced overfitting through shared representations~\cite{multitask1, multitask2}.
Despite the effectiveness of multi-task learning in a number of other domains~\cite{MTL1, MTL2}, most deep gait recognition solutions in the literature focus on the single task of identification. Thus, most existing works learn features that are sensitive to identity without considering interactions with other latent factors, such as affective states, gender, and age~\cite{B7, SOTA1,MTLG}. 
% However, in the real world, gait-based identification can be intertwined with the auxiliary tasks for the improved recognition. 
% To improve the performance of gait recognition methods, multi-task learning strategies~\cite{multitask1, multitask2} can be used, thus simultaneously learning different factors by a shared model that can be used for gait recognition
% . This reduces overfitting through shared representations, increases the learning speed by leveraging auxiliary information, and disentangles nuisance factors.
In this context, simultaneous learning of multiple tasks for gait recognition may present new design paradigms and optimization challenges, notably in terms of task identification and loss functions~\cite{MLTLoss}. 
% Future gait recognition methods based on multi-task learning are expected to identify tasks that are complementary to recognition, while also ignoring the noisy and outlier tasks. 
We expect these challenges to attract further attention in the near future and be tackled in the context of gait recognition with multi-task learning.

% that can be more widely studied for future research in the field.

% \subsection{Few-shot and zero-shot}

% \subsection{Contrastive Learning}

\subsection{Data Synthesis and Domain Adaptation}

Deep gait recognition methods require large amounts of data for effective training and reliable evaluation. This issue is evident in Figure~\ref{fig:tax}(b) where most of the deep gait recognition solutions~\cite{ICPR, TBiom, CVPR20201, ACCV, TIP,CSVT, MM, ECCV} used large-scale gait datasets for instance CASIA-B~\cite{CASIA}, OU-ISIR~\cite{DB-OU}, and OU-MVLP~\cite{MVLP}. In the context of deep gait recognition, data synthesis, for instance using GANs~\cite{stylegan1, stylegan2}, can be considered for creating large datasets or data augmentation~\cite{frameGAN, MVG,Ali2}. Furthermore, developing synthesized datasets can also be advantageous in that subject privacy concerns could be alleviated with fake subject data. Similar approaches have been carried out for the more \textit{privacy-sensitive} area of facial recognition~\cite{faceDB1, faceDB2}, where large datasets comprised only of fake data have been developed to be used in deep learning research~\cite{100kface, faceDB3, faceDB4}. In addition, such approaches can be used to increase the variance of existing datasets. For instance, large-scale gait datasets such as OU-ISIR~\cite{DB-OU} and OU-MVLP~\cite{MVLP} only provide normal walking sequences with no variations in occlusion or carrying and clothing conditions. Thus, solutions trained on these datasets usually fail to generalize well when facing variations in appearance and environment during the testing phase. Here, domain adaptation~\cite{domainadapt3, domainadapt4,domainadapt, domainadapt2, cycleGAN} is a potential remedy for this problem that can modify existing datasets to include the desired variations, thus eliminating the necessity for collecting new data. 
% In this context, GAN architectures~\cite{domainadapt, domainadapt2, cycleGAN} can transfer the data across different domains, e.g., from normal walking to walking with a bag. 
Furthermore, gait synthesis can be performed for computer animation~\cite{anim} and by game engines~\cite{GANDB}
% , and radial basis function neural networks~\cite{Ali2}
to generate large-scale synthetic gait datasets. Hence, we anticipate that with advances in gait data synthesis and domain adaptation techniques, more complementary gait datasets will be constructed to enable the development of more robust solutions.

% , as very recently demonstrated in VersatileGait dataset~\cite{GANDB}. 

\subsection{Cross-Dataset Evaluation}
The practical value of gait recognition systems is strongly dependant in its ability to generalize to unseen data. To the best of our knowledge, cross-dataset gait recognition on well-known datasets such as CASIA-B~\cite{CASIA}, OU-ISIR dataset~\cite{DB-OU}, and OU-MVLP~\cite{MVLP}, has not been performed in the literature as notable solutions available in the literature all use the same gait dataset for both training and testing. However, in many real applications such as deployed products, test or run-time data are often obtained in a variety of different conditions with respect to the training data. In order to examine the generalizability of gait recognition systems in real-world applications, cross-dataset evaluations should be adopted, for example using transfer learning techniques~\cite{transfer}. In this context, a solution trained on one dataset can be used to extract features from the test data (gallery and probe sets) of another dataset. The extracted features can then feed a classifier to perform gait recognition. Cross-dataset gait recognition can potentially be formulated as an out-of-distribution (OOD) testing problem, where the generalization ability of a deep model beyond the biases of the training set is evaluated~\cite{OOD2}. We expect that OOD tests~\cite{OOD} become increasingly popular for evaluating the generalization ability of gait recognition methods.
% This type of experiment will reveal the generalization ability of the gait recognition systems and thus foster their real-world applications.

\subsection{Multi-View Recognition}

A large number of gait datasets contain multi-view sequences, providing gait information captured from different view-points. Most of the current methods available in the literature only perform single-view gait recognition. These methods generally learn intra-view relationships and ignore inter-view information between multiple viewpoints.
% available among the different sequences simultaneously captured using separate cameras. 
By casting the problem as multi-view, descriptors such as gate-level fusion LSTM~\cite{joint}, state-level fusion LSTM~\cite{joint}, spatio-temporal LSTM~\cite{LSTMV1}, multi-perspective LSTM~\cite{mplstm}, and multi-view LSTM~\cite{MVLSTM}, can be adopted to jointly learn both the intra-view and inter-view relationships. Another challenge in multi-view gait recognition is that most existing multi-view descriptors consider a well defined camera network topology with fixed camera positions. However, data collection in real-world environments is often uncontrollable, i.e. data might be captured from unpredictable viewing angles~\cite{non-stationary, liem2014joint} or even from moving cameras~\cite{moving}. To this end, existing multi-view methods, which mostly rely on pre-trained descriptors, fail to bridge the domain gap between the training and run-time multi-view data. We expect that future research direction in this area will be shaped by proposing novel approaches, for example using clustering algorithms~\cite{cluster}, combinatorial optimization~\cite{comb}, and self-supervised learning~\cite{PAMINew}, for adopting generic gait descriptors for multi-view geometry. 
% Additionally, as discussed earlier, self-supervised approaches can be considered ...

\subsection{Multi-biometric Recognition}

Some literature in the field have fused gait information with other biometric information such as face~\cite{deepsurvey, face} and ear~\cite{ear1,ear2,ear3}, which can be obtained from high-quality gait videos. As we discussed earlier, gait recognition systems are generally challenged when facing variations in subject appearance and clothing, camera view-points, and body occlusions. On the other hand, additional sources of biometric information, notably face and ear, are less sensitive to some of these challenging factors. Instead, face and ear recognition systems can be negatively affected by some other factors such as low image quality, for instance blurred or low resolution images, varying lighting, or facial occlusions, which in turn have limited impact on the performance of gait recognition systems. Hence, various biometric modalities and gait can complement one another to compensate each others' weaknesses in the context of a multi-biometric system~\cite{fusebio, fusebio2}. 
% For instance, gait recognition systems are less sensitive to factors that affect face recognition, including low image quality and lighting changes, while face recognition systems are known to be more robust to covariates that negatively affect gait recognition systems, including carrying and clothing conditions. 
Apart from the complementary (hard-)biometric traits, soft-biometric traits such as age~\cite{age}, height~\cite{height, height2}, weight~\cite{weight}, gender~\cite{gender}, and particular body marks including tattoos~\cite{tattoo} can also be included to boost overall performance. The combination of other soft- and hard- biometric traits with gait has mostly been done in the literature based on non-deep methods~\cite{multi1, multi2, multi3, multi4, multi5, multi6}, while multi-modal deep learning methods~\cite{fusedeep1, fusedeep2}, notably based on fusion~\cite{fusefuse}, joint learning~\cite{joint}, and attention~\cite{fuseatt} networks, can also be adopted. Hence, we anticipate that research on deep multi-biometric recognition systems that include gait, gain popularity in the coming years.

% \subsection{Using Shadow Information}
% Lastly, as an additional modality, the usage of shadow information has recently been proven to be advantageous, notably for in-the-wild scenarios~\cite{TanmayThesis, R7}. This is due to the fact that they are less sensitive to view-point changes and self-occlusions. Shadow information can be treated as a singular but often weak modality, or alternatively as an additional sources of information to be combined with body information in the context of multimodal gait recognition. 
% It is expected that deep gait recognition using shadow information will receive more attentions in the future. 

\section{Summary}

We provided a survey of deep gait recognition methods that was driven by a novel taxonomy with four dimensions namely body representation, temporal representation, feature representation, and neural architectures. Following our taxonomy, we reviewed the most representative deep gait recognition methods and provided discussions on their characteristics, advantages, and limitations. We additionally reviewed the most commonly used datasets along with their evaluation protocols and corresponding performance results reported in the literature. We finally concluded this survey with a discussion on current challenges, pointing out a few promising future research directions in this domain. We expect that this survey provides insights into the technological landscape of gait recognition for guiding researchers in advancing future research.

\bibliographystyle{ieeetran}
\bibliography{references}

\begin{thebibliography}{100}
\providecommand{\url}[1]{#1}
\csname url@samestyle\endcsname
\providecommand{\newblock}{\relax}
\providecommand{\bibinfo}[2]{#2}
\providecommand{\BIBentrySTDinterwordspacing}{\spaceskip=0pt\relax}
\providecommand{\BIBentryALTinterwordstretchfactor}{4}
\providecommand{\BIBentryALTinterwordspacing}{\spaceskip=\fontdimen2\font plus
\BIBentryALTinterwordstretchfactor\fontdimen3\font minus
  \fontdimen4\font\relax}
\providecommand{\BIBforeignlanguage}[2]{{%
\expandafter\ifx\csname l@#1\endcsname\relax
\typeout{** WARNING: IEEEtran.bst: No hyphenation pattern has been}%
\typeout{** loaded for the language `#1'. Using the pattern for}%
\typeout{** the default language instead.}%
\else
\language=\csname l@#1\endcsname
\fi
#2}}
\providecommand{\BIBdecl}{\relax}
\BIBdecl

\bibitem{R1}
F.~Deligianni, Y.~Guo, and G.~Yang, ``From emotions to mood disorders: A survey
  on gait analysis methodology,'' \emph{IEEE Journal of Biomedical and Health
  Informatics}, vol.~23, no.~6, pp. 2302--2316, November 2019.

\bibitem{Ali1}
A.~{Etemad} and A.~{Arya}, ``Expert-driven perceptual features for modeling
  style and affect in human motion,'' \emph{IEEE Transactions on Human-Machine
  Systems}, vol.~46, no.~4, pp. 534--545, August 2016.

\bibitem{Ali2}
A.~Etemad and A.~Arya, ``Classification and translation of style and affect in
  human motion using {RBF} neural networks,'' \emph{Neurocomputing}, vol. 129,
  no.~1, pp. 585 -- 595, April 2014.

\bibitem{Ali3}
A.~Etemad and A.Arya, ``Correlation-optimized time warping for motion,''
  \emph{The Visual Computer volume}, vol.~31, no.~1, p. 1569–1586, October
  2015.

\bibitem{sport1}
J.~M. Echterhoff, J.~Haladjian, and B.~Br\"{u}gge, ``Gait and jump
  classification in modern equestrian sports,'' in \emph{ACM International
  Symposium on Wearable Computers}, New York, NY, USA, October 2018.

\bibitem{sport2}
H.~{Zhang}, Y.~{Guo}, and D.~{Zanotto}, ``Accurate ambulatory gait analysis in
  walking and running using machine learning models,'' \emph{IEEE Transactions
  on Neural Systems and Rehabilitation Engineering}, vol.~28, no.~1, pp.
  191--202, January 2020.

\bibitem{pato}
T.~T. {Verlekar}, P.~{Lobato Correia}, and L.~D. {Soares}, ``Using transfer
  learning for classification of gait pathologies,'' in \emph{International
  Conference on Bioinformatics and Biomedicine}, Madrid, Spain, January 2019.

\bibitem{clinic}
A.~Muro-de-la Herran, B.~Garcia-Zapirain, and A.~Mendez-Zorrilla, ``Gait
  analysis methods: An overview of wearable and non-wearable systems,
  highlighting clinical applications,'' \emph{Sensors}, vol.~14, no.~2, pp.
  3362--3394, February 2014.

\bibitem{clinic2}
D.~{Jarchi}, J.~{Pope}, T.~K.~M. {Lee}, L.~{Tamjidi}, A.~{Mirzaei}, and
  S.~{Sanei}, ``A review on accelerometry-based gait analysis and emerging
  clinical applications,'' \emph{IEEE Reviews in Biomedical Engineering},
  vol.~11, pp. 177 -- 194, February 2018.

\bibitem{survey1}
C.~Wan, L.~Wang, and V.~V. Phoha, ``A survey on gait recognition,'' \emph{ACM
  Computing Surveys}, vol.~51, no.~5, pp. 1--35, August 2018.

\bibitem{survey2}
I.~Rida, N.~Almaadeed, and S.~Almaadeed, ``Robust gait recognition: a
  comprehensive survey,'' \emph{IET Biometrics}, vol.~8, no.~1, pp. 14--28,
  January 2019.

\bibitem{nambiar}
A.~Nambiar, A.~Bernardino, and J.~C. Nascimento, ``Gait-based person
  re-identification: A survey,'' \emph{ACM Computing Surveys}, vol.~52, no.~2,
  pp. 1--34, April 2019.

\bibitem{wear}
M.~D. Marsico and A.~Mecca, ``A survey on gait recognition via wearable
  sensors,'' \emph{ACM Computing Surveys}, vol.~52, no.~4, pp. 1--39, September
  2019.

\bibitem{survey4}
J.~P. Singh, S.~Jain, S.~Arora, and U.~P. Singh, ``Vision-based gait
  recognition: A survey,'' \emph{IEEE Access}, vol.~6, pp. 70\,497--70\,527,
  November 2018.

\bibitem{incomp}
C.~Chen, J.~Liang, H.~Zhao, H.~Hu, and J.~Tian, ``Frame difference energy image
  for gait recognition with incomplete silhouettes,'' \emph{Pattern Recognition
  Letters}, vol.~30, no.~11, pp. 977 -- 984, August 2009.

\bibitem{bar3}
M.~Z. Uddin, D.~Muramatsu, N.~Takemura, M.~A.~R. Ahad, and Y.~Yagi,
  ``Spatio-temporal silhouette sequence reconstruction for gait recognition
  against occlusion,'' \emph{{IPSJ} Transactions on Computer Vision and
  Applications}, vol.~11, no.~1, p.~9, November 2019.

\bibitem{background}
A.~Ibrahim, W.-N. Mohd-Isa, and C.~C. Ho, ``Background subtraction on gait
  videos containing illumination variates,'' in \emph{AIP Conference
  Proceedings}, vol. 2016, no.~1, 2018.

\bibitem{R7}
T.~Verlekar, L.~Soares, and P.~Correia, ``Gait recognition in the wild using
  shadow silhouettes,'' \emph{Image and Vision Computing}, vol.~76, no.~1, pp.
  1 -- 13, August 2018.

\bibitem{face}
A.~Sepas-Moghaddam, F.~Pereira, and P.~Correia, ``Face recognition: A novel
  multi-level taxonomy based survey,'' \emph{IET Biometrics}, vol.~9, no.~2,
  pp. 1--12, March 2020.

\bibitem{ear1}
{\v{Z}}.~Emer{\v{s}}i{\v{c}}, V.~{\v{S}}truc, and P.~Peer, ``Ear recognition:
  More than a survey,'' \emph{Neurocomputing}, vol. 255, pp. 26--39, September
  2017.

\bibitem{iris}
K.~Nguyen, C.~Fookes, R.~Jillela, S.~Sridharan, and A.~Ross, ``Long range iris
  recognition: A survey,'' \emph{Pattern Recognition}, vol.~72, no.~1, pp.
  123--143, December 2017.

\bibitem{iris2}
------, ``Long range iris recognition: A survey,'' \emph{Pattern Recognition},
  vol.~72, pp. 123--143, December 2017.

\bibitem{finger}
K.~Cao and A.~K. Jain, ``Automated latent fingerprint recognition,'' \emph{IEEE
  transactions on pattern analysis and machine intelligence}, vol.~41, no.~4,
  pp. 788--800, March 2019.

\bibitem{taa1}
A.~Hadid, M.~Ghahramani, V.~Kellokumpu, M.~Pietikäinen, J.~Bustard, and
  M.~Nixon, ``Can gait biometrics be spoofed?'' in \emph{International
  Conference on Pattern Recognition}, Tsukuba, Japan, November 2012.

\bibitem{reid_survey}
M.~Ye, J.~Shen, G.~Lin, T.~Xiang, L.~Shao, and S.~C. Hoi, ``Deep learning for
  person re-identification: A survey and outlook,'' \emph{IEEE Transactions on
  Pattern Analysis and Machine Intelligence}, vol. in press, 2021.

\bibitem{activity}
K.~Chen, D.~Zhang, L.~Yao, B.~Guo, Z.~Yu, and Y.~Liu, ``Deep learning for
  sensor-based human activity recognition: Overview, challenges, and
  opportunities,'' \emph{ACM Computing Surveys (CSUR)}, vol.~54, no.~4, pp.
  1--40, July 2021.

\bibitem{reid_feat}
C.~Liu, S.~Gong, C.~C. Loy, and X.~Lin, ``Person re-identification: What
  features are important?'' in \emph{European Conference on Computer Vision},
  Florence, Italy, October 2012.

\bibitem{wang2019deep}
J.~Wang, Y.~Chen, S.~Hao, X.~Peng, and L.~Hu, ``Deep learning for sensor-based
  activity recognition: A survey,'' \emph{Pattern Recognition Letters}, vol.
  119, pp. 3--11, March 2019.

\bibitem{fatigue}
J.~L. Helbostad, S.~Leirfall, R.~Moe-Nilssen, and O.~Sletvold, ``Physical
  fatigue affects gait characteristics in older persons,'' \emph{The Journals
  of Gerontology Series A: Biological Sciences and Medical Sciences}, vol.~62,
  no.~9, pp. 1010--1015, September 2007.

\bibitem{gaitaffect}
M.~Karg, K.~K{\"u}hnlenz, and M.~Buss, ``Recognition of affect based on gait
  patterns,'' \emph{IEEE Transactions on Systems, Man, and Cybernetics, Part B
  (Cybernetics)}, vol.~40, no.~4, pp. 1050--1061, August 2010.

\bibitem{Injury}
I.~Kl{\"o}pfer-Kr{\"a}mer, A.~Brand, H.~Wackerle, J.~M{\"u}{\ss}ig,
  I.~Kr{\"o}ger, and P.~Augat, ``Gait analysis--available platforms for outcome
  assessment,'' \emph{Injury}, vol.~51, pp. S90--S96, May 2020.

\bibitem{CASIA}
S.~Yu, D.~Tan, and T.~Tan, ``A framework for evaluating the effect of view
  angle, clothing and carrying condition on gait recognition,'' in
  \emph{International Conference on Pattern Recognition}, Hong Kong, China,
  August 2006.

\bibitem{R3}
D.~Cunado, M.~Nixon, and J.~Carter, ``Using gait as a biometric, via
  phase-weighted magnitude spectra,'' in \emph{International Conference on
  Audio- and Video-Based Biometric Person Authentication}, Crans-Montana,
  Switzerland, March 1997.

\bibitem{firstNN}
J.~{Yoo}, D.~{Hwang}, K.~{Moon}, and M.~S. {Nixon}, ``Automated human
  recognition by gait using neural network,'' in \emph{Workshops on Image
  Processing Theory, Tools and Applications}, Sousse, Tunisia, November 2008.

\bibitem{SOTA1}
C.~{Yan}, B.~{Zhang}, and F.~{Coenen}, ``Multi-attributes gait identification
  by convolutional neural networks,'' in \emph{International Congress on Image
  and Signal Processing}, Shenyang, China, October 2015.

\bibitem{SOTA2}
Y.~Feng, Y.~Li, and J.~Luo, ``Learning effective gait features using {LSTM},''
  in \emph{International Conference on Pattern Recognition}, Cancun, Mexico,
  December 2016.

\bibitem{B8}
K.~{Shiraga}, Y.~{Makihara}, D.~{Muramatsu}, T.~{Echigo}, and Y.~{Yagi},
  ``{GEINet}: View-invariant gait recognition using a convolutional neural
  network,'' in \emph{International Conference on Biometrics}, Halmstad,
  Sweden, June 2016.

\bibitem{DBN2}
B.~M. Nair and K.~D. Kendricks, ``Deep network for analyzing gait patterns in
  low resolution video towards threat identification.'' \emph{Electronic
  Imaging}, vol. 2016, no.~11, pp. 1--8, February 2016.

\bibitem{B6}
Z.~{Wu}, Y.~{Huang}, L.~{Wang}, X.~{Wang}, and T.~{Tan}, ``A comprehensive
  study on cross-view gait based human identification with deep {CNNs},''
  \emph{IEEE Transactions on Pattern Analysis and Machine Intelligence},
  vol.~39, no.~2, February 2017.

\bibitem{DIC}
L.~{Yao}, W.~{Kusakunniran}, Q.~{Wu}, J.~{Zhang}, and Z.~{Tang}, ``Robust
  cnn-based gait verification and identification using skeleton gait energy
  image,'' in \emph{Digital Image Computing: Techniques and Applications},
  Canberra, Australia, December 2018.

\bibitem{IETBiom}
A.~{Sokolova} and A.~{Konushin}, ``Pose-based deep gait recognition,''
  \emph{IET Biometrics}, vol.~8, no.~2, pp. 134--143, February 2019.

\bibitem{B3}
Z.~Zhang, L.~Tran, X.~Yin, Y.~Atoum, X.~Liu, J.~Wan, and N.~Wang, ``Gait
  recognition via disentangled representation learning,'' in \emph{Computer
  Vision and Pattern Recognition}, Long Beach, CA, USA, June 2019.

\bibitem{B1}
H.~Chao, Y.~He, J.~Zhang, and J.~Feng, ``Gaitset: Regarding gait as a set for
  cross-view gait recognition,'' in \emph{AAAI Conference on Artificial
  Intelligence}, Honolulu, HW, USA, February 2019.

\bibitem{TBiom}
A.~{Sepas-Moghaddam} and A.~{Etemad}, ``View-invariant gait recognition with
  attentive recurrent learning of partial representations,'' \emph{IEEE
  Transactions on Biometrics, Behavior, and Identity Science}, vol.~3, no.~1,
  pp. 124--137, January 2021.

\bibitem{CVPR20201}
C.~Fan, Y.~Peng, C.~Cao, X.~Liu, S.~Hou, J.~Chi, Y.~Huang, Q.~Li, and Z.~He,
  ``Gaitpart: Temporal part-based model for gait recognition,'' in
  \emph{Computer Vision and Pattern Recognition}, Seattle, WA, USA, June 2020.

\bibitem{ECCV}
S.~Hou, C.~Cao, X.~Liu, and Y.~Huang, ``Gait lateral network: Learning
  discriminative and compact representations for gait recognition,'' in
  \emph{European Conference on Computer Vision}, Glasgow, UK, August 2020.

\bibitem{ACCV}
X.~Li, Y.~Makihara, C.~Xu, Y.~Yagi, S.~Yu, and M.~Ren, ``End-to-end model-based
  gait recognition,'' in \emph{Asian Conference on Computer Vision}, Kyoto,
  Japan, November 2020.

\bibitem{MM}
B.~Lin, S.~Zhang, and F.~Bao, ``Gait recognition with multiple-temporal-scale
  {3D} convolutional neural network,'' in \emph{ACM International Conference on
  Multimedia}, Seattle, WA, USA, October 2020.

\bibitem{survey3}
P.~Connor and A.~Ross, ``Biometric recognition by gait: A survey of modalities
  and features,'' \emph{Computer Vision and Image Understanding}, vol. 167, pp.
  1--27, February 2018.

\bibitem{skl3}
M.~J. Nordin and A.~Saadoon, ``A survey of gait recognition based on skeleton
  model for human identification,'' \emph{Research Journal of Applied Sciences,
  Engineering and Technology}, vol.~12, no.~7, pp. 756--763, April 2016.

\bibitem{tax1}
J.~P. Singh, S.~Jain, S.~Arora, and U.~P. Singh, ``A survey of behavioral
  biometric gait recognition: Current success and future perspectives,''
  \emph{Archives of Computational Methods in Engineering}, vol.~28, no.~1, pp.
  107--148, November 2019.

\bibitem{google}
\BIBentryALTinterwordspacing
Google scholar. [Online]. Available: \url{http://scholar.google.com/}
\BIBentrySTDinterwordspacing

\bibitem{IEEE}
\BIBentryALTinterwordspacing
{IEEE Xplore} digital library. [Online]. Available:
  \url{https://ieeexplore.ieee.org/}
\BIBentrySTDinterwordspacing

\bibitem{ACM}
\BIBentryALTinterwordspacing
{ACM} digital library. [Online]. Available: \url{https://www.dl.acm.org/}
\BIBentrySTDinterwordspacing

\bibitem{ELS}
\BIBentryALTinterwordspacing
{ScienceDirect} digital library. [Online]. Available:
  \url{https://www.sciencedirect.com/}
\BIBentrySTDinterwordspacing

\bibitem{CVF}
\BIBentryALTinterwordspacing
Computer vision foundation ({CVF}) open access. [Online]. Available:
  \url{https://openaccess.thecvf.com/}
\BIBentrySTDinterwordspacing

\bibitem{DB-MOBO}
R.~Gross and J.~Shi, ``The {CMU} motion of body ({MoBo}) database,''
  CMU-RI-TR-01-18, Pittsburgh, PA, USA, Tech. Rep., June 2001.

\bibitem{DB-SOTON}
J.~D. Shutler, M.~G. Grant, M.~S. Nixon, and J.~N. Carter, ``On a large
  sequence-based human gait database,'' in \emph{Applications and Science in
  Soft Computing}.\hskip 1em plus 0.5em minus 0.4em\relax Springer, 2004, pp.
  339--346.

\bibitem{DB-CASIAA}
L.~Wang, T.~Tan, H.~Ning, and W.~Hu, ``Silhouette analysis-based gait
  recognition for human identification,'' \emph{IEEE Transactions on Pattern
  Analysis and Machine Intelligence}, vol.~25, no.~12, pp. 1505--1518, December
  2003.

\bibitem{DB-HumanID}
S.~Sarkar, P.~J. Phillips, Z.~Liu, I.~R. Vega, P.~Grother, and K.~W. Bowyer,
  ``The {humanID} gait challenge problem: Data sets, performance, and
  analysis,'' \emph{IEEE Transactions on Pattern Analysis and Machine
  intelligence}, vol.~27, no.~2, pp. 162--177, January 2005.

\bibitem{DB-CASIAC}
D.~Tan, K.~Huang, S.~Yu, and T.~Tan, ``Efficient night gait recognition based
  on template matching,'' in \emph{International Conference on Pattern
  Recognition}, Hong Kong, China, August 2006.

\bibitem{ISRT1}
A.~Tsuji, Y.~Makihara, and Y.~Yagi, ``Silhouette transformation based on
  walking speed for gait identification,'' in \emph{Computer Vision and Pattern
  Recognition}, San Francisco, CA, USA, June 2010.

\bibitem{ISRT2}
M.~A. Hossain, Y.~Makihara, J.~Wang, and Y.~Yagi, ``Clothing-invariant gait
  identification using part-based clothing categorization and adaptive weight
  control,'' \emph{Pattern Recognition}, vol.~43, no.~6, pp. 2281--2291, 2010.

\bibitem{ISRT4}
Y.~Makihara, H.~Mannami, and Y.~Yagi, ``Gait analysis of gender and age using a
  large-scale multi-view gait database,'' in \emph{Asian Conference on Computer
  Vision}, Queenstown, New Zealand, November 2010.

\bibitem{DB-OU}
H.~{Iwama}, M.~{Okumura}, Y.~{Makihara}, and Y.~{Yagi}, ``The {OU-ISIR} gait
  database comprising the large population dataset and performance evaluation
  of gait recognition,'' \emph{IEEE Transactions on Information Forensics and
  Security}, vol.~7, no.~5, pp. 1511--1521, June 2012.

\bibitem{DB-TUM}
M.~{Hofmann}, S.~{Bachmann}, and G.~{Rigoll}, ``{2.5D} gait biometrics using
  the depth gradient histogram energy image,'' in \emph{IEEE International
  Conference on Biometrics: Theory, Applications and Systems}, Washington, DC,
  USA, September 2012.

\bibitem{OUBAG}
M.~Z. Uddin, T.~T. Ngo, Y.~Makihara, N.~Takemura, X.~Li, D.~Muramatsu, and
  Y.~Yagi, ``The {OU-ISIR} large population gait database with real-life
  carried object and its performance evaluation,'' \emph{IPSJ Transactions on
  Computer Vision and Applications}, vol.~10, no.~1, pp. 1--11, May 2018.

\bibitem{MVLP}
N.~Takemura, Y.~Makihara, D.~Muramatsu, T.~Echigo, and Y.~Yagi, ``Multi-view
  large population gait dataset and its performance evaluation for cross-view
  gait recognition,'' \emph{IPSJ Transactions on Computer Vision and
  Applications}, vol.~10, no.~1, p.~4, February 2018.

\bibitem{SOTA23}
Y.~Zhang, Y.~Huang, L.~Wang, and S.~Yu, ``A comprehensive study on gait
  biometrics using a joint cnn-based method,'' \emph{Pattern Recognition},
  vol.~93, pp. 228--236, September 2019.

\bibitem{CASIAE}
C.~Song, Y.~Huang, W.~Wang, L.~Wang, and T.~Tan, ``{CASIA-E}: a big dataset for
  gait recognition,'' \emph{Submitted to IEEE Transactions on Pattern Analysis
  and Machine Intelligence}, 2020.

\bibitem{Tbiom2}
W.~An, S.~Yu, Y.~Makihara, X.~Wu, C.~Xu, Y.~Yu, R.~Liao, and Y.~Yagi,
  ``Performance evaluation of model-based gait on multi-view very large
  population database with pose sequences,'' \emph{IEEE Transactions on
  Biometrics, Behavior, and Identity Science}, vol.~2, no.~4, pp. 421--430,
  October 2020.

\bibitem{deepsurvey}
M.~Wang and W.~Deng, ``Deep face recognition: A survey,'' \emph{arXiv
  1804.06655}, February 2019.

\bibitem{sundararajan2018deep}
K.~Sundararajan and D.~L. Woodard, ``Deep learning for biometrics: A survey,''
  \emph{ACM Computing Surveys}, vol.~51, no.~3, pp. 1--34, May 2018.

\bibitem{Compet}
\BIBentryALTinterwordspacing
{TC4} competition and workshop on human identification at a distance 2020.
  [Online]. Available: \url{http://hid2020.iapr-tc4.org/}
\BIBentrySTDinterwordspacing

\bibitem{openpose}
Z.~Cao, T.~Simon, S.-E. Wei, and Y.~Sheikh, ``Realtime multi-person {2D} pose
  estimation using part affinity fields,'' in \emph{Computer Vision and Pattern
  Recognition}, Honolulu, HW, USA, July 2017.

\bibitem{Alpha}
H.-S. Fang, S.~Xie, Y.-W. Tai, and C.~Lu, ``{RMPE}: Regional multi-person pose
  estimation,'' in \emph{International Conference on Computer Vision}, Venice,
  Italy, October 2017.

\bibitem{tax2}
T.~K. Lee, M.~Belkhatir, and S.~Sanei, ``A comprehensive review of past and
  present vision-based techniques for gait recognition,'' \emph{Multimedia
  tools and applications}, vol.~72, no.~3, pp. 2833--2869, July 2014.

\bibitem{sil}
M.~Nieto-Hidalgo, F.~J. Ferr{\'a}ndez-Pastor, R.~J. Valdivieso-Sarabia,
  J.~Mora-Pascual, and J.~M. Garc{\'\i}a-Chamizo, ``Vision based extraction of
  dynamic gait features focused on feet movement using rgb camera,'' in
  \emph{Ambient Intelligence for Health}, Puerto Varas, Chile, December 2015.

\bibitem{TanmayThesis}
T.~Verlekar, ``Gait analysis in unconstrained environments,'' Ph.D.
  dissertation, Electrical and Computer Engineering, Instituto Superior
  Técnico, University of Lisbon, Lisbon, Portugal, June 2019.

\bibitem{opt1}
A.~Sokolova and A.~Konushin, ``View resistant gait recognition,'' in
  \emph{International Conference on Video and Image Processing}, Shanghai,
  China, December 2019.

\bibitem{opt2}
F.~M. {Castro}, M.~J. {Marin-Jimenez}, N.~{Guil}, S.~{Lopez-Tapia}, and
  N.~{Perez de la Blanca}, ``Evaluation of {CNN} architectures for gait
  recognition based on optical flow maps,'' in \emph{International Conference
  of the Biometrics Special Interest Group}, Darmstadt, Germany, October 2017.

\bibitem{skl1}
E.~Gianaria, N.~Balossino, M.~Grangetto, and M.~Lucenteforte, ``Gait
  characterization using dynamic skeleton acquisition,'' in \emph{International
  Workshop on Multimedia Signal Processing}, Pula, Italy, October 2013.

\bibitem{pose}
R.~Alp~G{\"u}ler, N.~Neverova, and I.~Kokkinos, ``Densepose: Dense human pose
  estimation in the wild,'' in \emph{Computer Vision and Pattern Recognition},
  Salt Lake City, UT, USA, June 2018.

\bibitem{skel}
Y.~{Wang}, J.~{Sun}, J.~{Li}, and D.~{Zhao}, ``Gait recognition based on {3D}
  skeleton joints captured by kinect,'' in \emph{International Conference on
  Image Processing}, Phoenix, AZ, USA, September 2016.

\bibitem{posecomp}
D.~Zhang and M.~Shah, ``Human pose estimation in videos,'' in
  \emph{International Conference on Computer Vision}, Santiago, Chile, December
  2015.

\bibitem{GEI}
J.~{Han} and B.~{Bhanu}, ``Individual recognition using gait energy image,''
  \emph{IEEE Transactions on Pattern Analysis and Machine Intelligence},
  vol.~28, no.~2, pp. 316--322, December 2006.

\bibitem{GEI1}
C.~{Wang}, J.~{Zhang}, p.~{Xiaoru}, and L.~{Wang}, ``{Chrono-Gait} image: A
  novel temporal template for gait recognition,'' in \emph{European Conference
  on Computer Vision}, Washington, DC, USA, December 2011.

\bibitem{FDEI}
C.~Chen, J.~Liang, H.~Zhao, H.~Hu, and J.~Tian, ``Frame difference energy image
  for gait recognition with incomplete silhouettes,'' \emph{Pattern Recognition
  Letters}, vol.~30, no.~11, pp. 977--984, August 2009.

\bibitem{GENI}
K.~{Bashir}, T.~{Xiang}, and S.~{Gong}, ``Gait recognition using gait entropy
  image,'' in \emph{International Conference on Imaging for Crime Detection and
  Prevention}, London, UK, December 2009.

\bibitem{B7}
Y.~{He}, J.~{Zhang}, H.~{Shan}, and L.~{Wang}, ``Multi-task {GANs} for
  view-specific feature learning in gait recognition,'' \emph{IEEE Transactions
  on Information Forensics and Security}, vol.~14, no.~1, pp. 102--113, January
  2019.

\bibitem{SOTA30}
R.~Liao, C.~Cao, E.~B. Garcia, S.~Yu, and Y.~Huang, ``Pose-based
  temporal-spatial network ({PTSN}) for gait recognition with carrying and
  clothing variations,'' in \emph{Chinese Conference on Biometric Recognition},
  Beijing, China, October 2017.

\bibitem{LSTM3}
D.~Liu, M.~Ye, X.~Li, F.~Zhang, and L.~Lin, ``Memory-based gait recognition.''
  in \emph{The British Machine Vision Conference}, York, UK, September 2016.

\bibitem{SOTA22}
W.~Xing, Y.~Li, and S.~Zhang, ``View-invariant gait recognition method by
  three-dimensional convolutional neural network,'' \emph{Journal of Electronic
  Imaging}, vol.~27, no.~1, p. 013010, January 2018.

\bibitem{arXiv}
B.~Lin, S.~Zhang, X.~Yu, Z.~Chu, and H.~Zhang, ``Learning effective
  representations from global and local features for cross-view gait
  recognition,'' \emph{arXiv:2011.01461}, November 2020.

\bibitem{GCN2}
N.~Li, X.~Zhaoa, and C.~Ma, ``{JointsGait}:a model-based gait recognition
  method based on gait graph convolutional networks and joints relationship
  pyramid mapping,'' \emph{arXiv:2005.08625}, December 2020.

\bibitem{ICPR}
A.~Sepas-Moghaddam, S.~Ghorbani, N.~F. Troje, and A.~Etemad, ``Gait recognition
  using multi-scale partial representation transformation with capsules,'' in
  \emph{International Conference on Pattern Recognition}, Milan, Italy, January
  2021.

\bibitem{part}
R.~Imad, ``Towards human body-part learning for model-free gait recognition,''
  \emph{arXiv:1904.01620}, April 2019.

\bibitem{R9}
Y.~Fu, Y.~Wei, Y.~Zhou, H.~Shi, G.~Huang, X.~Wang, Z.~Yao, and T.~Huang,
  ``Horizontal pyramid matching for person re-identification,'' in \emph{AAAI
  Conference on Artificial Intelligence}, Honolulu, HW, USA, February 2019.

\bibitem{ReLU}
A.~F. Agarap, ``Deep learning using rectified linear units ({ReLU}),''
  \emph{arXiv:1803.08375}, February 2019.

\bibitem{tanh}
B.~Karlik and A.~V. Olgac, ``Performance analysis of various activation
  functions in generalized {MLP} architectures of neural networks,''
  \emph{International Journal of Artificial Intelligence and Expert Systems},
  vol.~1, no.~4, pp. 111--122, December 2011.

\bibitem{resnet}
K.~He, X.~Zhang, S.~Ren, and J.~Sun, ``Deep residual learning for image
  recognition,'' in \emph{Proceedings of the IEEE conference on computer vision
  and pattern recognition}, Las Vegas, NV, USA, July 2016.

\bibitem{inception}
C.~Szegedy, V.~Vanhoucke, S.~Ioffe, J.~Shlens, and Z.~Wojna, ``Rethinking the
  inception architecture for computer vision,'' in \emph{computer vision and
  pattern recognition}, Las Vegas, NV, USA, July 2016.

\bibitem{DBN}
G.~E. Hinton, S.~Osindero, and Y.-W. Teh, ``A fast learning algorithm for deep
  belief nets,'' \emph{Neural computation}, vol.~18, no.~7, pp. 1527--1554, May
  2006.

\bibitem{RBM}
G.~E. Hinton, T.~J. Sejnowski \emph{et~al.}, ``Learning and relearning in
  boltzmann machines,'' \emph{Parallel distributed processing: Explorations in
  the microstructure of cognition}, vol.~1, no.~2, pp. 282--317, 1986.

\bibitem{Biom}
M.~Benouis, M.~Senouci, R.~Tlemsani, and L.~Mostefai, ``Gait recognition based
  on model-based methods and deep belief networks,'' \emph{International
  Journal of Biometrics}, vol.~8, no.~3, pp. 237--253, March 2017.

\bibitem{RNNSurvey}
Z.~C. Lipton, J.~Berkowitz, and C.~Elkan, ``A critical review of recurrent
  neural networks for sequence learning,'' \emph{arXiv:1506.00019}, 2015.

\bibitem{LSTM1}
Y.~Feng, Y.~Li, and J.~Luo, ``Learning effective gait features using {LSTM},''
  in \emph{International Conference on Pattern Recognition}, Cancun, Mexico,
  September 2016.

\bibitem{LSTM2}
F.~Battistone and A.~Petrosino, ``{TGLSTM}: A time based graph deep learning
  approach to gait recognition,'' \emph{Pattern Recognition Letters}, vol. 126,
  pp. 132--138, September 2019.

\bibitem{B4}
Z.~{Zhang}, L.~{Tran}, F.~{Liu}, and X.~{Liu}, ``On learning disentangled
  representations for gait recognition,'' \emph{IEEE Transactions on Pattern
  Analysis and Machine Intelligence}, vol. in press, pp. 1--16, August 2020.

\bibitem{LSTM}
S.~Hochreiter and J.~Schmidhuber, ``Long short-term memory,'' \emph{Neural
  computation}, vol.~9, no.~8, pp. 1735--1780, November 1997.

\bibitem{GRU1}
K.~Cho, B.~V. Merri{\"e}nboer, D.~Bahdanau, and Y.~Bengio, ``On the properties
  of neural machine translation: Encoder-decoder approaches,''
  \emph{arXiv:1409.1259}, October 2014.

\bibitem{B2}
Y.~Wang, B.~Du, Y.~Shen, K.~Wu, G.~Zhao, J.~Sun, and H.~Wen, ``{EV-Gait}:
  Event-based robust gait recognition using dynamic vision sensors,'' in
  \emph{Computer Vision and Pattern Recognition}, Long Beach, CA, USA, June
  2019.

\bibitem{B10}
S.~Yu, R.~Liao, W.~An, H.~Chen, E.~García, Y.~Huang, and N.~Poh, ``Gaitnet: An
  end-to-end network for gait based human identification,'' \emph{Pattern
  Recognition}, vol.~96, no.~1, pp. 1--11, December 2019.

\bibitem{R6}
X.~{Li}, Y.~{Makihara}, C.~{Xu}, Y.~{Yagi}, and M.~{Ren}, ``Joint intensity
  transformer network for gait recognition robust against clothing and carrying
  status,'' \emph{IEEE Transactions on Information Forensics and Security},
  vol.~14, no.~12, pp. 3102--3115, December 2019.

\bibitem{DAE1}
S.~Yu, H.~Chen, Q.~Wang, L.~Shen, and Y.~Huang, ``Invariant feature extraction
  for gait recognition using only one uniform model,'' \emph{Neurocomputing},
  vol. 239, pp. 81--93, May 2017.

\bibitem{CVPR20202}
X.~Li, Y.~Makihara, C.~Xu, Y.~Yagi, and M.~Ren, ``Gait recognition via
  semi-supervised disentangled representation learning to identity and
  covariate features,'' in \emph{Computer Vision and Pattern Recognition},
  Seattle, WA, USA, June 2020.

\bibitem{GoogleNet}
C.~{Szegedy}, {Wei Liu}, {Yangqing Jia}, P.~{Sermanet}, S.~{Reed},
  D.~{Anguelov}, D.~{Erhan}, V.~{Vanhoucke}, and A.~{Rabinovich}, ``Going
  deeper with convolutions,'' in \emph{Computer Vision and Pattern
  Recognition}, Boston, MA, USA, June 2015.

\bibitem{goodfellow}
I.~J. Goodfellow, J.~Pouget-Abadie, M.~Mirza, B.~Xu, D.~Warde-Farley, S.~Ozair,
  A.~Courville, and Y.~Bengio, ``Generative adversarial networks,''
  \emph{arXiv:1406.2661}, June 2014.

\bibitem{GAN5}
S.~{Yu}, H.~{Chen}, E.~B.~G. {Reyes}, and N.~{Poh}, ``{GaitGAN}: Invariant gait
  feature extraction using generative adversarial networks,'' in \emph{Computer
  Vision and Pattern Recognition Workshops}, Honolulu, HI, USA, August 2017.

\bibitem{GAN6}
S.~Yu, R.~Liao, W.~An, H.~Chen, E.~B. García, Y.~Huang, and N.~Poh,
  ``{GaitGANv2}: Invariant gait feature extraction using generative adversarial
  networks,'' \emph{Pattern Recognition}, vol.~87, pp. 179 -- 189, March 2019.

\bibitem{B9}
B.~Hu, Y.~Gao, Y.~Guan, Y.~Long, N.~Lane, and T.~Ploetz, ``Robust cross-view
  gait identification with evidence: A discriminant gait {GAN} {(DiGGAN)}
  approach on 10000 people,'' \emph{arXiv:1811.10493}, 2018.

\bibitem{GAN3}
Y.~Wang, C.~Song, Y.~Huang, Z.~Wang, and L.~Wang, ``Learning view invariant
  gait features with two-stream {GAN},'' \emph{Neurocomputing}, vol. 339, pp.
  245--254, April 2019.

\bibitem{GAN4}
X.~Li, Y.~Makihara, C.~Xu, Y.~Yagi, and M.~Ren, ``Gait recognition invariant to
  carried objects using alpha blending generative adversarial networks,''
  \emph{Pattern Recognition}, vol. 105, pp. 1--12, September 2020.

\bibitem{caps}
S.~Sabour, N.~Frosst, and G.~E. Hinton, ``Dynamic routing between capsules,''
  in \emph{Advances in neural information processing systems}, Long Beach, CA,
  USA, December 2017.

\bibitem{capsule1}
Z.~Xu, W.~Lu, Q.~Zhang, Y.~Yeung, and X.~Chen, ``Gait recognition based on
  capsule network,'' \emph{Journal of Visual Communication and Image
  Representation}, vol.~59, pp. 159--167, February 2019.

\bibitem{spyder}
A.~Zhao, J.~Li, and M.~Ahmed, ``{SpiderNet}: A spiderweb graph neural network
  for multi-view gait recognition,'' \emph{Knowledge-Based Systems}, vol. 206,
  pp. 1--14, October 2020.

\bibitem{3DCNNICIP}
T.~{Wolf}, M.~{Babaee}, and G.~{Rigoll}, ``Multi-view gait recognition using 3d
  convolutional neural networks,'' in \emph{IEEE International Conference on
  Image Processing}, Phoenix, AZ, USA, September 2016.

\bibitem{GCNFilter}
F.~Monti, D.~Boscaini, J.~Masci, E.~Rodola, J.~Svoboda, and M.~M. Bronstein,
  ``Geometric deep learning on graphs and manifolds using mixture model cnns,''
  in \emph{Computer Vision and Pattern Recognition}, Honolulu, Hawaii, July
  2017.

\bibitem{CNNLSTM1}
G.~{Batchuluun}, H.~S. {Yoon}, J.~K. {Kang}, and K.~R. {Park}, ``Gait-based
  human identification by combining shallow convolutional neural
  network-stacked long short-term memory and deep convolutional neural
  network,'' \emph{IEEE Access}, vol.~6, pp. 63\,164--63\,186, October 2018.

\bibitem{TIP}
Y.~{Zhang}, Y.~{Huang}, S.~{Yu}, and L.~{Wang}, ``Cross-view gait recognition
  by discriminative feature learning,'' \emph{IEEE Transactions on Image
  Processing}, vol.~29, no.~1, pp. 1001--1015, 2020.

\bibitem{DAEGAN2}
S.~{Li}, W.~{Liu}, H.~{Ma}, and S.~{Zhu}, ``Beyond view transformation:
  Cycle-consistent global and partial perception gan for view-invariant gait
  recognition,'' in \emph{International Conference on Multimedia and Expo}, San
  Diego, CA, USA, July 2018.

\bibitem{TMM}
Z.~{Wu}, Y.~{Huang}, and L.~{Wang}, ``Learning representative deep features for
  image set analysis,'' \emph{IEEE Transactions on Multimedia}, vol.~17,
  no.~11, pp. 1960--1968, November 2015.

\bibitem{SOTA3}
C.~{Zhang}, W.~{Liu}, H.~{Ma}, and H.~{Fu}, ``Siamese neural network based gait
  recognition for human identification,'' in \emph{International Conference on
  Acoustics, Speech and Signal Processing}, Shanghai, China, March 2016.

\bibitem{SOTA4}
M.~Alotaibi and A.~Mahmood, ``Improved gait recognition based on specialized
  deep convolutional neural network,'' \emph{Computer Vision and Image
  Understanding}, vol. 164, pp. 103--110, November 2017.

\bibitem{SOTA5}
C.~Li, X.~Min, S.~Sun, W.~Lin, and Z.~Tang, ``Deepgait: A learning deep
  convolutional representation for view-invariant gait recognition using joint
  bayesian,'' \emph{Applied Sciences}, vol.~7, no.~3, p. 210, February 2017.

\bibitem{SOTA6}
N.~{Takemura}, Y.~{Makihara}, D.~{Muramatsu}, T.~{Echigo}, and Y.~{Yagi}, ``On
  input/output architectures for convolutional neural network-based cross-view
  gait recognition,'' \emph{IEEE Transactions on Circuits and Systems for Video
  Technology}, vol.~29, no.~9, pp. 2708--2719, October 2019.

\bibitem{SOTA7}
F.~M. {Castro}, M.~J. {Marin-Jimenez}, N.~{Guil}, S.~{Lopez-Tapia}, and
  N.~{Perez de la Blanca}, ``Evaluation of cnn architectures for gait
  recognition based on optical flow maps,'' in \emph{International Conference
  of the Biometrics Special Interest Group}, Darmstadt, Germany, September
  2017.

\bibitem{SOTA31}
S.~Tong, H.~Ling, Y.~Fu, and D.~Wang, ``Cross-view gait identification with
  embedded learning,'' in \emph{ACM Multimedia}, Mountain View, CA, USA,
  October 2017.

\bibitem{SOTA21}
W.~Liu, C.~Zhang, H.~Ma, and S.~Li, ``Learning efficient spatial-temporal gait
  features with deep learning for human identification,''
  \emph{Neuroinformatics}, vol.~16, no. 3-4, pp. 457--471, February 2018.

\bibitem{SOTA8}
S.~{Tong}, Y.~{Fu}, X.~{Yue}, and H.~{Ling}, ``Multi-view gait recognition
  based on a spatial-temporal deep neural network,'' \emph{IEEE Access},
  vol.~6, pp. 57\,583--57\,596, October 2018.

\bibitem{SOTA9}
D.~{Thapar}, A.~{Nigam}, D.~{Aggarwal}, and P.~{Agarwal}, ``{VGR-net}: A view
  invariant gait recognition network,'' in \emph{International Conference on
  Identity, Security, and Behavior Analysis}, Singapore, Singapore, March 2018.

\bibitem{SOTA17}
H.~Wu, J.~Weng, X.~Chen, and W.~Lu, ``Feedback weight convolutional neural
  network for gait recognition,'' \emph{Journal of Visual Communication and
  Image Representation}, vol.~55, pp. 424--432, August 2018.

\bibitem{SOTA18}
W.~An, R.~Liao, S.~Yu, Y.~Huang, and P.~C. Yuen, ``Improving gait recognition
  with 3d pose estimation,'' in \emph{Chinese Conference on Biometric
  Recognition}, Urumchi, China, August 2018.

\bibitem{PRL}
F.~Battistone and A.~Petrosino, ``{TGLSTM}: A time based graph deep learning
  approach to gait recognition,'' \emph{Pattern Recognition Letters}, vol. 126,
  pp. 132--138, September 2019.

\bibitem{SOTA27}
S.~Tong, Y.~Fu, and H.~Ling, ``Gait recognition with cross-domain transfer
  networks,'' \emph{Journal of Systems Architecture}, vol.~93, pp. 40--47,
  February 2019.

\bibitem{SOTA24}
S.-b. Tong, Y.-z. Fu, and H.-f. Ling, ``Cross-view gait recognition based on a
  restrictive triplet network,'' \emph{Pattern Recognition Letters}, vol. 125,
  pp. 212--219, July 2019.

\bibitem{SOTA10}
K.~{Zhang}, W.~{Luo}, L.~{Ma}, W.~{Liu}, and H.~{Li}, ``Learning joint gait
  representation via quintuplet loss minimization,'' in \emph{Computer Vision
  and Pattern Recognition}, Long Beach, CA, USA, June 2019.

\bibitem{SOTA26}
P.~Zhang, Q.~Wu, and J.~Xu, ``Vt-gan: View transformation gan for gait
  recognition across views,'' in \emph{International Joint Conference on Neural
  Networks}, Budapest, Hungary, July 2019.

\bibitem{SOTA25}
S.~Li, W.~Liu, and H.~Ma, ``Attentive spatial--temporal summary networks for
  feature learning in irregular gait recognition,'' \emph{IEEE Transactions on
  Multimedia}, vol.~21, no.~9, pp. 2361--2375, September 2019.

\bibitem{SOTA29}
P.~Zhang, Q.~Wu, and J.~Xu, ``{VN-GAN}: Identity-preserved variation
  normalizing gan for gait recognition,'' in \emph{International Joint
  Conference on Neural Networks}, Budapest, Hungary, July 2019.

\bibitem{SOTA28}
X.~Wang, J.~Zhang, and W.~Q. Yan, ``Gait recognition using multichannel
  convolution neural networks,'' \emph{Neural Computing and Applications},
  vol.~32, p. 14275–14285, October 2019.

\bibitem{NCAA}
X.~Wang and W.~Q. Yan, ``Cross-view gait recognition through ensemble
  learning,'' \emph{Neural computing and applications}, vol.~32, no.~11, pp.
  7275--7287, May 2019.

\bibitem{capsule2}
Y.~Wu, J.~Hou, Y.~Su, C.~Wu, M.~Huang, and Z.~Zhu, ``Gait recognition based on
  feedback weight capsule network,'' in \emph{Information Technology,
  Networking, Electronic and Automation Control Conference}, Chongqing, China,
  May 2020.

\bibitem{RNNAE}
K.~Jun, D.-W. Lee, K.~Lee, S.~Lee, and M.~S. Kim, ``Feature extraction using an
  rnn autoencoder for skeleton-based abnormal gait recognition,'' \emph{IEEE
  Access}, vol.~8, pp. 19\,196--19\,207, January 2020.

\bibitem{PR5}
R.~Liao, S.~Yu, W.~An, and Y.~Huang, ``A model-based gait recognition method
  with body pose and human prior knowledge,'' \emph{Pattern Recognition},
  vol.~98, pp. 1--11, February 2020.

\bibitem{SOTA11}
X.~Wang and J.~Zhang, ``Gait feature extraction and gait classification using
  two-branch cnn,'' \emph{Multimedia Tools and Applications}, vol.~79, no.~3,
  pp. 2917--2930, January 2020.

\bibitem{SOTA16}
X.~Wang and W.~Q. Yan, ``Human gait recognition based on frame-by-frame gait
  energy images and convolutional long short-term memory,'' \emph{International
  journal of neural systems}, vol.~30, no.~1, pp. 1--12, January 2020.

\bibitem{CSVT}
C.~{Xu}, Y.~{Makihara}, X.~{Li}, Y.~{Yagi}, and J.~{Lu}, ``Cross-view gait
  recognition using pairwise spatial transformer networks,'' \emph{IEEE
  Transactions on Circuits and Systems for Video Technology}, vol.~31, no.~1,
  pp. 260--274, January 2021.

\bibitem{MTAP2}
X.~Wang and W.~Q. Yan, ``Non-local gait feature extraction and human
  identification,'' \emph{Multimedia Tools and Applications}, vol.~80, pp.
  6065–--6078, October 2020.

\bibitem{MTAP3}
X.~Wang and K.~Yan, ``Gait classification through {CNN-based} ensemble
  learning,'' \emph{Multimedia Tools and Applications}, vol.~80, pp. 1--17,
  September 2020.

\bibitem{SOTA15}
J.~Wen, ``Gait recognition based on {GF-CNN} and metric learning,''
  \emph{Journal of Information Processing Systems}, vol.~16, no.~5, pp.
  1105--1112, October 2020.

\bibitem{MTAP4}
A.~Mehmood, M.~A. Khan, M.~Sharif, S.~A. Khan, M.~Shaheen, T.~Saba, N.~Riaz,
  and I.~Ashraf, ``Prosperous human gait recognition: an end-to-end system
  based on pre-trained cnn features selection,'' \emph{Multimedia Tools and
  Applications}, April 2020.

\bibitem{Supercom}
O.~Elharrouss, N.~Almaadeed, S.~Al-Maadeed, and A.~Bouridane, ``Gait
  recognition for person re-identification,'' \emph{The Journal of
  Supercomputing}, vol. in press, pp. 1--20, August 2020.

\bibitem{ACCVW1}
J.~Pan, H.~Sun, Y.~Wu, S.~Yin, and S.~Wang, ``Optimization of {GaitSet} for
  gait recognition,'' in \emph{Asian Conference on Computer Vision}, Kyoto,
  Japan, November 2020.

\bibitem{ACCVW2}
P.~Zhang, Z.~Song, and X.~Xing, ``Multi-grid spatial and temporal feature
  fusion for human identification at a distance,'' in \emph{Asian Conference on
  Computer Vision}, Kyoto, Japan, November 2020.

\bibitem{SOTA20}
G.~Huang, Z.~Lu, C.-M. Pun, and L.~Cheng, ``Flexible gait recognition based on
  flow regulation of local features between key frames,'' \emph{IEEE Access},
  vol.~8, pp. 75\,381--75\,392, April 2020.

\bibitem{ICASSP3}
J.~Su, Y.~Zhao, and X.~Li, ``Deep metric learning based on center-ranked loss
  for gait recognition,'' in \emph{International Conference on Acoustics,
  Speech and Signal Processing}, Barcelona, Spain, May 2020.

\bibitem{SOTA19}
R.~Liao, W.~An, S.~Yu, Z.~Li, and Y.~Huang, ``Dense-view geis set: View space
  covering for gait recognition based on dense-view gan,'' in
  \emph{International Joint Conference on Biometrics}, Houston, USA, September
  2020.

\bibitem{skl2}
Z.~Liu, J.~Zhu, J.~Bu, and C.~Chen, ``A survey of human pose estimation: the
  body parts parsing based methods,'' \emph{Journal of Visual Communication and
  Image Representation}, vol.~32, pp. 10--19, October 2015.

\bibitem{cent}
P.-T. De~Boer, D.~P. Kroese, S.~Mannor, and R.~Y. Rubinstein, ``A tutorial on
  the cross-entropy method,'' \emph{Annals of operations research}, vol. 134,
  no.~1, pp. 19--67, February 2005.

\bibitem{TS}
A.~Hermans, L.~Beyer, and B.~Leibe, ``In defense of the triplet loss for person
  re-identification,'' in \emph{Computer Vision and Pattern Recognition}, Long
  Beach, CA, USA, June 2019.

\bibitem{contloss}
B.~Ghojogh, M.~Sikaroudi, S.~Shafiei, H.~R. Tizhoosh, F.~Karray, and
  M.~Crowley, ``Fisher discriminant triplet and contrastive losses for training
  siamese networks,'' in \emph{International Joint Conference on Neural
  Networks}, Glasgow, UK, July 2020.

\bibitem{softloss}
R.~Ranjan, C.~D. Castillo, and R.~Chellappa, ``L2-constrained softmax loss for
  discriminative face verification,'' \emph{arXiv:1703.09507}, March 2017.

\bibitem{arc}
J.~Deng, J.~Guo, N.~Xue, and S.~Zafeiriou, ``Arcface: Additive angular margin
  loss for deep face recognition,'' in \emph{Computer Vision and Pattern
  Recognition}, Long Beach, CA, USA, June 2019.

\bibitem{centerloss}
Y.~Wen, K.~Zhang, Z.~Li, and Y.~Qiao, ``A discriminative feature learning
  approach for deep face recognition,'' in \emph{European conference on
  computer vision}, Amsterdam, Netherlands, October 2016.

\bibitem{euc}
G.~De~Soete and J.~D. Carroll, ``K-means clustering in a low-dimensional
  euclidean space,'' in \emph{New approaches in classification and data
  analysis}.\hskip 1em plus 0.5em minus 0.4em\relax Springer, 1994, pp.
  212--219.

\bibitem{taa2}
B.~Biggio, Z.~Akhtar, G.~Fumera, G.~L. Marcialis, and F.~Roli, ``Security
  evaluation of biometric authentication systems under real spoofing attacks,''
  \emph{IET biometrics}, vol.~1, no.~1, pp. 11--24, March 2012.

\bibitem{taa3}
D.~Gafurov, E.~Snekkenes, and P.~Bours, ``Spoof attacks on gait authentication
  system,'' \emph{IEEE Transactions on Information Forensics and Security},
  vol.~2, no.~3, pp. 491--502, Agust 2007.

\bibitem{aas1}
N.~Akhtar and A.~Mian, ``Threat of adversarial attacks on deep learning in
  computer vision: A survey,'' \emph{IEEE Access}, vol.~6, February 2018.

\bibitem{aas2}
N.~Akhtar, A.~Mian, N.~Kardan, and M.~Shah, ``Threat of adversarial attacks on
  deep learning in computer vision: {Survey II},'' \emph{arXiv:2108.00401},
  August 2021.

\bibitem{aas3}
N.~Akhtar, M.~Jalwana, M.~Bennamoun, and A.~S. Mian, ``Attack to fool and
  explain deep networks,'' \emph{IEEE Transactions on Pattern Analysis and
  Machine Intelligence}, vol. in press, pp. 1--16, May 2021.

\bibitem{aafirst}
C.~Szegedy, W.~Zaremba, I.~Sutskever, J.~Bruna, D.~Erhan, I.~Goodfellow, and
  R.~Fergus, ``Intriguing properties of neural networks,'' in
  \emph{International Conference on Learning Representations}, Banff, Canada,
  April 2014.

\bibitem{aagan1}
S.~Tulyakov, M.-Y. Liu, X.~Yang, and J.~Kautz, ``Mocogan: Decomposing motion
  and content for video generation,'' in \emph{computer vision and pattern
  recognition}, Salt Lake City, UT, USA, June 2018.

\bibitem{aagan2}
Y.~Wang, P.~Bilinski, F.~Bremond, and A.~Dantcheva, ``Imaginator: Conditional
  spatio-temporal gan for video generation,'' in \emph{Winter Conference on
  Applications of Computer Vision}, Snowmass Village, CO, USA, March 2020.

\bibitem{aagan3}
M.~Treu, T.-N. Le, H.~H. Nguyen, J.~Yamagishi, and I.~Echizen, ``Fashion-guided
  adversarial attack on person segmentation,'' in \emph{Computer Vision and
  Pattern Recognition}, Virtual, June 2021.

\bibitem{aa1}
M.~Jia, H.~Yang, D.~Huang, and Y.~Wang, ``Attacking gait recognition systems
  via silhouette guided {GANs},'' in \emph{ACM International Conference on
  Multimedia}, Nice, France, October 2019.

\bibitem{aa2}
Z.~He, W.~Wang, J.~Dong, and T.~Tan, ``Temporal sparse adversarial attack on
  sequence-based gait recognition,'' \emph{arXiv:2002.09674}, February 2020.

\bibitem{distface1}
X.~Peng, X.~Yu, K.~Sohn, D.~N. Metaxas, and M.~Chandraker,
  ``Reconstruction-based disentanglement for pose-invariant face recognition,''
  in \emph{International Conference on Computer Vision}, Venice, Italy, October
  2017.

\bibitem{distface2}
L.~Tran, X.~Yin, and X.~Liu, ``Disentangled representation learning gan for
  pose-invariant face recognition,'' in \emph{Computer Vision and Pattern
  Recognition}, Honolulu, Hawaii, July 2017.

\bibitem{distact}
Z.~Liu, H.~Zhang, Z.~Chen, Z.~Wang, and W.~Ouyang, ``Disentangling and unifying
  graph convolutions for skeleton-based action recognition,'' in \emph{Computer
  Vision and Pattern Recognition}, Seattle, WA, USA, June 2020.

\bibitem{dist}
X.~Liu, ``Disentanglement for discriminative visual recognition,''
  \emph{arXiv:2006.07810}, July 2020.

\bibitem{distpos}
J.~Gu, Z.~Wang, W.~Ouyang, J.~Li, L.~Zhuo \emph{et~al.}, ``3d hand pose
  estimation with disentangled cross-modal latent space,'' in \emph{Computer
  Vision and Pattern Recognition}, Seattle, WA, USA, June 2020.

\bibitem{dist1}
Y.~Bengio, A.~Courville, and P.~Vincent, ``Representation learning: A review
  and new perspectives,'' \emph{IEEE transactions on pattern analysis and
  machine intelligence}, vol.~35, no.~8, pp. 1798--1828, March 2013.

\bibitem{dist2}
H.~Li, S.~Wang, R.~Wan, and A.~K. Chichung, ``{GMFAD}: Towards generalized
  visual recognition via multi-layer feature alignment and disentanglement,''
  \emph{IEEE Transactions on Pattern Analysis and Machine Intelligence}, vol.
  in press, pp. 1--15, August 2020.

\bibitem{dist3}
A.~Achille and S.~Soatto, ``Emergence of invariance and disentanglement in deep
  representations,'' \emph{The Journal of Machine Learning Research}, vol.~19,
  no.~1, pp. 1947--1980, September 2018.

\bibitem{dist5}
E.~Denton and V.~Birodkar, ``Unsupervised learning of disentangled
  representations from video,'' \emph{arXiv:1705.10915}, May 2017.

\bibitem{dist6}
Y.~Khraimeche, G.-A. Bilodeau, D.~Steele, and H.~Mahadik, ``Unsupervised
  disentanglement gan for domain adaptive person re-identification,''
  \emph{arXiv:2007.15560}, July 2020.

\bibitem{dist7}
K.~Ridgeway and M.~C. Mozer, ``Learning deep disentangled embeddings with the
  f-statistic loss,'' \emph{arXiv:1802.05312}, May 2018.

\bibitem{self}
L.~Jing and Y.~Tian, ``Self-supervised visual feature learning with deep neural
  networks: A survey,'' \emph{IEEE Transactions on Pattern Analysis and Machine
  Intelligence}, vol. in press, pp. 1--24, May 2020.

\bibitem{self1}
H.~Rao, S.~Wang, X.~Hu, M.~Tan, H.~Da, J.~Cheng, and B.~Hu, ``Self-supervised
  gait encoding with locality-aware attention for person re-identification,''
  in \emph{International Joint Conference on Artificial Intelligence},
  Yokohama, Japan, January 2020.

\bibitem{self2}
H.~Rao, S.~Wang, X.~Hu, M.~Tan, Y.~Guo, J.~Cheng, B.~Hu, and X.~Liu, ``A
  self-supervised gait encoding approach with locality-awareness for {3D}
  skeleton based person re-identification,'' \emph{arXiv:2009.03671}, September
  2020.

\bibitem{ssl1}
P.~Khosla, P.~Teterwak, C.~Wang, A.~Sarna, Y.~Tian, P.~Isola, A.~Maschinot,
  C.~Liu, and D.~Krishnan, ``Supervised contrastive learning,''
  \emph{arXiv:2004.11362}, December 2020.

\bibitem{ssl2}
T.~Chen, S.~Kornblith, M.~Norouzi, and G.~Hinton, ``A simple framework for
  contrastive learning of visual representations,'' in \emph{International
  conference on machine learning}, Vienna, Austria, July 2020.

\bibitem{self3}
I.~Misra and L.~v.~d. Maaten, ``Self-supervised learning of pretext-invariant
  representations,'' in \emph{Computer Vision and Pattern Recognition},
  Seattle, WA, USA, 2020.

\bibitem{pritam1}
P.~{Sarkar} and A.~{Etemad}, ``Self-supervised learning for ecg-based emotion
  recognition,'' in \emph{International Conference on Acoustics, Speech and
  Signal Processing}, Barcelona, Spain, May 2020.

\bibitem{pritam2}
P.~Sarkar and A.~Etemad, ``Self-supervised ecg representation learning for
  emotion recognition,'' \emph{IEEE Transactions on Affective Computing}, vol.
  in press, pp. 1--13, August 2020.

\bibitem{pritamR}
A.~Saeed, T.~Ozcelebi, and J.~Lukkien, ``Multi-task self-supervised learning
  for human activity detection,'' \emph{ACM on Interactive, Mobile, Wearable
  and Ubiquitous Technologies}, vol.~3, no.~2, pp. 1--30, 2019.

\bibitem{multitask1}
M.~Crawshaw, ``Multi-task learning with deep neural networks: A survey,''
  \emph{arXiv preprint arXiv:2009.09796}, September 2020.

\bibitem{multitask2}
Y.~Zhang and Q.~Yang, ``A survey on multi-task learning,'' \emph{arXiv preprint
  arXiv:1707.08114}, July 2017.

\bibitem{MLT6}
S.~Liu, E.~Johns, and A.~J. Davison, ``End-to-end multi-task learning with
  attention,'' in \emph{Computer Vision and Pattern Recognition}, Long Beach,
  CA, USA, June 2019.

\bibitem{MTL1}
N.~Sarafianos, T.~Giannakopoulos, C.~Nikou, and I.~A. Kakadiaris, ``Curriculum
  learning for multi-task classification of visual attributes,'' in
  \emph{International Conference on Computer Vision Workshops}, Venice, Italy,
  October 2017.

\bibitem{MTL2}
D.~Kollias and S.~Zafeiriou, ``Expression, affect, action unit recognition:
  Aff-wild2, multi-task learning and arcface,'' \emph{arXiv:1910.04855},
  September 2019.

\bibitem{MTLG}
M.~J. {Marín-Jiménez}, F.~M. {Castro}, N.~{Guil}, F.~{de la Torre}, and
  R.~{Medina-Carnicer}, ``Deep multi-task learning for gait-based biometrics,''
  in \emph{IEEE International Conference on Image Processing}, Beijing, China,
  September 2017.

\bibitem{MLTLoss}
S.~Chennupati, G.~Sistu, S.~Yogamani, and S.~A~Rawashdeh, ``Multinet++:
  Multi-stream feature aggregation and geometric loss strategy for multi-task
  learning,'' in \emph{Computer Vision and Pattern Recognition}, Long Beach,
  CA, USA, June 2019.

\bibitem{stylegan1}
T.~Karras, S.~Laine, and T.~Aila, ``A style-based generator architecture for
  generative adversarial networks,'' in \emph{Computer Vision and Pattern
  Recognition}, Long Beach, CA, USA, June 2019.

\bibitem{stylegan2}
T.~Karras, S.~Laine, M.~Aittala, J.~Hellsten, J.~Lehtinen, and T.~Aila,
  ``Analyzing and improving the image quality of {StyleGAN},'' in
  \emph{Computer Vision and Pattern Recognition}, Seattle, WA, USA, 2020.

\bibitem{frameGAN}
W.~Xue, H.~Ai, T.~Sun, C.~Song, Y.~Huang, and L.~Wang, ``Frame-gan: Increasing
  the frame rate of gait videos with generative adversarial networks,''
  \emph{Neurocomputing}, vol. 380, pp. 95--104, March 2020.

\bibitem{MVG}
X.~{Chen}, X.~{Luo}, J.~{Weng}, W.~{Luo}, H.~{Li}, and Q.~{Tian}, ``Multi-view
  gait image generation for cross-view gait recognition,'' \emph{IEEE
  Transactions on Image Processing}, vol. in press, pp. 1--15, February 2021.

\bibitem{faceDB1}
A.~Kortylewski, B.~Egger, A.~Schneider, T.~Gerig, A.~Morel-Forster, and
  T.~Vetter, ``Analyzing and reducing the damage of dataset bias to face
  recognition with synthetic data,'' in \emph{Computer Vision and Pattern
  Recognition Workshops}, Long Beach, CA, USA, June 2019.

\bibitem{faceDB2}
B.~Dolhansky, J.~Bitton, B.~Pflaum, J.~Lu, R.~Howes, M.~Wang, and C.~C. Ferrer,
  ``The deepfake detection challenge dataset,'' \emph{arXiv:2006.07397},
  October 2020.

\bibitem{100kface}
\BIBentryALTinterwordspacing
100,000 faces generated by {AI}. [Online]. Available:
  \url{https://generated.photos/}
\BIBentrySTDinterwordspacing

\bibitem{faceDB3}
H.~Dang, F.~Liu, J.~Stehouwer, X.~Liu, and A.~K. Jain, ``On the detection of
  digital face manipulation,'' in \emph{Computer Vision and Pattern
  Recognition}, Seattle, WA, USA, 2020.

\bibitem{faceDB4}
J.~C. Neves, R.~Tolosana, R.~Vera-Rodriguez, V.~Lopes, H.~Proen{\c{c}}a, and
  J.~Fierrez, ``Ganprintr: Improved fakes and evaluation of the state of the
  art in face manipulation detection,'' \emph{IEEE Journal of Selected Topics
  in Signal Processing}, vol.~14, no.~5, pp. 1038--1048, July 2020.

\bibitem{domainadapt3}
M.~Wang and W.~Deng, ``Deep visual domain adaptation: A survey,''
  \emph{Neurocomputing}, vol. 312, pp. 135--153, October 2018.

\bibitem{domainadapt4}
G.~Csurka, ``Domain adaptation for visual applications: A comprehensive
  survey,'' \emph{arXiv:1702.05374}, March 2017.

\bibitem{domainadapt}
J.~Chen, Y.~Li, K.~Ma, and Y.~Zheng, ``Generative adversarial networks for
  video-to-video domain adaptation.'' in \emph{AAAI Conference on Artificial
  Intelligence}, New York City, NY, USA, February 2020.

\bibitem{domainadapt2}
S.~Huang, C.~Lin, S.~Chen, Y.~Wu, P.~Hsu, and S.~Lai, ``{AugGAN}: Cross domain
  adaptation with {GAN-based} data augmentation,'' in \emph{European Conference
  on Computer Vision}, Munich, Germany, September 2018.

\bibitem{cycleGAN}
J.-Y. Zhu, T.~Park, P.~Isola, and A.~A. Efros, ``Unpaired image-to-image
  translation using cycle-consistent adversarial networks,'' in
  \emph{International Conference on Computer Vision}, Venice, Italy, October
  2017.

\bibitem{anim}
F.~Multon, L.~France, M.-P. Cani-Gascuel, and G.~Debunne, ``Computer animation
  of human walking: a survey,'' \emph{The journal of visualization and computer
  animation}, vol.~10, no.~1, pp. 39--54, June 1999.

\bibitem{GANDB}
H.~Dou, W.~Zhang, P.~Zhang, Y.~Zhao, S.~Li, Z.~Qin, F.~Wu, L.~Dong, and X.~Li,
  ``{VersatileGait}: A large-scale synthetic gait dataset with fine-grained
  attributes and complicated scenarios,'' \emph{arXiv:2101.01394}, January
  2021.

\bibitem{transfer}
J.~Zhang, W.~Li, and P.~Ogunbona, ``Transfer learning for cross-dataset
  recognition: a survey,'' \emph{arXiv:1705.04396}, May 2017.

\bibitem{OOD2}
Y.~Bengio, F.~Bastien, A.~Bergeron, N.~Boulanger-Lewandowski, T.~Breuel,
  Y.~Chherawala, M.~Cisse, M.~C{\^o}t{\'e}, D.~Erhan, J.~Eustache
  \emph{et~al.}, ``Deep learners benefit more from out-of-distribution
  examples,'' in \emph{International Conference on Artificial Intelligence and
  Statistics}, Lauderdale, FL, USA, April 2011.

\bibitem{OOD}
D.~Teney, K.~Kafle, R.~Shrestha, E.~Abbasnejad, C.~Kanan, and A.~Hengel, ``On
  the value of out-of-distribution testing: An example of {Goodhart's} law,''
  \emph{arXiv:2005.09241}, May 2020.

\bibitem{joint}
A.~Sepas-Moghaddam, A.~Etemad, F.~Pereira, and P.~Correia, ``Long short-term
  memory with gate and state level fusion for light field-based face
  recognition,'' \emph{IEEE Transactions on Information Forensics and
  Security}, vol.~16, no.~1, pp. 1365--1379, November 2020.

\bibitem{LSTMV1}
J.~{Liu}, A.~{Shahroudy}, D.~{Xu}, A.~C. {Kot}, and G.~{Wang}, ``Skeleton-based
  action recognition using spatio-temporal {LSTM} network with trust gates,''
  \emph{IEEE Transactions on Pattern Analysis and Machine Intelligence},
  vol.~40, no.~12, pp. 3007--3021, November 2018.

\bibitem{mplstm}
A.~Sepas-Moghaddam, F.~Pereira, P.~L. Correia, and A.~Etemad,
  ``Multi-perspective lstm for joint visual representation learning,'' in
  \emph{Computer Vision and Pattern Recognition}, Virtual, June 2021.

\bibitem{MVLSTM}
S.~S. Rajagopalan, L.-P. Morency, T.~Baltrusaitis, and R.~Goecke, ``Extending
  long short-term memory for multi-view structured learning,'' in
  \emph{European Conference on Computer Vision}, Amsterdam, Netherlands,
  October 2016.

\bibitem{non-stationary}
B.~H.~D. Koh and W.~L. Woo, ``Multi-view temporal ensemble for classification
  of non-stationary signals,'' \emph{IEEE Access}, vol.~7, pp.
  32\,482--32\,491, March 2019.

\bibitem{liem2014joint}
M.~C. Liem and D.~M. Gavrila, ``Joint multi-person detection and tracking from
  overlapping cameras,'' \emph{Computer Vision and Image Understanding}, vol.
  128, pp. 36--50, November 2014.

\bibitem{moving}
B.~Heo, K.~Yun, and J.~Y. Choi, ``Appearance and motion based deep learning
  architecture for moving object detection in moving camera,'' in
  \emph{International Conference on Image Processing}, Beijing, China,
  September 2017.

\bibitem{cluster}
S.~A. Shah and V.~Koltun, ``Robust continuous clustering,'' \emph{Proceedings
  of the National Academy of Sciences}, vol. 114, no.~37, pp. 9814--9819,
  August 2017.

\bibitem{comb}
H.~Chen, P.~Guo, P.~Li, G.~H. Lee, and G.~Chirikjian, ``Multi-person 3d pose
  estimation in crowded scenes based on multi-view geometry,'' in
  \emph{European Conference on Computer Vision}, Glasgow, UK, August 2020.

\bibitem{PAMINew}
M.~P. {Vo}, E.~{Yumer}, K.~{Sunkavalli}, S.~{Hadap}, Y.~A. {Sheikh}, and S.~G.
  {Narasimhan}, ``Self-supervised multi-view person association and its
  applications,'' \emph{IEEE Transactions on Pattern Analysis and Machine
  Intelligence}, vol. in press, pp. 1--14, February 2020.

\bibitem{ear2}
S.~Dodge, J.~Mounsef, and L.~Karam, ``Unconstrained ear recognition using deep
  neural networks,'' \emph{IET Biometrics}, vol.~7, no.~3, pp. 207--214, May
  2018.

\bibitem{ear3}
A.~Sepas-Moghaddam, F.~Pereira, and P.~L. Correia, ``Ear recognition in a light
  field imaging framework: a new perspective,'' \emph{IET Biometrics}, vol.~7,
  no.~3, pp. 224--231, May 2018.

\bibitem{fusebio}
L.~M. {Dinca} and G.~P. {Hancke}, ``The fall of one, the rise of many: A survey
  on multi-biometric fusion methods,'' \emph{IEEE Access}, vol.~5, pp.
  6247--6289, April 2017.

\bibitem{fusebio2}
E.~L. Oliveira, C.~A. Lima, and S.~M. Peres, ``Fusion of face and gait for
  biometric recognition: Systematic literature review,'' in \emph{Brazilian
  Symposium on Information Systems on Brazilian Symposium on Information
  Systems: Information Systems in the Cloud Computing Era}, Porto Alegre,
  Brazil, May 2016.

\bibitem{age}
A.~Sakata, N.~Takemura, and Y.~Yagi, ``Gait-based age estimation using
  multi-stage convolutional neural network,'' \emph{IPSJ Transactions on
  Computer Vision and Applications}, vol.~11, no.~1, pp. 1--10, June 2019.

\bibitem{height}
C.~BenAbdelkader, R.~Cutler, and L.~Davis, ``View-invariant estimation of
  height and stride for gait recognition,'' in \emph{International Workshop on
  Biometric Authentication}, Copenhagen, Denmark, June 2002.

\bibitem{height2}
K.~Koide and J.~Miura, ``Identification of a specific person using color,
  height, and gait features for a person following robot,'' \emph{Robotics and
  Autonomous Systems}, vol.~84, pp. 76--87, October 2016.

\bibitem{weight}
M.~M. Samson, A.~Crowe, P.~De~Vreede, J.~A. Dessens, S.~A. Duursma, and H.~J.
  Verhaar, ``Differences in gait parameters at a preferred walking speed in
  healthy subjects due to age, height and body weight,'' \emph{Aging clinical
  and experimental research}, vol.~13, no.~1, pp. 16--21, May 2001.

\bibitem{gender}
A.~Jain and V.~Kanhangad, ``Gender classification in smartphones using gait
  information,'' \emph{Expert Systems with Applications}, vol.~93, pp.
  257--266, March 2018.

\bibitem{tattoo}
H.~Han, J.~Li, A.~K. Jain, S.~Shan, and X.~Chen, ``Tattoo image search at
  scale: Joint detection and compact representation learning,'' \emph{IEEE
  transactions on pattern analysis and machine intelligence}, vol.~41, no.~10,
  pp. 2333--2348, January 2019.

\bibitem{multi1}
X.~{Xing}, K.~{Wang}, and Z.~{Lv}, ``Fusion of gait and facial features using
  coupled projections for people identification at a distance,'' \emph{IEEE
  Signal Processing Letters}, vol.~22, no.~12, pp. 2349--2353, September 2015.

\bibitem{multi2}
Y.~{Guan}, X.~{Wei}, C.~{Li}, G.~L. {Marcialis}, F.~{Roli}, and
  M.~{Tistarelli}, ``Combining gait and face for tackling the elapsed time
  challenges,'' in \emph{IEEE International Conference on Biometrics: Theory,
  Applications and Systems}, Arlington, VA, USA, October 2013.

\bibitem{multi3}
A.~E.~K. Ghalleb and N.~E.~B. Amara, ``Remote person authentication in
  different scenarios based on gait and face in front view,'' in
  \emph{International Multi-Conference on Systems, Signals \& Devices},
  Marrakech, Morocco, December 2017.

\bibitem{multi4}
A.~{El Kissi Ghalleb}, R.~{Ben Slamia}, and N.~E. {Ben Amara}, ``Contribution
  to the fusion of soft facial and body biometrics for remote people
  identification,'' in \emph{International Conference on Advanced Technologies
  for Signal and Image Processing}, Monastir, Tunisia, July 2016.

\bibitem{multi5}
A.~{Kale}, A.~K. {Roychowdhury}, and R.~{Chellappa}, ``Fusion of gait and face
  for human identification,'' in \emph{International Conference on Acoustics,
  Speech, and Signal Processing}, Montreal, Canada, May 2004.

\bibitem{multi6}
Q.~Yu, Y.~Yin, G.~Yang, Y.~Ning, and Y.~Li, ``Face and gait recognition based
  on semi-supervised learning,'' in \emph{Chinese Conference on Pattern
  Recognition}, Beijing, China, September 2012.

\bibitem{fusedeep1}
J.~Gao, P.~Li, Z.~Chen, and J.~Zhang, ``A survey on deep learning for
  multimodal data fusion,'' \emph{Neural Computation}, vol.~32, no.~5, pp.
  829--864, May 2020.

\bibitem{fusedeep2}
V.~{Talreja}, M.~C. {Valenti}, and N.~M. {Nasrabadi}, ``Multibiometric secure
  system based on deep learning,'' in \emph{IEEE Global Conference on Signal
  and Information Processing}, Montreal, Canada, March 2017.

\bibitem{fusefuse}
J.~Kim, J.~Koh, Y.~Kim, J.~Choi, Y.~Hwang, and J.~W. Choi, ``Robust deep
  multi-modal learning based on gated information fusion network,'' in
  \emph{Asian Conference on Computer Vision}, Perth, WA, Australia, December
  2018.

\bibitem{fuseatt}
S.~Liu, S.~Yao, J.~Li, D.~Liu, T.~Wang, H.~Shao, and T.~Abdelzaher,
  ``{GIobalFusion}: A global attentional deep learning framework for
  multisensor information fusion,'' \emph{ACM on Interactive, Mobile, Wearable
  and Ubiquitous Technologies}, vol.~4, no.~1, pp. 1--27, March 2020.

\end{thebibliography}

% \ifCLASSOPTIONcaptionsoff
%   \newpage
% \fi

\begin{IEEEbiography}[
{
\includegraphics[width=1in,height=1.21in,clip,keepaspectratio]{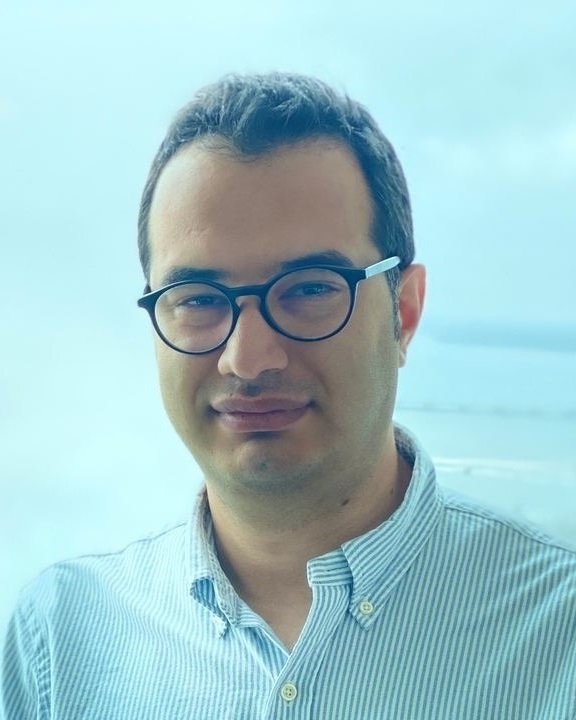}
}
]
{Alireza Sepas-Moghaddam}
received the B.Sc. and M.Sc. (first class honors) degrees in Computer Engineering in 2007 and 2010, respectively. From 2011 to 2015, he was with the Shamsipour Technical University, Tehran, Iran, as a lecturer. In 2015, he joined Instituto Superior Técnico, University of Lisbon, Lisbon, Portugal, where he completed his Ph.D. degree with distinction and honour in Electrical and Computer Engineering in 2019. From 2019 to 2021, he held a postdoctoral fellow position at Queen’s University, Kingston, ON, Canada, where he worked on different research projects funded by the Natural Sciences and Engineering Research Council of Canada (NSERC) and Mitacs, as well as private sector. He is currently working as a senior computer vision data scientist at Socure, dealing with digital identity verification and fraud detection from visual signals. His main research interests are theoretical and applied machine learning, notably deep learning, for biometrics, forensics, and affective computing. He has contributed more than 45 papers in notable conferences and journals in his area, including \textit{CVPR}, \textit{ICCV}, \textit{IEEE T-IP}, \textit{IEEE T-IFS}, \textit{IEEE T-CSVT}, and \textit{IEEE T-Biom}.
He has served as a program chair, publicity chair, and area chair for several conferences and workshops, including \textit{ICPR}, \textit{AAAI-HCSSL}, and \textit{EUVIP} and has been a reviewer for multiple top-tier conferences and journals in the field.  
\end{IEEEbiography}

\begin{IEEEbiography}[
{
\includegraphics[width=1in,height=1.25in,clip,keepaspectratio]{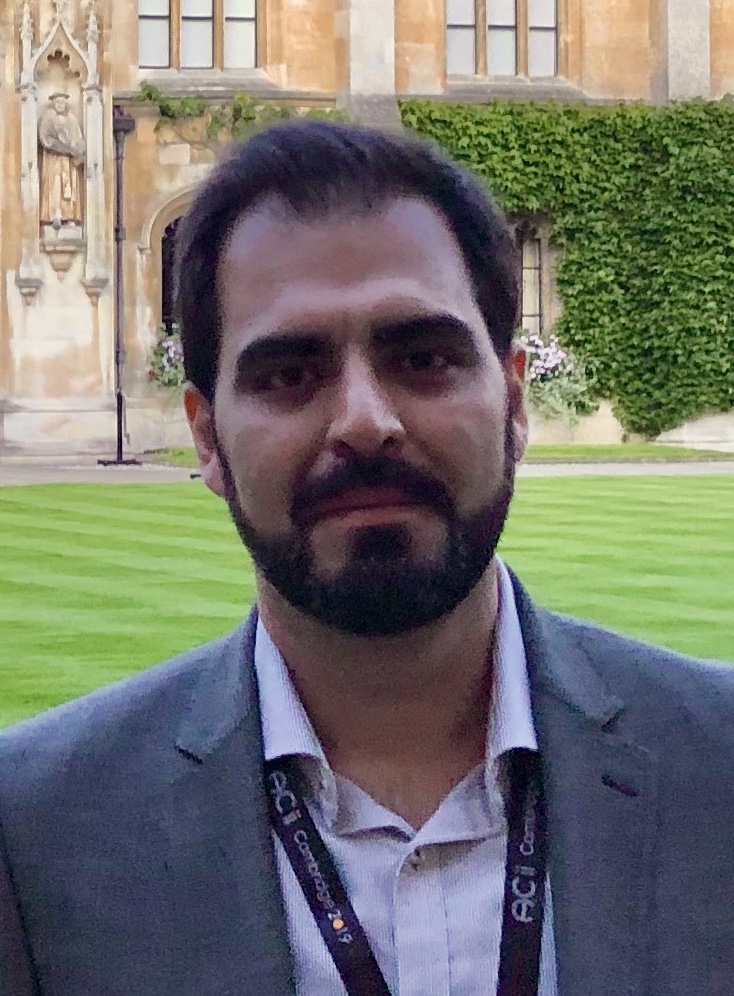}
}
]
{Ali Etemad} 
is an Assistant Professor, as well as a Mitchell Professor in AI for Human Sensing \& Understanding at the Department of Electrical and Computer Engineering, and Ingenuity Labs Research Institute, Queen’s University, Canada. He leads the Ambient Intelligence and Interactive Machines (Aiim) lab, where his main area of research is machine learning and deep learning focused on human-centered applications with wearables, smart devices, and smart environments. 
% Prior to joining Queen’s, he held several industrial positions as lead scientist.
His work has appeared in top-tier venues such as CVPR, AAAI, ICCV, ACM CHI, ICASSP, Interspeech, ICPR, FG, T-AFFC, T-IP, T-AI, T-ASLP, T-IFS, T-BIOM, T-HMS, T-NSRE, IEEE T-GRS, IEEE IoT J., and others. He is a co-inventor of 9 patents and has given over 20 invited talks at different venues. 
Dr. Etemad is an Associate Editor for IEEE Transactions on Artificial Intelligence, and has been a PC member/reviewer for many notable conferences and journals in the field. 
% including NeurIPS, ICML, CVPR, ICLR, ACII (senior PC), ICASSP, ISWC, and ICMI, 
% VRST, BSN, Interspeech, CHI, SMC, AIVR, SAP (senior PC), among others. 
% He is also an active reviewer for journals such as T-AFFC, T-ASLP, T-NNLS, T-AI, T-HMS, T-KDD, T-NSRE, JBHI, Patt. Rec., Neural. Comput. Appl., CVIU, Appl. Intell., and many others. 
He has been the General Chair for the AAAI Workshop on Human-Centric Self-Supervised Learning (2022), Publicity Co-Chair for European Workshop on Visual Information Processing (2022), and Industry Relations Chair for Canadian Conference on AI (2019).
% He has received a number of awards including Supervisor of the Year Award (at Queen’s), Instructor of the Year Award (at Queen’s), and several Best Paper Awards (e.g., at CGI). 
Dr. Etemad’s lab and research program have been funded by the Natural Sciences and Engineering Research Council (NSERC) of Canada, Ontario Centers of Excellence (OCE), Canadian Foundation for Innovation (CFI), Mitacs, and other organizations, as well as the private sector.

\end{IEEEbiography}

\end{document}